\documentclass[a4paper, 10pt]{article}
\usepackage[preprint]{tmlr}

\usepackage{amsmath,amsfonts,bm}

\def\eqref#1{equation~\ref{#1}}

\def\1{\bm{1}}

\DeclareMathAlphabet{\mathsfit}{\encodingdefault}{\sfdefault}{m}{sl}
\SetMathAlphabet{\mathsfit}{bold}{\encodingdefault}{\sfdefault}{bx}{n}

\usepackage[margin=1in]{geometry}
\usepackage{setspace}
\usepackage{microtype}
\usepackage{quantikz}

\usepackage{xcolor}
\usepackage{tikz}
\usetikzlibrary{tikzmark}
\newcommand{\algohlmath}[1]{%
  \colorbox{yellow!30}{$\displaystyle #1$}%
}

\newcommand{\phasehighlight}[2]{%
\begin{tikzpicture}[remember picture,overlay]
\fill[
  yellow!30,
  rounded corners=1pt
]
  ([xshift=-0.4em,yshift=2.2ex]pic cs:#1)
  rectangle
  ([xshift=13.5em,yshift=-0.3ex]pic cs:#2);
\end{tikzpicture}%
}

\usepackage{amsmath} \allowdisplaybreaks
\usepackage{amssymb}
\usepackage{amsthm} \theoremstyle{definition}

\usepackage{natbib}
\usepackage{bibentry}

\usepackage{booktabs}
\usepackage{array}

\usepackage{graphicx}
\usepackage{multirow}
\usepackage{makecell}
\usepackage{threeparttable}
\usepackage{caption}
\usepackage{quantikz}

\usepackage{appendix}
\usepackage{soul} %
\usepackage[hyphens]{url}
\usepackage[bookmarks=false,hidelinks]{hyperref}
\usepackage{cleveref}

\usepackage{subcaption}
\usepackage{algorithm}
\usepackage{algpseudocode}
\usepackage{float}

\usepackage{tikz}
\usepackage{pgfplots}
\pgfplotsset{compat=1.18}
\usetikzlibrary{positioning}
\usetikzlibrary{arrows,shapes}
\usetikzlibrary{arrows.meta}
\usetikzlibrary{snakes}
\usetikzlibrary{calc, backgrounds}
\definecolor{slategray}{RGB}{112,128,144}

\usepackage{authblk}

\usepackage[shortlabels]{enumitem}

\usepackage[makeroom]{cancel}

\usepackage{adjustbox}

\usepackage[most]{tcolorbox}
\usepackage{caption}

\newtcolorbox{promptbox}{
  breakable,
  colback=gray!5,
  colframe=gray!35,
  boxrule=0.6pt,
  arc=3pt,
  left=6pt,
  right=6pt,
  top=6pt,
  bottom=6pt,
  width=\linewidth,
  before skip=2mm,
  after skip=1mm,
  fontupper=\ttfamily\small\setstretch{1.0}
}

\usepackage{bm}

\usepackage{xcolor}

\usepackage{etoolbox}

\usepackage[normalem]{ulem}

\makeatletter
\def\do#1{\@namedef{#1c}{\ensuremath{\mathcal{#1}}}}
\docsvlist{A,B,C,D,E,F,G,H,I,J,K,L,M,N,O,P,Q,R,S,T,U,V,W,X,Y,Z}
\makeatother

\renewcommand{\bar}[1]{\mkern 1.5mu\overline{\mkern-1.5mu#1\mkern-1.5mu}\mkern 1.5mu}

\renewcommand{\epsilon}{\varepsilon}

\title{LLM-Driven Algorithm Design for Quantum Circuit Synthesis based on Binary Decision Diagrams}

\author{%
    \name Yoonju Sim$^{1}$\email syj5268@kaist.ac.kr\par
    \name Federico Berto$^{2}$\email berto.federico2@gmail.com\par
    \name Chuanbo Hua$^{1}$\email cbhua@kaist.ac.kr\par
    \name Jinkyoo Park$^{1,3}$\email jinkyoo.park@kaist.ac.kr\par
    \name Changhyun Kwon$^{1,3,\dag}$\email chkwon@kaist.ac.kr\par
    \vspace{0.3em}%
    \addr $^1$ Department of Industrial and Systems Engineering, KAIST, Daejeon, 34141, Republic of Korea\par
    \addr $^2$ Radical Numerics, California, USA\par
    \addr $^3$ Omelet, Inc., Daejeon, 34051, Republic of Korea\par
    \addr $\dag$ Corresponding author
}

\def\month{09}  %
\def\year{2026} %
\def\openreview{\url{https://openreview.net/forum?id=iL7YeQp7je}} %

\begin{document}
\maketitle

\begin{abstract}
Quantum circuits are central to implementing quantum algorithms on quantum devices, where quantum gates must be reversible. Many quantum algorithms rely on Boolean functions, which must therefore be implemented reversibly within quantum circuits. Reversible circuit synthesis provides a way to translate such Boolean functions into reversible circuits. Binary decision diagrams (BDDs) offer a scalable approach to this task, but the resulting BDDs and circuits depend heavily on variable ordering. Existing ordering heuristics commonly minimize BDD size because it is closely tied to the circuit size. However, BDD size is an imperfect proxy for the quantum cost of the synthesized circuit (QCC). We propose \texttt{QuantumEvo}, an evolutionary framework that uses an LLM as a heuristic generator for QCC-aware BDD variable ordering. Instead of predicting orderings directly, \texttt{QuantumEvo} searches over ordering heuristics initialized from multiple heuristic families. Candidate heuristics directly manipulate variable orderings using standard BDD operations and are selected by downstream QCC. The discovered heuristic, HGA-QE, modifies the sifting step inside a genetic algorithm so that the procedure is better aligned with QCC. Across the benchmark set, HGA-QE achieves a 70.9\% tie-or-win rate against the per-function best baseline and is strictly best on 13.5\% of the functions. The results demonstrate broadly competitive QCC performance, with HGA-QE showing a clearer relative advantage in strict wins on the two benchmark suites drawn from sources different from the data used for heuristic discovery.

\end{abstract}

\section{Introduction} 
\label{sec:introduction}

Quantum computing has attracted increasing attention with advances in both quantum algorithms and quantum hardware~\citep{fedorov2022quantum, blekos2024review, bluvstein2024logical}. 
To run a quantum algorithm on a quantum device, its operations must be expressed as a quantum circuit, i.e., a cascade of quantum gates. 
Since quantum operations are unitary, quantum circuits are \emph{reversible} in nature.
Many quantum algorithms contain Boolean components, such as oracle functions~\citep{grover1996fast, simon1997power} and arithmetic subroutines~\citep{vedral1996quantum}. 
To implement such components in a quantum circuit, a Boolean function must be embedded into a reversible transformation.
Reversible circuit synthesis addresses this task by translating Boolean functions into reversible circuits that can be mapped to a target gate library~\citep{shende2003synthesis, wille2016reversible}.
The quality of a synthesized circuit is often evaluated by its quantum cost, which reflects the cost of implementing the circuit under a quantum gate library.
We refer to this metric as the quantum cost of the synthesized circuit (QCC).

Several approaches have been developed for reversible circuit synthesis, including transformation-based~\citep{miller2003transformation} and cycle-based~\citep{saeedi2010reversible}.
Such methods can produce high-quality circuits for small functions, but their scalability is limited as the number of inputs grows. 
A BDD-based synthesis aims to improve scalability by representing a Boolean function as a binary decision diagram (BDD) and translating the resulting BDD structure into a reversible circuit~\citep{wille2009bdd}.
Since variable ordering strongly affects the resulting BDD, and the BDD size provides an upper bound on the circuit size, minimizing the BDD size is a common objective for ordering methods.
Previous studies on variable ordering methods, including metaheuristics~\citep{bollig1995simulated, awad2022genetic} and learning-based methods~\citep{miao2026bdd2seq}, have used the BDD size as an objective.
However, BDD size is an imperfect proxy for QCC. 
A smaller BDD does not necessarily imply a lower QCC. 
This motivates the search for BDD variable ordering heuristics that directly reduce downstream QCC.

We formulate this as a heuristic design problem.
The goal is to discover a reusable BDD variable ordering heuristic that maps each Boolean function to a variable ordering with low QCC. 
Recent LLM-driven algorithm design methods use LLMs to explore a broader space of heuristic programs beyond a predefined set~\citep{novikov2025alphaevolve, ye2024reevo, liu2026eohs, huang2026calm}. 
Building on this idea, we propose \texttt{QuantumEvo}\footnote{Source code: \url{https://github.com/syj5268/Quantumevo}}, an evolutionary framework that uses an LLM as a heuristic generator.
It initializes the search from multiple heuristic families and evaluates executable heuristics that manipulate variable ordering using standard BDD operations.

Our main contributions are as follows: 
(1) We formulate QCC-aware BDD variable ordering as a heuristic design problem and introduce \texttt{QuantumEvo}, an LLM-driven heuristic search framework that evaluates executable ordering heuristics using downstream QCC rather than BDD size;
(2) \texttt{QuantumEvo} discovers HGA-QE, an effective hybrid genetic algorithm that replaces standard sifting with an LLM-discovered targeted sifting procedure better aligned with downstream QCC. Across the benchmark set, HGA-QE demonstrates broadly competitive QCC performance against both classical and learning-based baselines, with a clearer relative advantage in strict wins on the two benchmark suites drawn from sources different from the data used for heuristic discovery.

\section{Background and Related Work} 
This section reviews the background and related work relevant to our approach.

\subsection{Reversible Logic and Circuit Synthesis}
\label{subsec:reversible_synthesis}

A Boolean function $f:\mathbb{B}^n \rightarrow \mathbb{B}^m$ maps $n$ binary input bits to $m$ binary output bits, where $\mathbb{B}=\{0,1\}$.
A reversible function is a bijection with the same number of input and output bits.
While general Boolean functions are not necessarily bijective,  they can be embedded into reversible functions by introducing ancilla inputs and garbage outputs~\citep{wille2010towards}. 
Due to its information-preserving property, reversible logic is relevant to a range of applications, most notably quantum computing, low power design, optical computing, and nanotechnologies~\citep{reversiblelogic1961, saeedi2013synthesis, cuykendall1987reversible, merkle1993reversible}.

Reversible circuit synthesis realizes Boolean functions as circuits over a gate library.
We focus on the widely used \emph{NCT library}, which comprises NOT, CNOT, and Toffoli gates; see \Cref{app:circuit-example}.
Each gate in the NCT library with \(k\in\{0,1,2\}\) control bits is assigned a QCC of
$\mathrm{Cost}(k) = 2k^2 - 2k + 1$ 
based on its decomposition into elementary quantum gates, following \citet{szyprowski2013low}.
The total QCC of a circuit is the sum of its gate costs.

Before synthesis, a Boolean function must be represented in a form suitable for manipulation.
Although truth tables provide a canonical representation of Boolean functions by enumerating all $2^n$ rows, they become impractical at scale due to their exponential size. 
BDDs provide compact intermediate representations for scalable reversible circuit synthesis~\citep{wille2009bdd, soeken2016fast}.
Following \citet{wille2009bdd}, the synthesis procedure considered in this work constructs reversible circuits by replacing each non-terminal BDD node with a cascade of reversible gates.

\subsection{Binary Decision Diagrams and Variable Ordering}
\label{subsec:bdd_ordering}
\begin{figure}[t]
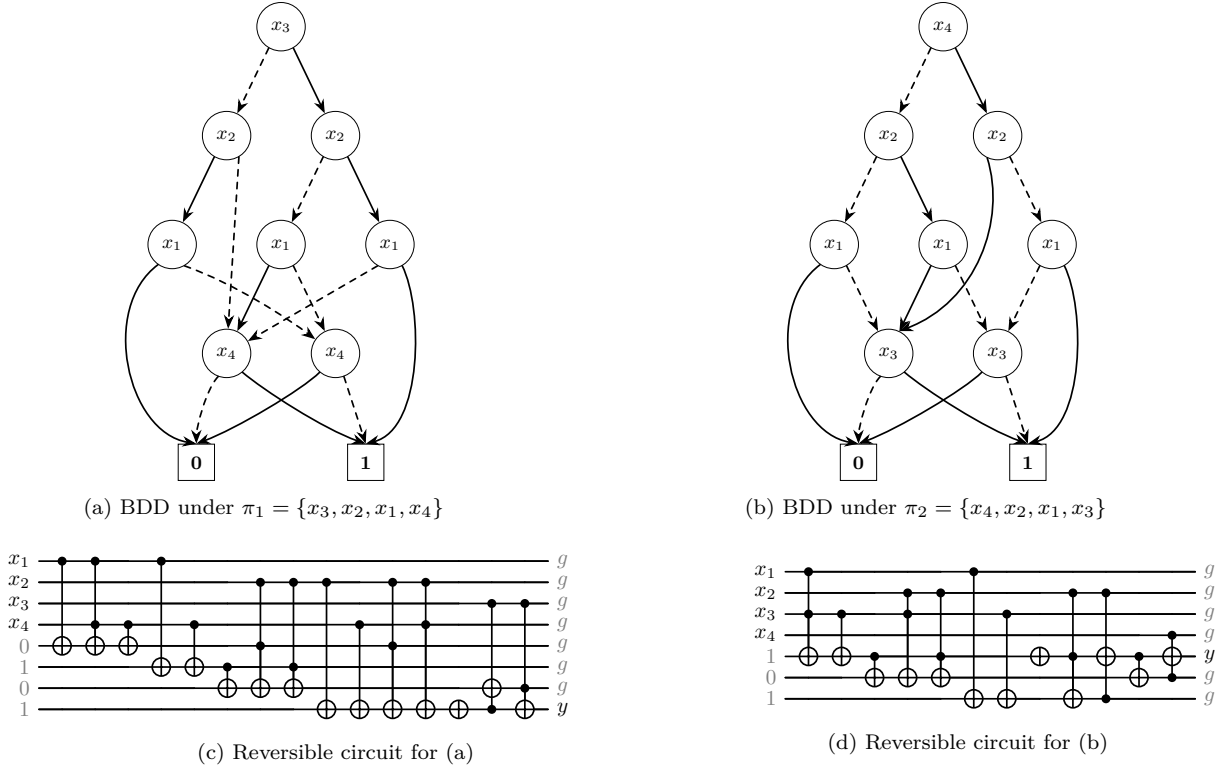

\captionsetup[subfigure]{justification=centering}
    \centering
    \begin{subfigure}[t]{0.45\textwidth}
        \centering
        \scalebox{0.8}{
            \begin{tikzpicture}[
    bdd node/.style={circle, draw, minimum size=8mm, inner sep=0pt, font=\small\bfseries},
    terminal/.style={rectangle, draw, minimum size=6mm, font=\small\bfseries},
    one edge/.style={-{Stealth}, thick, line cap=round, line join=round},
    zero edge/.style={-{Stealth}, dashed, thick, line cap=round, line join=round}
]

    \node[bdd node] (n7) at (0.00,0.00) {$x_3$};
    \node[bdd node] (n3) at (-0.90,-1.80) {$x_2$};
    \node[bdd node] (n6) at (0.90,-1.80) {$x_2$};
    \node[bdd node] (n2) at (-1.80,-3.60) {$x_1$};
    \node[bdd node] (n4) at (0.00,-3.60) {$x_1$};
    \node[bdd node] (n5) at (1.80,-3.60) {$x_1$};
    \node[bdd node] (n0) at (-0.90,-5.40) {$x_4$};
    \node[bdd node] (n1) at (0.90,-5.40) {$x_4$};

    \node[terminal] (const0) at (-1.40,-7.20) {0};
    \node[terminal] (const1) at (1.40,-7.20) {1};

    \draw[zero edge] (n0.250) .. controls (-1.25,-6.00) and (-1.40,-6.65) .. (const0.north);
    \draw[one edge] (n0.310) .. controls (-0.30,-6.00) and (0.70,-6.65) .. (const1.north);
    \draw[zero edge] (n1.290) -- (const1.north);
    \draw[one edge] (n1.230) .. controls (0.35,-6.00) and (-0.75,-6.65) .. (const0.north);
    \draw[zero edge] (n2.300) .. controls (-0.75,-4.25) and (0.15,-4.85) .. (n1.145);
    \draw[one edge] (n2.235) .. controls (-3.00,-4.65) and (-2.55,-6.70) .. (const0.north);
    \draw[zero edge] (n3.300) -- (n0);
    \draw[one edge] (n3.240) -- (n2);
    \draw[zero edge] (n4.300) -- (n1);
    \draw[one edge] (n4.240) -- (n0);
    \draw[zero edge] (n5.240) -- (n0);
    \draw[one edge] (n5.300) .. controls (2.35,-4.65) and (2.35,-6.65) .. (const1.north);
    \draw[zero edge] (n6.240) -- (n4);
    \draw[one edge] (n6.300) -- (n5);
    \draw[zero edge] (n7.240) -- (n3);
    \draw[one edge] (n7.300) -- (n6);

\end{tikzpicture}
        }
        \caption{BDD under $\pi_1 = \{x_3, x_2, x_1, x_4\}$}
    \end{subfigure}
    \hfill
    \begin{subfigure}[t]{0.45\textwidth}
        \centering
        \scalebox{0.8}{
            \begin{tikzpicture}[
    bdd node/.style={circle, draw, minimum size=8mm, inner sep=0pt, font=\small\bfseries},
    terminal/.style={rectangle, draw, minimum size=6mm, font=\small\bfseries},
    one edge/.style={-{Stealth}, thick, line cap=round, line join=round},
    zero edge/.style={-{Stealth}, dashed, thick, line cap=round, line join=round}
]

    \node[bdd node] (n7) at (0.00,0.00) {$x_4$};
    \node[bdd node] (n4) at (-0.90,-1.80) {$x_2$};
    \node[bdd node] (n6) at (0.90,-1.80) {$x_2$};
    \node[bdd node] (n1) at (-1.80,-3.60) {$x_1$};
    \node[bdd node] (n3) at (0.00,-3.60) {$x_1$};
    \node[bdd node] (n5) at (1.80,-3.60) {$x_1$};
    \node[bdd node] (n0) at (-0.90,-5.40) {$x_3$};
    \node[bdd node] (n2) at (0.90,-5.40) {$x_3$};

    \node[terminal] (const0) at (-1.40,-7.20) {0};
    \node[terminal] (const1) at (1.40,-7.20) {1};

    \draw[zero edge] (n0.250) .. controls (-1.25,-6.00) and (-1.40,-6.65) .. (const0.north);
    \draw[one edge] (n0.310) .. controls (-0.30,-6.00) and (0.70,-6.65) .. (const1.north);
    \draw[zero edge] (n1.300) -- (n0);
    \draw[one edge] (n1.235) .. controls (-3.00,-4.65) and (-2.55,-6.70) .. (const0.north);
    \draw[zero edge] (n2.290) -- (const1.north);
    \draw[one edge] (n2.230) .. controls (0.35,-6.00) and (-0.75,-6.65) .. (const0.north);
    \draw[zero edge] (n3.300) -- (n2);
    \draw[one edge] (n3.240) -- (n0);
    \draw[zero edge] (n4.240) -- (n1);
    \draw[one edge] (n4.300) -- (n3);
    \draw[zero edge] (n5.240) -- (n2);
    \draw[one edge] (n5.300) .. controls (2.35,-4.65) and (2.35,-6.65) .. (const1.north);
    \draw[zero edge] (n6.300) -- (n5);
    \draw[one edge] (n6.245) .. controls (1.05,-3.20) and (0.55,-4.55) .. (n0.65);
    \draw[zero edge] (n7.240) -- (n4);
    \draw[one edge] (n7.300) -- (n6);

\end{tikzpicture}
        }
        \caption{BDD under $\pi_2 = \{x_4, x_2, x_1, x_3\}$}
    \end{subfigure}

\vspace{0.3cm}
    \begin{subfigure}[t]{0.57\textwidth}
        \scalebox{0.8}{
            \input{fig/circuit_ga.tex}
        }
        \caption{Reversible circuit for (a)}
    \end{subfigure}
    \hfill
    \begin{subfigure}[t]{0.38\textwidth}
        \scalebox{0.8}{
            \input{fig/circuit_sift.tex}
        }
        \caption{Reversible circuit for (b)}
    \end{subfigure}
\caption{Impact of variable ordering on QCC for the function \texttt{sf\_232}. Solid/dashed edges denote 1/0-branches. BDD size is 8 for both orderings, whereas QCC is 43 for $\pi_1$ and 36 for $\pi_2$. Here, $y=x_2 \oplus x_3 \oplus x_4 \oplus x_1\cdot(x_2 \oplus x_3 \oplus x_2x_3 \oplus x_2x_4)$}.
\label{fig:bdd_ordering}
\end{figure}

A BDD is a directed acyclic graph representation of a Boolean function~\citep{bryant1986graph}.
Each non-terminal node is labeled by one variable and has two branches pointing to terminal or non-terminal child nodes.
When the same subfunction appears in multiple places, it is represented once and reused as a shared subgraph.
The variable ordering is important when constructing a BDD, as it strongly affects the size of the resulting graph.
The BDD size is typically measured by the number of non-terminal nodes, and finding an optimal ordering that minimizes this size is $\mathcal{NP}$-complete~\citep{bollig1996improving}.
Although the BDD size is a natural proxy for the QCC because the size of the resulting circuit is bounded by the BDD size~\citep{wille2009bdd}, it still remains an indirect proxy.
\Cref{fig:bdd_ordering} shows that two orderings of the same function can produce BDDs with the same size but different QCCs.

Prior work on BDD variable ordering includes exact algorithms~\citep{friedman1987finding, jeong1993efficient}, heuristic algorithms such as sifting~\citep{rudell1993dynamic, panda1994symmetry, panda1995variables, meinel1997linear}, and metaheuristics including genetic algorithms (GA), simulated annealing (SA), and swarm-based search~\citep{drechsler1996genetic, bollig1995simulated, awad2022genetic}. 
More recently, BDD2Seq~\citep{miao2026bdd2seq} introduced a learning-based approach that trains a graph-to-sequence model using supervised learning to predict the orderings.
However, all of these methods primarily select the orderings based on BDD size, while downstream QCC is observed only after synthesis.
This motivates a QCC-aware objective that optimizes the orderings by the QCC rather than by the BDD size.

\subsection{Related Work}
We review the LLM-based approaches related to our work.
The first line examines how LLMs have been used in quantum-computing workflows.
Recent Quantum--LLM studies address broad problem settings within the quantum-computing workflow.
Some works evaluate LLMs on generating quantum code from textual task prompts~\citep{QiskitHumanEval, QuanBench}.
Others fine-tune LLMs to generate parameterized quantum circuits for quantum optimization tasks~\citep{AgentQ}.
QCircuitBench~\citep{QCircuitBench} benchmarks LLMs on quantum algorithm design tasks, including oracle construction and the implementation of standard quantum algorithms.
Oracle construction is related to our setting because it often requires implementing Boolean functions as reversible circuits.
However, QCircuitBench focuses mainly on textbook-level or algorithm-specific oracles, whereas \texttt{QuantumEvo} discovers BDD variable ordering heuristics for synthesizing large Boolean functions under downstream QCC as the objective.

The second covers LLM-driven algorithm design, which provides the methodological basis for our approach.
A complementary line of work uses LLMs to generate heuristic programs rather than predict solutions directly.
FunSearch~\citep{romero2024mathematical} introduced LLM-driven algorithm search for mathematical and heuristic design, and ReEvo~\citep{ye2024reevo} extended it with reflective evolution for combinatorial optimization.
Recent frameworks have further advanced LLM-driven algorithm design~\citep{mctsahd2025, novikov2025alphaevolve, ye2025large, meoh2025, hsevo2025, liu2026eohs, huang2026calm}.
This paradigm has also been applied to domain-specific heuristic design~\citep{zhao2026trajevo, llmvd2026, wang2026llm}.
\texttt{QuantumEvo} extends this line of domain-specific work to BDD variable ordering for reversible circuit synthesis. %

\section{Methodology}
\label{sec:methodology}

\begin{figure}
\centering
\includegraphics[width=\textwidth]{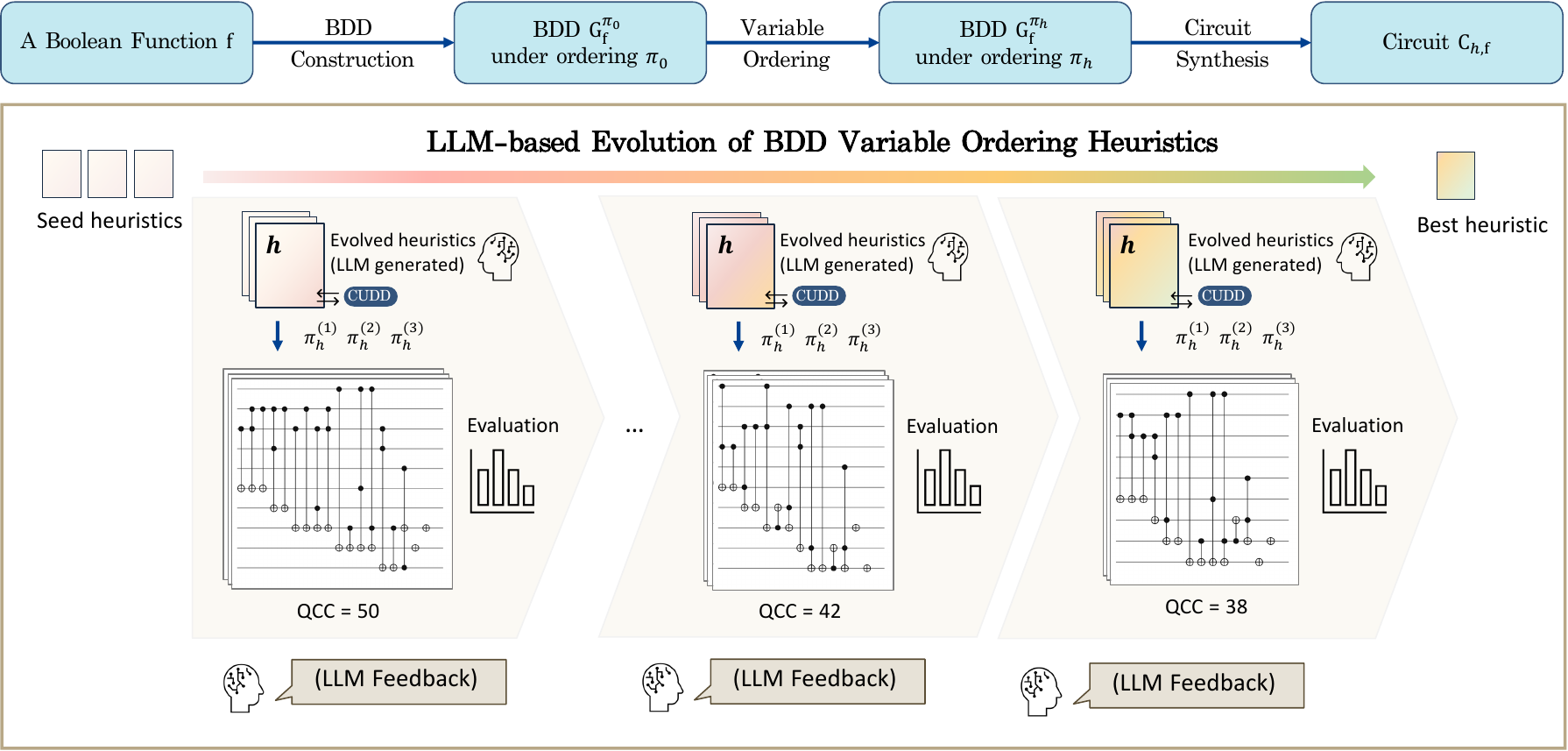}
\caption{Overview of the \texttt{QuantumEvo} framework. Each individual heuristic reorders the initial variable order of BDD, and the resulting circuit is evaluated using QCC.}
\label{fig:framework}
\end{figure}

This section describes our methodology.
We first define the objective used to evaluate BDD variable orderings in terms of QCC.
We then introduce \texttt{QuantumEvo}, an LLM-driven evolutionary framework for discovering BDD variable ordering heuristics.
\Cref{fig:framework} gives an overview of the framework.

\subsection{Problem Formulation}
Given a Boolean function $f:\mathbb{B}^n \rightarrow \mathbb{B}^m$, 
let $\mathcal{H}$ denote the search space of candidate heuristics for BDD variable ordering.
An initial BDD $G_f^{\pi_0}$ is first constructed under an initial variable ordering $\pi_0$.
A heuristic $h\in\mathcal{H}$ then reorders the BDD and produces a variable ordering $\pi_h$ for $f$.
The reordered BDD $G_f^{\pi_h}$ is synthesized into a reversible circuit by a fixed BDD-based synthesis procedure $\mathcal{S}$ following \citet{wille2009bdd}, as implemented in RevKit~\citep{soeken2012revkit}.
The resulting circuit is denoted by
$C_{h,f}=\mathcal{S}\!\left(G_f^{\pi_h}\right).$
Let $q(C)$ denote the QCC of a circuit $C$.
For the given Boolean function $f$, the goal is to find a heuristic that minimizes the QCC of the synthesized circuit within a time budget:
\begin{equation}
\begin{aligned}
    h^* = \arg \min_{h\in\mathcal{H}}
    \quad &
    q\!\left(C_{h,f}\right)
    \\
    \text{s.t.}\quad
    &
    t_{\mathrm{syn}}\!\left(h,f\right)\le\tau_U,
\end{aligned}
\label{eq:heuristic_objective}
\end{equation}
where $t_{\mathrm{syn}}(h,f)$ denotes the total time required to construct the BDD for $f$ using the ordering produced by heuristic $h$ and synthesize the resulting circuit, and $\tau_U$ is the time budget.

\subsection{QuantumEvo: LLM-driven Evolutionary Heuristic Design}
\label{subsec:evolutionary_search}

\texttt{QuantumEvo} primarily builds on the reflective evolutionary search framework of ReEvo~\citep{ye2024reevo}, in which LLMs generate executable heuristic programs that are iteratively improved through evaluation and feedback. 
Starting from a set of seed heuristics, the LLM proposes new heuristics by revising existing ones, guided by natural-language feedback on previous fitness results.
We instantiate this paradigm for BDD variable ordering through three domain-specific design choices that shape the search process: how individual heuristics are represented, how the initial heuristics are constructed, and how the fitness function is defined.

\paragraph{Individual representation}
Each individual is represented as a \texttt{C} language function linked with CUDD (Colorado University Decision Diagram), a widely used \texttt{C} library for constructing and manipulating decision diagrams.
This allows generated heuristics to operate directly on the BDD manager, avoiding the need to implement low-level BDD operations from scratch.
The relevant CUDD API calls are listed in \Cref{app:impl-notes}.

\paragraph{Initial heuristics construction}
The initial population is seeded using three established BDD variable ordering methods: sifting~\citep{rudell1993dynamic}, GA~\citep{drechsler1996genetic}, and SA~\citep{bollig1995simulated}.
Whereas ReEvo initializes its search from a single seed heuristic, \texttt{QuantumEvo} uses seed heuristics from these three distinct algorithmic families.
Each method represents a different search strategy, allowing the LLM-driven evolutionary process to recombine and adapt their algorithmic components when generating new heuristics.

\paragraph{Fitness Function}
\texttt{QuantumEvo} defines the fitness function using the downstream QCC.
Each valid individual is compiled, executed on the search set, and evaluated after full reversible circuit synthesis.
Let $\mathcal{T}$ denote the search set used during evolution.
For each instance $f \in \mathcal{T}$, let $q_f^{\mathrm{base}}$ denote the lowest QCC achieved on $f$ among the classical baseline methods considered in this work.
The fitness of a generated heuristic $h$ is defined as the average relative QCC gap to the instance-wise best baseline:
\begin{equation}
    \mathrm{fitness}(h)
    =\frac{1}{|\mathcal{T}|}
    \sum_{f \in \mathcal{T}}
    \frac{q(C_{h,f}) - q_f^{\mathrm{base}}}{q_f^{\mathrm{base}}
    }.
\end{equation}
Lower values are better, and negative values indicate improvement over the best baseline for that instance.
Because generated heuristics may contain stochastic components, each valid individual is evaluated multiple times per instance.

\subsection{Analysis of the Discovered Algorithm}
\label{subsec:proposed_algorithm}

\begin{figure}
\centering
\includegraphics[width=0.9\textwidth]{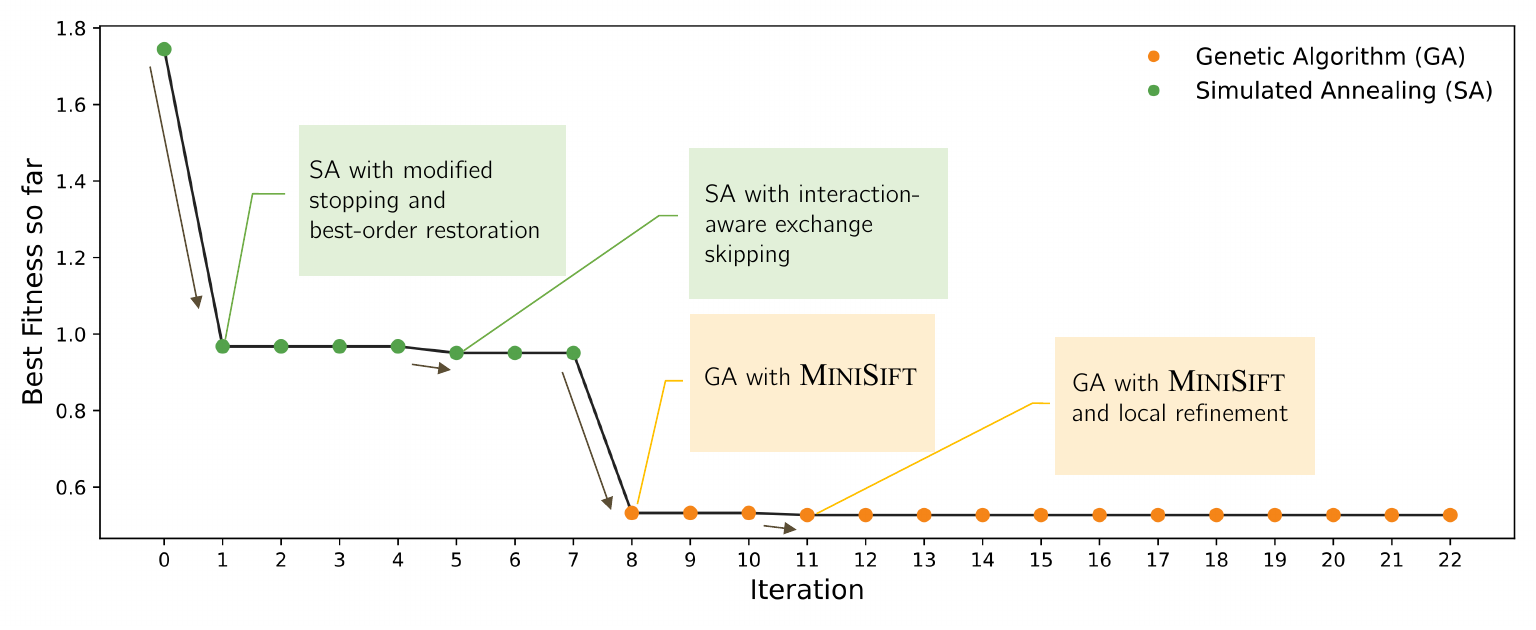}
\caption{Fitness of the best individual over iterations, showing the family transition at iteration~8. Lower values indicate better fitness.}
\label{fig:discovered_lineage}
\end{figure}

We analyze the best heuristic selected by \texttt{QuantumEvo}, tracing its evolution across iterations.
\Cref{fig:discovered_lineage} shows the fitness trajectory of the overall best individual.
In the best run, the best-so-far individual starts from an SA-family heuristic, transitions to a GA-family heuristic at iteration~8, and later incorporates a local refinement step.
We refer to the best heuristic as HGA-QE, a hybrid GA discovered by \texttt{QuantumEvo} using a targeted sifting procedure.
HGA-QE uses two generated subroutines, \textsc{PartialSift} and \textsc{MiniSift}. 
The algorithm is summarized in \Cref{alg:cuddFunc}, where the highlighted phase indicates the modification from the original GA.
The subroutines are detailed in \Cref{app:impl-notes}.
\textsc{PartialSift} reorders the current BDD according to a candidate variable order, identifies the affected levels, and calls \textsc{MiniSift} to locally sift only those levels.
This replaces the standard sifting step used in the original GA, while keeping the same fitness function based on inverse BDD size. 

\Cref{tab:ga_sift_ablation_train} shows that this replacement improves QCC while reducing runtime. 
The main gain comes from \textsc{MiniSift}, while the additional local refinement step gives only marginal improvement. 
We infer that the improvement is related to greater diversity among candidate orderings and the resulting BDD structures.
The standard sifting tends to pull different candidates toward similar BDD structures optimized for BDD size, whereas \textsc{MiniSift} produces more diverse orderings with comparable BDD size. 
The diagnostic analysis in \Cref{app:sift_minisift_compare} supports this interpretation and further examines cases where the BDD size is unchanged but \textsc{MiniSift} lowers QCC. 
In these cases, the resulting BDDs have synthesis-relevant structural differences, accompanied by fewer Toffoli gates and total controls.
This suggests that \textsc{MiniSift} may expose a more diverse set of variable orderings, some of which are more favorable for synthesis and are associated with lower QCC, even though the GA itself still selects candidates by the BDD size.

\begin{algorithm}[h]
\caption{\textbf{HGA-QE}: Hybrid GA discovered by \texttt{QuantumEvo}. Subroutines \textsc{PartialSift} and \textsc{MiniSift} are defined in \Cref{app:impl-notes}. }
\label{alg:cuddFunc}
\begin{algorithmic}[1]
\Require BDD manager $T$, variable range $[\ell, u]$
\Ensure Improved variable ordering in $T$

\Statex \textbf{Phase 1 --- Baseline sifting}
\State $\textsc{Sift}(T,\, \ell,\, u)$ \Comment{existing CUDD API; sift variables to reduce BDD size}
\Statex
\Statex \textbf{Phase 2 --- Population initialization}
\State $n \gets u - \ell + 1$;\quad
$\mathrm{popsize} \gets \max\{4,\min(2n,\;40)\}$;\quad
$\mathrm{nCross} \gets \min(3n,\;\mathrm{popsize},\;60)$  %
\State $P[0] \gets$ current ordering of $T$ after sift;\quad
$P[0].\mathrm{cost} \gets |T.\mathrm{nodes}|$
\State $P[1] \gets$ reverse of $P[0]$
\For{$i = 2$ \textbf{to} $\mathrm{popsize}-1$}
    \State $P[i] \gets$ uniformly random permutation of variables in $[\ell, u]$
\EndFor
\For{$i = 1$ \textbf{to} $\mathrm{popsize}-1$}
    \State $(P[i],\,P[i].\mathrm{cost}) \gets \algohlmath{\textsc{PartialSift}(T,\, P[i],\, \ell,\, u)}$
    \Comment{\Cref{alg:partialsift}}
\EndFor
\Statex
\Statex \textbf{Phase 3 --- Evolutionary loop}
\For{$m = 1$ \textbf{to} $\mathrm{nCross}$}
    \State $(p_1,\, p_2) \gets \textsc{RouletteSelect}(P)$ such that $p_1 \neq p_2$
    \Comment{roulette selection weighted by $1/\mathrm{cost}$}
    \State $(c_1,\, c_2) \gets \textsc{PMX}(P[p_1],\, P[p_2])$ \Comment{PMX crossover for valid permutations}
    \For{each offspring $c \in \{c_1,\, c_2\}$}
        \State $(c,\,c.\mathrm{cost}) \gets \algohlmath{\textsc{PartialSift}(T,\, c,\, \ell,\, u)}$
        \State $w \gets$ worst replaceable individual in $P$ \Comment{do not discard the only tracked copy of a repeated order}
        \If{$c.\mathrm{cost} < P[w].\mathrm{cost}$}
            \State $P[w] \gets c$ \Comment{replace worst with offspring}
        \EndIf
    \EndFor
\EndFor
\State $b \gets \arg\min_{i}\; P[i].\mathrm{cost}$
\State $(P[b],\,P[b].\mathrm{cost}) \gets \algohlmath{\textsc{PartialSift}(T,\, P[b],\, \ell,\, u)}$
\Comment{restore $T$ to the best ordering}
\State $O^\star \gets \text{current ordering of } T;\quad C^\star \gets P[b].\mathrm{cost}$

\Statex
\phasehighlight{phase4start}{phase4end}
\Statex \tikzmark{phase4start}\textbf{Phase 4 --- Local refinement}
\If{$n \ge 2$}
    \State Choose $x \sim \mathrm{Unif}\{\ell,\ldots,u-1\}$
    \State $\textsc{SwapAdjacent}(T,\, x,\, x+1)$
    \Comment{existing CUDD API; exchange two adjacent BDD levels}
    \State $\textsc{MiniSift}(T,\, \ell,\, u,\, \{0,\ldots,n-1\})$
    \State $\mathrm{newCost} \gets |T.\mathrm{nodes}|$
    \If{$\mathrm{newCost} < C^\star$}
        \State $O^\star \gets \text{current ordering of } T$
        \State $C^\star \gets \mathrm{newCost}$
    \Else
        \State $\textsc{RestoreOrder}(T,\, O^\star,\, \ell,\, u)$
        \Comment{revert to the saved best order}
    \EndIf
\EndIf
\tikzmark{phase4end}
\end{algorithmic}
\end{algorithm}

\begin{table}[t]
\centering
\caption{Comparison of GA variants on the search set using best-of-3 seeds. The mean gaps are signed percentage differences relative to the best known value per instance. The win/tie/loss counts are relative to GA with sifting.}
\label{tab:ga_sift_ablation_train}
\resizebox{\linewidth}{!}{%
\begin{tabular}{l rr rr r}
\toprule
       & \multicolumn{2}{c}{QCC} & \multicolumn{2}{c}{BDD} & \\
       \cmidrule(lr){2-3} \cmidrule(lr){4-5}
Method & Mean Gap ($\downarrow$) & W/T/L & Mean Gap ($\downarrow$) & W/T/L & Total Time (s) \\
\midrule
GA with sifting
& 2.096\% & - & $-$0.211\% & - & 23.24 \\
GA with \textsc{MiniSift}
& 0.532\% & 34/53/13 & $-$0.350\% & 4/94/2 & 21.95  \\
GA with \textsc{MiniSift} + local refinement
& 0.527\% & 34/53/13 & $-$0.354\% & 4/94/2 & 22.01 \\
\bottomrule
\end{tabular}
}

\begin{tablenotes}[flushleft]
\footnotesize
\item \emph{Note.} The initial GA heuristic, corresponding to GA with sifting, uses a reduced population for runtime efficiency. The negative mean BDD gap indicates smaller BDDs than the best baseline value.
\end{tablenotes}

\end{table}

\section{Experiment}
\label{sec:experiments}
We compare the best heuristic discovered by \texttt{QuantumEvo} against baselines, and conduct additional studies on the design of the framework.
All experiments are conducted using a single thread on an AMD Ryzen 9 5900X machine with 64 GB of RAM.

\subsection{Experimental Setup}
This subsection describes the datasets, baselines, and LLM configuration used in our experiments.
The hyperparameters are summarized in \Cref{tab:hyperparams}.

\paragraph{Dataset}
We use benchmark functions from RevLib, LGSynth91, and ISCAS85/89.
RevLib~\citep{wille2008revlib} provides benchmarks for reversible and quantum circuit design,\footnote{\url{https://www.revlib.org/index.php}}
while LGSynth91 and ISCAS85/89 are taken from the \texttt{hdl-benchmarks} repository.\footnote{\url{https://github.com/ispras/hdl-benchmarks}}
We augment RevLib functions using input negation, output complementation, and don't-care expansion. 
The resulting instances are split into a search set for evaluating generated heuristics during evolution and a validation set for selecting the best heuristic across independent runs.
Final evaluation uses a benchmark set of 148 reversible functions not used during search or validation, consisting of 96 RevLib, 48 LGSynth91, and 4 ISCAS85/89 circuits.
A detailed comparison between the search/validation set and the benchmark set is provided in \Cref{subsec:dataset-analysis}.

\paragraph{Baseline Algorithms}
We compare against five classical BDD variable ordering methods implemented in CUDD: sifting, symmetric sifting, group sifting, GA and SA (details in \Cref{app:cudd}).
We also compare against a learning-based method, BDD2Seq~\citep{miao2026bdd2seq}, using the QCC reported in the original paper; our benchmark set is constructed to match the functions evaluated therein.

\paragraph{LLM configuration}
The selected heuristic is obtained from an evolutionary run using the local LLM \texttt{gpt-oss-120b},\footnote{\url{https://huggingface.co/openai/gpt-oss-120b}} an open-source 120B-parameter mixture-of-experts model released by OpenAI under the Apache~2.0 license.
\Cref{app:ablation} compares independent runs with two model settings: the local model \texttt{gpt-oss-120b} and the commercial model \texttt{gpt-5.4-mini}. \footnote{\url{https://platform.openai.com/docs/models/gpt-5.4-mini}}

\subsection{Experimental Results}
\label{subsec:results}
We organize the experimental results around two main findings, followed by an ablation study of \texttt{QuantumEvo}.
Finding 1 examines whether BDD size is a reliable proxy for downstream QCC, and Finding 2 evaluates the discovered heuristic on the benchmark. 
We then present an ablation study of \texttt{QuantumEvo} to analyze the impact of key search design choices.

\subsubsection{Finding 1: BDD size is an imperfect proxy for downstream QCC.}
\begin{figure}[t]
\captionsetup[subfigure]{justification=centering}
\centering
\begin{subfigure}[t]{0.48\textwidth}
    \centering
    \includegraphics[width=\linewidth]{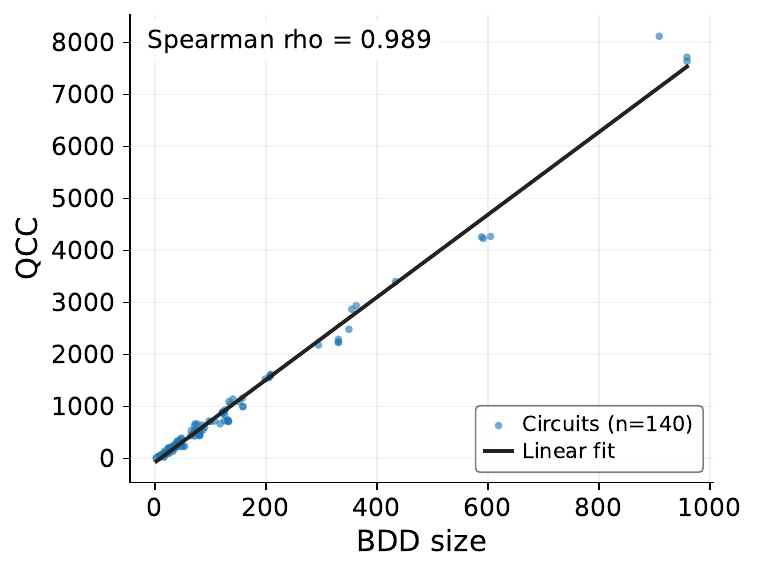}
    \caption{Overall trend with Spearman $\rho = 0.989$ and linear fit }
\end{subfigure}
\hfill
\begin{subfigure}[t]{0.48\textwidth}
    \centering
    \includegraphics[width=\linewidth]{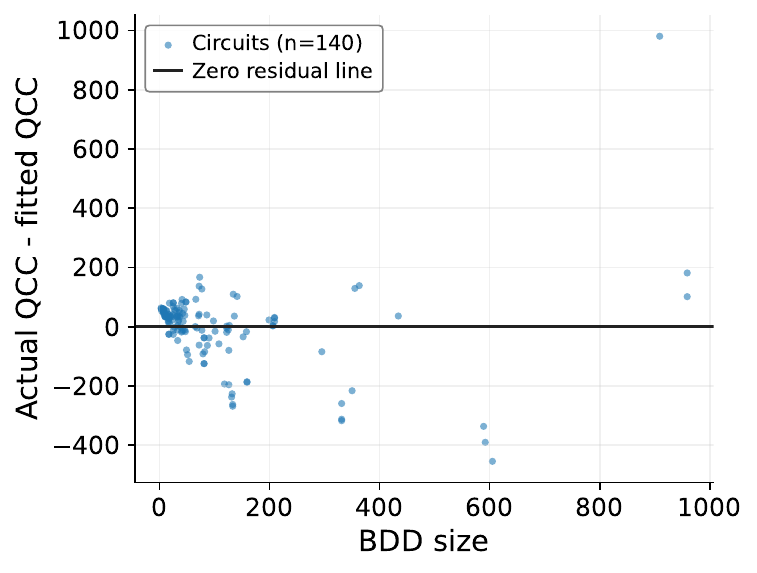}
    \caption{Residuals from the linear fit, showing function-level deviations}
\end{subfigure}
\caption{BDD size vs.\ QCC for the 140 search and validation functions with BDD size at most 1000. }
\label{fig:bdd-qc-residual}
\end{figure}

The BDD variable-ordering algorithms commonly aim to minimize BDD size.
Since our objective is to minimize downstream QCC, we first examine whether BDD size serves as a reliable proxy for QCC.
\Cref{fig:bdd-qc-residual} examines the relationship between BDD size and QCC on the 140 search/validation functions with BDD size at most 1000. 
The residual plot shows the difference between the actual and fitted QCC, with the horizontal line indicating zero residual.
QCC is strongly correlated with BDD size, as indicated by Spearman $\rho = 0.989$, but residuals around the fitted trend are substantial. The mean absolute deviation is 79.3, with individual deviations reaching 980.
Moreover, 35\% of these functions exhibit at least one ordering pair in which a smaller BDD yields a higher QCC, suggesting that BDD size is not a reliable proxy.
These observations motivate the QCC-aware heuristic design used in \texttt{QuantumEvo}.

\subsubsection{Finding 2: HGA-QE is competitive with existing variable ordering baselines.}
\begin{figure}[t]
\captionsetup[subfigure]{justification=centering}
\centering
\begin{subfigure}[t]{0.48\textwidth}
    \centering
    \includegraphics[width=\linewidth]{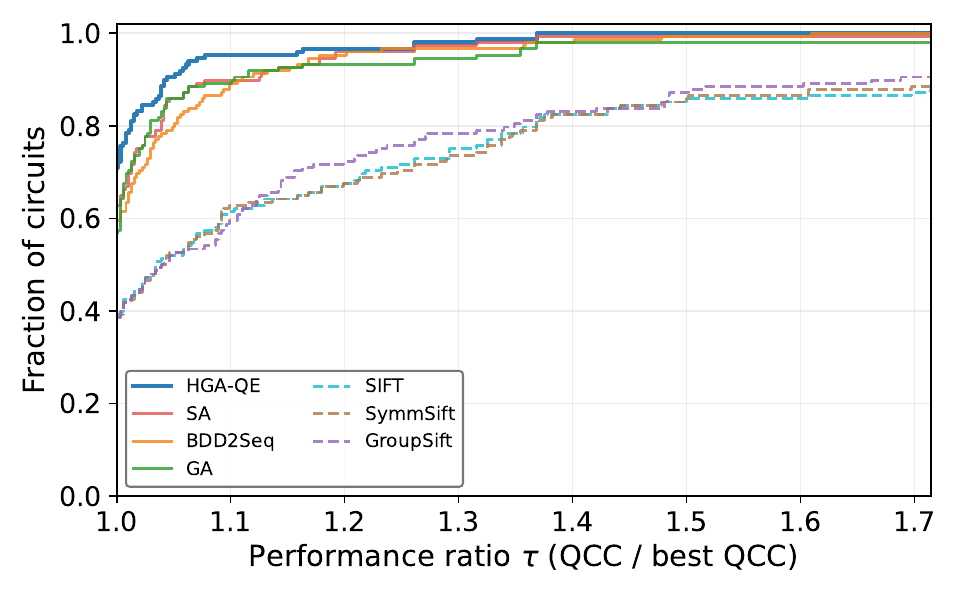}
    \caption{Performance profile}
    \label{fig:performance-a}
\end{subfigure}
\hfill
\begin{subfigure}[t]{0.48\textwidth}
    \centering
    \includegraphics[width=\linewidth]{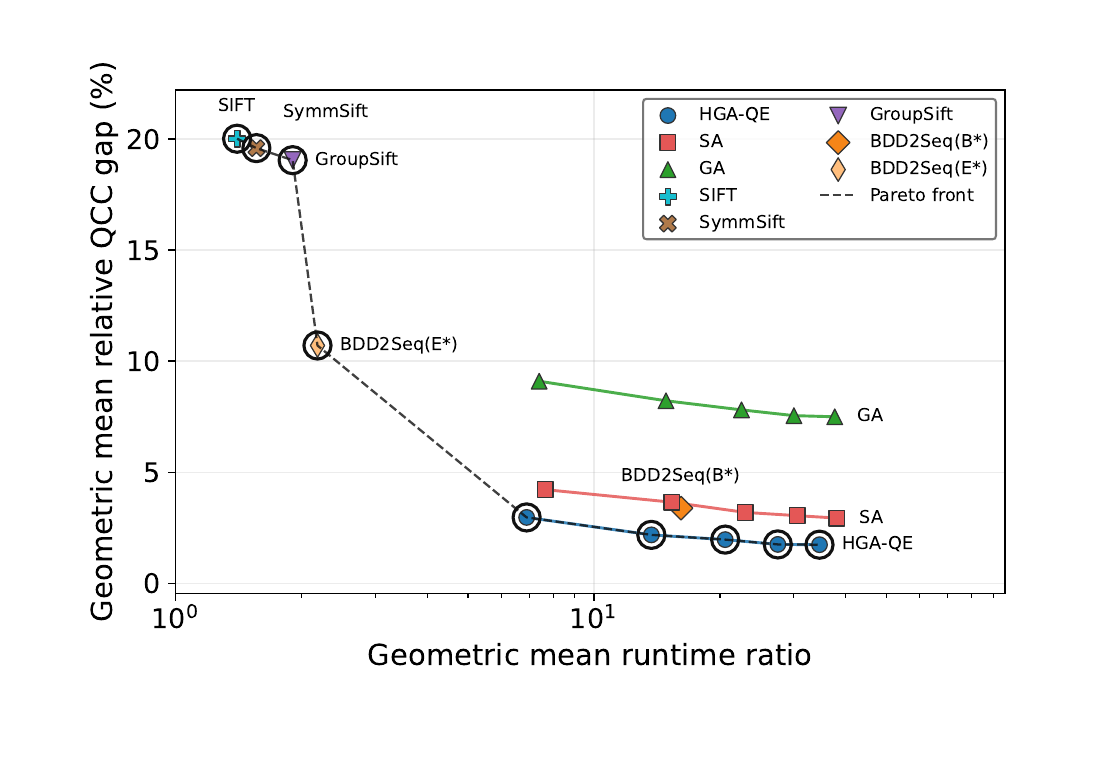}
    \caption{Quality--runtime Pareto curve}
    \label{fig:performance-b}
\end{subfigure}
\caption{(a) Performance profile over the whole benchmark set, using the ratio between each method's QCC and the best QCC achieved on the same function. (b) Quality--runtime Pareto curve over all 148 benchmark circuits. The $x$-axis reports the geometric mean runtime ratio relative to the per-circuit fastest result, and the $y$-axis reports the geometric mean relative QCC gap. Connected points correspond to cumulative run budgets for stochastic methods, whereas deterministic methods and BDD2Seq variants are shown as single points. Lower-left is better.}
\label{fig:performance-profile-qc}
\end{figure}

We select the heuristic with the best validation fitness across independent runs, corresponding to the one analyzed in \Cref{subsec:proposed_algorithm}. 
\Cref{tab:final-avg-results} reports results by suite, where Best counts wins or ties and Strict counts unique wins. 
For stochastic methods, QCC statistics are computed over five seeds, and for BDD2Seq we use the better of BDD2Seq(B*) and BDD2Seq(E*) reported in the original paper.
Overall, HGA-QE matches or improves upon the best baseline on 105 of 148 functions and is uniquely best on 20 functions. 
On RevLib, it achieves the highest Best count, although its mean QCC is comparable to SA and BDD2Seq has more strict wins. 
Its gains are clearer on LGSynth91 and ISCAS85/89, suggesting transfer beyond the augmented RevLib functions used for search and validation.
The performance profile in \Cref{fig:performance-a} shows that HGA-QE remains competitive across the benchmark set with respect to QCC.
The quality--runtime Pareto curve in \Cref{fig:performance-b}, aggregated using the geometric mean, summarizes the overall trade-off without being dominated by a few expensive instances.
After two runs, HGA-QE achieves a trade-off comparable to BDD2Seq, while additional runs further improve QCC at the cost of increased cumulative runtime. 
While the overall trade-off is favorable, reducing runtime on some expensive instances remains an important direction for further optimization.
\Cref{tab:final-individual-best} details the 20 functions on which our method achieves a lower QCC than all the baselines.
The average time denotes the mean runtime per seed, whereas the total time denotes the cumulative runtime across all seeds. 
Both measurements include BDD construction, variable ordering, and reversible circuit synthesis. 
Because these steps are performed once for a given Boolean function and the resulting circuit can be reused for subsequent executions, we prioritize circuit quality as long as the synthesis runtime remains practically manageable.

\begin{table}
\centering
\caption{Per-suite benchmark performance. Best includes ties for the lowest Best QCC; Strict is counted only when the method is the unique best (5 random seeds for stochastic methods).}
\label{tab:final-avg-results}
\centering
\small
\begin{tabular}{llrrrrr}
\toprule
Suite & Method & Best & Strict & Mean Best QCC & Mean Avg QCC & Mean Worst QCC \\
\midrule
RevLib (96) & HGA-QE & \textbf{70} & 6 & \textbf{1416.04} & \textbf{1434.31} & \textbf{1450.76} \\
 & SA & 68 & 5 & 1416.25 & 1437.27 & 1463.44 \\
 & BDD2Seq & 65 & \textbf{13} & 1456.03 & -- & -- \\
 & GA & 58 & 1 & 1445.28 & 1457.55 & 1468.35 \\
 & SIFT & 42 & 0 & 1767.24 & -- & -- \\
 & SYMM SIFT & 41 & 0 & 1757.90 & -- & -- \\
 & GROUP SIFT & 41 & 1 & 1688.56 & -- & -- \\
\midrule
LGSynth91 (48) & HGA-QE & \textbf{31} & \textbf{13} & \textbf{2088.54} & \textbf{2218.05} & \textbf{2344.25} \\
 & SA & 22 & 3 & 2153.58 & 2267.06 & 2447.79 \\
 & BDD2Seq & 22 & 5 & 2466.00 & -- & -- \\
 & GA & 23 & 6 & 125739.94 & 202778.95 & 293303.17 \\
 & SIFT & 16 & 1 & 4687.90 & -- & -- \\
 & SYMM SIFT & 16 & 0 & 4682.71 & -- & -- \\
 & GROUP SIFT & 15 & 0 & 5762.81 & -- & -- \\
\midrule
ISCAS85/89 (4) & HGA-QE & \textbf{4} & \textbf{1} & \textbf{12134.25} & \textbf{12211.23} & \textbf{12288.75} \\
 & SA & 3 & 0 & 12286.00 & 12403.25 & 12589.00 \\
 & BDD2Seq & 1 & 0 & 13549.50 & -- & -- \\
 & GA & 3 & 0 & 12296.00 & 12518.60 & 12681.25 \\
 & SIFT & 0 & 0 & 19230.00 & -- & -- \\
 & SYMM SIFT & 0 & 0 & 19230.00 & -- & -- \\
 & GROUP SIFT & 1 & 0 & 23486.00 & -- & -- \\
\bottomrule
\end{tabular}

\end{table}

\begin{table}
\centering
\caption{Compact comparison for functions where \texttt{QuantumEvo} is strictly best.}
\label{tab:final-individual-best}
\renewcommand{\arraystretch}{1.4}
\resizebox{\linewidth}{!}{%
\begin{tabular}{lrr|rrr|rrr|rrr}
\toprule
Circuit & I & O & \multicolumn{3}{c|}{QuantumEvo} & \multicolumn{3}{c|}{GA} & \multicolumn{3}{c}{SA} \\
\cmidrule(lr){4-6}\cmidrule(lr){7-9}\cmidrule(lr){10-12}
 & & & Best & Avg & Avg Time & Best & Avg & Avg Time & Best & Avg & Avg Time \\
\midrule
\texttt{apex5\_104}\textsuperscript{R} & 117 & 88 & \textbf{9184} & 9812.40 & 0.97 & 9551 & 10041.40 & 1.24 & 9397 & \textbf{9809.00} & 1.97 \\
\texttt{apex6\_orig}\textsuperscript{L} & 135 & 99 & \textbf{3736} & \textbf{3787.20} & 0.65 & 4159 & 4368.40 & 0.64 & 4392 & 4551.40 & 1.08 \\
\texttt{apex7\_orig}\textsuperscript{L} & 49 & 37 & \textbf{1431} & \textbf{1440.60} & 0.28 & 1441 & 1443.40 & 0.58 & 1703 & 1759.20 & 0.44 \\
\texttt{b9\_orig}\textsuperscript{L} & 41 & 21 & \textbf{713} & 719.40 & 0.17 & 715 & \textbf{719.00} & 0.27 & 804 & 804.00 & 0.19 \\
\texttt{c432}\textsuperscript{I} & 36 & 7 & \textbf{10494} & \textbf{11053.20} & 245.19 & 11141 & 11170.60 & 6.20 & 11101 & 11148.60 & 1.46 \\
\texttt{comp\_orig}\textsuperscript{L} & 32 & 3 & \textbf{796} & \textbf{796.00} & 2.41 & 824 & 824.00 & 2.28 & 897 & 936.00 & 2.01 \\
\texttt{cordic\_138}\textsuperscript{R} & 23 & 2 & \textbf{304} & \textbf{308.00} & 0.16 & 306 & 312.00 & 0.23 & 316 & 323.20 & 0.17 \\
\texttt{cordic\_orig}\textsuperscript{L} & 23 & 2 & \textbf{304} & \textbf{308.00} & 0.12 & 306 & 312.00 & 0.20 & 316 & 323.20 & 0.14 \\
\texttt{cu\_orig}\textsuperscript{L} & 14 & 11 & \textbf{219} & \textbf{219.80} & 0.08 & 224 & 224.00 & 0.09 & 220 & 220.00 & 0.09 \\
\texttt{dalu\_orig}\textsuperscript{L} & 75 & 16 & \textbf{5230} & 5301.60 & 71.73 & 5240 & \textbf{5291.20} & 71.72 & 5316 & 5663.20 & 69.72 \\
\texttt{frg1\_160}\textsuperscript{R} & 28 & 3 & \textbf{559} & \textbf{597.40} & 0.17 & 560 & 610.20 & 0.24 & 567 & 637.40 & 0.17 \\
\texttt{frg1\_orig}\textsuperscript{L} & 28 & 3 & \textbf{559} & \textbf{597.40} & 0.15 & 560 & 610.20 & 0.27 & 567 & 637.40 & 0.17 \\
\texttt{k2\_orig}\textsuperscript{L} & 45 & 45 & \textbf{10156} & \textbf{10212.60} & 0.97 & 10200 & 10265.40 & 1.78 & 10276 & 10410.00 & 1.49 \\
\texttt{misex1\_178}\textsuperscript{R} & 8 & 7 & \textbf{279} & 290.40 & 0.06 & 287 & \textbf{287.00} & 0.06 & 288 & 288.00 & 0.07 \\
\texttt{pair\_orig}\textsuperscript{L} & 173 & 137 & \textbf{20236} & \textbf{22344.20} & 6.24 & 47110 & 602976.40 & 8.28 & 20729 & 22438.00 & 20.16 \\
\texttt{rot\_orig}\textsuperscript{L} & 135 & 107 & \textbf{23926} & 26838.00 & 845.19 & 5928978 & 9069721.00 & 62.87 & 24124 & \textbf{25464.60} & 25.99 \\
\texttt{seq\_201}\textsuperscript{R} & 41 & 35 & \textbf{9080} & 9151.00 & 1.17 & 9113 & \textbf{9142.00} & 3.03 & 9165 & 9237.40 & 1.99 \\
\texttt{urf4\_89}\textsuperscript{R} & 11 & 11 & \textbf{28406} & \textbf{28537.40} & 0.41 & 28538 & 28591.80 & 0.54 & 28489 & 28590.80 & 0.97 \\
\texttt{x3\_orig}\textsuperscript{L} & 135 & 99 & \textbf{3736} & \textbf{3787.20} & 0.62 & 4159 & 4368.40 & 0.66 & 4392 & 4551.40 & 1.15 \\
\texttt{x4\_orig}\textsuperscript{L} & 94 & 71 & \textbf{2437} & \textbf{2513.60} & 0.37 & 2505 & 2809.40 & 0.65 & 2738 & 2863.20 & 0.69 \\
\bottomrule
\end{tabular}
}

\bigskip

\resizebox{\linewidth}{!}{%
\begin{tabular}{l|rr|rr|rr|rr|rr|rr}
\toprule
Circuit & \multicolumn{2}{|c|}{QuantumEvo} & \multicolumn{2}{c|}{BDD2Seq(B*)} & \multicolumn{2}{c|}{BDD2Seq(E*)} & \multicolumn{2}{c|}{SIFT} & \multicolumn{2}{c|}{SYMM\_SIFT} & \multicolumn{2}{c}{GROUP\_SIFT} \\
\cmidrule(lr){2-3}\cmidrule(lr){4-5}\cmidrule(lr){6-7}\cmidrule(lr){8-9}\cmidrule(lr){10-11}\cmidrule(lr){12-13}
 & Best & Total Time & Best & Time & Best & Time & Best & Time & Best & Time & Best & Time \\
\midrule
\texttt{apex5\_104}\textsuperscript{R} & \textbf{9184} & 4.87 & 9852 & 12.83 & 10697 & 0.20 & 10358 & 0.06 & 10228 & 0.03 & 10227 & 0.06 \\
\texttt{apex6\_orig}\textsuperscript{L} & \textbf{3736} & 3.24 & 3895 & 16.72 & 4021 & 0.16 & 4822 & 0.05 & 4810 & 0.03 & 4269 & 0.03 \\
\texttt{apex7\_orig}\textsuperscript{L} & \textbf{1431} & 1.40 & 1455 & 2.40 & 1456 & 0.07 & 2653 & 0.04 & 2653 & 0.04 & 2377 & 0.04 \\
\texttt{b9\_orig}\textsuperscript{L} & \textbf{713} & 0.83 & 737 & 1.71 & 737 & 0.05 & 804 & 0.03 & 810 & 0.03 & 836 & 0.03 \\
\texttt{c432}\textsuperscript{I} & \textbf{10494} & 1225.95 & 12255 & 10.01 & 20391 & 6.79 & 11101 & 0.04 & 11101 & 0.02 & 11141 & 0.04 \\
\texttt{comp\_orig}\textsuperscript{L} & \textbf{796} & 12.04 & 824 & 40.21 & 824 & 28.75 & 961 & 1.77 & 1071 & 1.96 & 884 & 1.79 \\
\texttt{cordic\_138}\textsuperscript{R} & \textbf{304} & 0.82 & 344 & 0.62 & 446 & 0.05 & 325 & 0.02 & 323 & 0.02 & 318 & 0.04 \\
\texttt{cordic\_orig}\textsuperscript{L} & \textbf{304} & 0.58 & 320 & 0.62 & 320 & 0.06 & 325 & 0.01 & 323 & 0.01 & 318 & 0.03 \\
\texttt{cu\_orig}\textsuperscript{L} & \textbf{219} & 0.39 & 224 & 0.26 & 231 & 0.02 & 220 & 0.01 & 220 & 0.01 & 224 & 0.03 \\
\texttt{dalu\_orig}\textsuperscript{L} & \textbf{5230} & 358.67 & 5297 & 55.53 & 5652 & 28.17 & 31145 & 66.13 & 31145 & 71.95 & 47170 & 75.07 \\
\texttt{frg1\_160}\textsuperscript{R} & \textbf{559} & 0.85 & 598 & 0.95 & 629 & 0.03 & 747 & 0.03 & 747 & 0.03 & 827 & 0.03 \\
\texttt{frg1\_orig}\textsuperscript{L} & \textbf{559} & 0.77 & 757 & 0.95 & 762 & 0.03 & 747 & 0.03 & 747 & 0.02 & 827 & 0.02 \\
\texttt{k2\_orig}\textsuperscript{L} & \textbf{10156} & 4.85 & 10432 & 2.38 & 10686 & 0.19 & 11058 & 0.07 & 11058 & 0.13 & 11275 & 0.09 \\
\texttt{misex1\_178}\textsuperscript{R} & \textbf{279} & 0.30 & 281 & 0.12 & 289 & 0.02 & 288 & 0.01 & 288 & 0.03 & 289 & 0.03 \\
\texttt{pair\_orig}\textsuperscript{L} & \textbf{20236} & 31.21 & 20754 & 28.19 & 20799 & 1.06 & 46491 & 0.11 & 46391 & 0.17 & 49917 & 0.15 \\
\texttt{rot\_orig}\textsuperscript{L} & \textbf{23926} & 4225.96 & 38453 & 37.98 & 48344 & 15.69 & 78639 & 0.56 & 78374 & 0.62 & 110936 & 0.74 \\
\texttt{seq\_201}\textsuperscript{R} & \textbf{9080} & 5.85 & 9349 & 2.12 & 9908 & 0.33 & 19362 & 0.24 & 19350 & 0.28 & 15309 & 0.27 \\
\texttt{urf4\_89}\textsuperscript{R} & \textbf{28406} & 2.06 & 28488 & 0.68 & 28833 & 0.47 & 28523 & 0.06 & 28523 & 0.10 & 28523 & 0.11 \\
\texttt{x3\_orig}\textsuperscript{L} & \textbf{3736} & 3.09 & 3941 & 16.78 & 4645 & 0.19 & 4822 & 0.03 & 4810 & 0.05 & 4269 & 0.03 \\
\texttt{x4\_orig}\textsuperscript{L} & \textbf{2437} & 1.85 & 2662 & 8.31 & 2782 & 0.11 & 4470 & 0.02 & 4470 & 0.04 & 4334 & 0.04 \\
\bottomrule
\end{tabular}
}
\par\smallskip{\footnotesize\raggedright 
\textit{Superscripts denote source dataset.} \textsuperscript{I}~ISCAS85/89; \textsuperscript{L}~LGSynth91; \textsuperscript{R}~RevLib. 
\textit{Hardware normalization.} BDD2Seq(B*) runtime is normalized from the original dual Xeon~8375C platform to our single-thread Ryzen~9~5900X platform. Because BDD2Seq(B*) uses beam search with a beam width of 20 and synthesizes all 20 candidate orderings, we normalize its additional CPU-based synthesis cost using the PassMark single-thread ratio between the two platforms. BDD2Seq(E*) is left unchanged because its single synthesis call constitutes only a small fraction of its total runtime. The normalization assumptions and derivation are provided in Appendix~\ref{app:full-results-per-instance}.}

\end{table}

\subsubsection{Ablation Study of \texttt{QuantumEvo}}
\label{subsec:ablation}
\begin{figure}[t]
\captionsetup[subfigure]{justification=centering}
\begin{subfigure}[t]{0.49\textwidth}
    \centering
    \includegraphics[width=\linewidth]{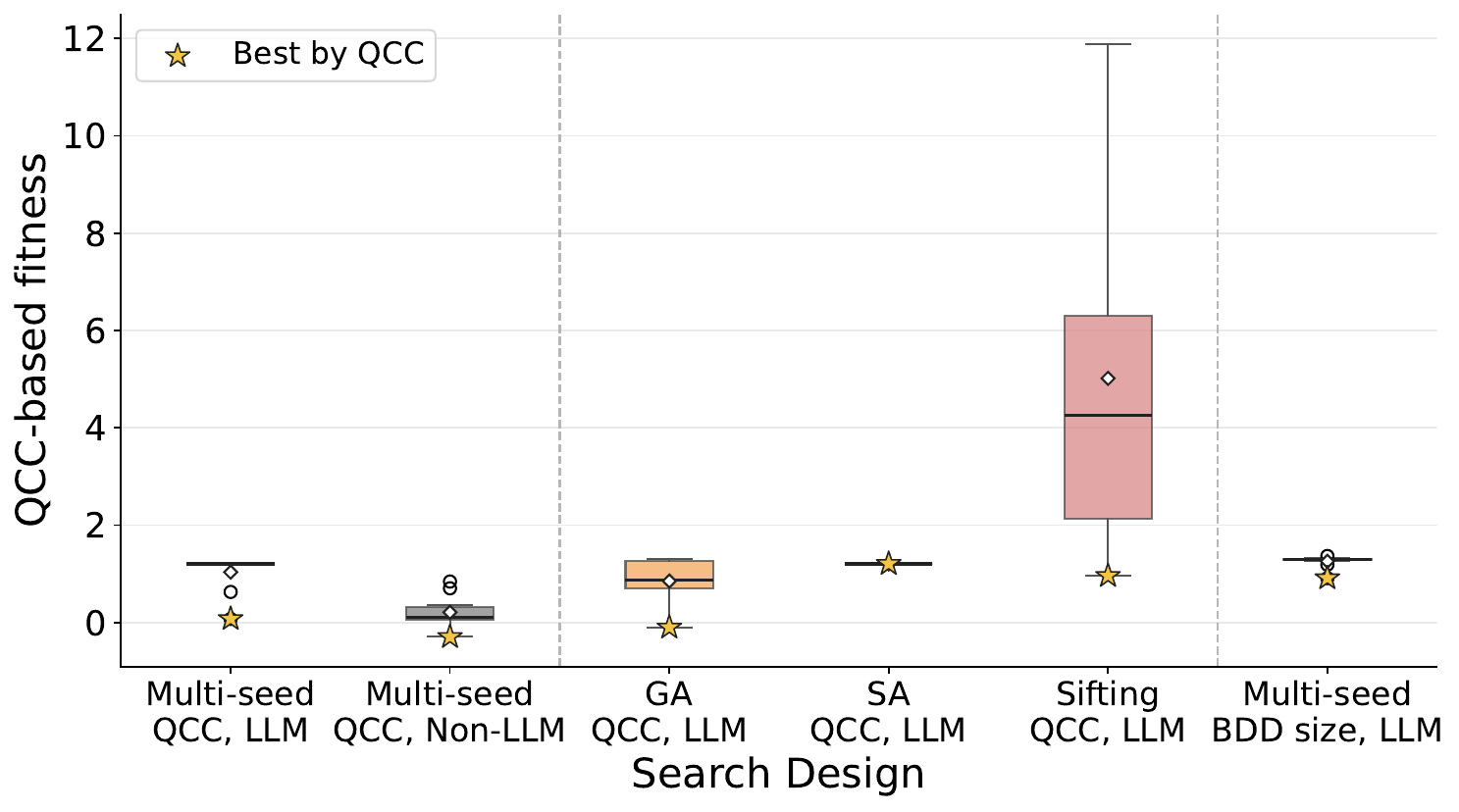}
    \caption{QCC fitness (lower is better, $\downarrow$)}
\end{subfigure}
\hfill
\begin{subfigure}[t]{0.49\textwidth}
    \centering
    \includegraphics[width=\linewidth]{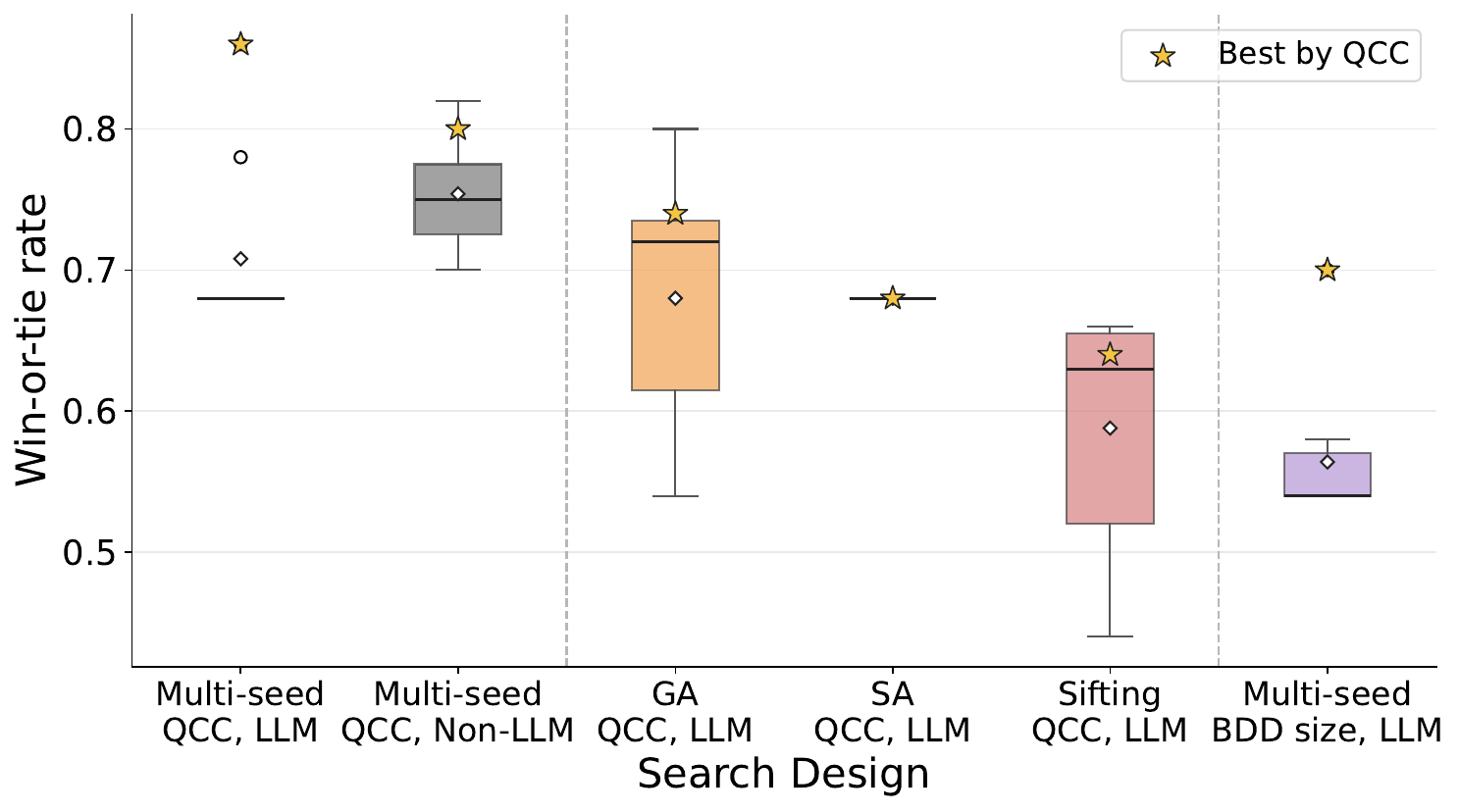}
    \caption{Win-or-tie rate (higher is better, $\uparrow$)}
\end{subfigure}
\caption{Validation-set distributions over ten independent runs per setting. The star marks the run selected by the lowest validation QCC fitness; the corresponding win-or-tie rate is reported for the same run.}
\label{fig:ablation-val-boxplot}
\end{figure}

We study three components of \texttt{QuantumEvo}: LLM guidance, multi-family initialization, and QCC-based fitness.
For each setting, we conduct ten independent searches and compare their validation performance in \Cref{fig:ablation-val-boxplot}.
We then select the individual with the best validation fitness from each setting and evaluate it on the full 148-function benchmark set.
The corresponding benchmark results are summarized in \Cref{tab:ablation-main}.

\paragraph{LLM guidance.}
To isolate the contribution of the LLM, we construct a non-LLM heuristic-generation procedure that retains the same evolutionary procedure, fitness function, initialization, and search and validation sets as \texttt{QuantumEvo}.
Instead of generating and revising code with an LLM, it assembles candidates from predefined components of sifting, GA, and SA, as detailed in \Cref{app:nonllm-baseline}.
As shown in \Cref{fig:ablation-val-boxplot}, the non-LLM search achieves stronger validation fitness, and all ten runs converge to GA-family individuals.
These results do not establish that LLM guidance is uniformly superior, but suggest that LLM-driven algorithm design beyond a predefined component space can yield heuristics that transfer more effectively to unseen functions.

\paragraph{Multi-family initialization.}
To assess seed diversity, we compare the multi-family initialization of sifting, GA, and SA against initialization from each family alone.
SA-only initialization has zero variance, with every run converging to the same individual despite LLM stochasticity, indicating that a single seed family can constrain the explored region.
GA-only initialization achieves better validation QCC fitness than multi-family initialization, whereas multi-family initialization attains a higher validation win-or-tie rate.
On the full benchmark, however, the heuristic selected under GA-only initialization no longer leads, and the heuristics selected under SA-only and sifting-only initialization also underperform HGA-QE.

\paragraph{QCC-based fitness.}
Finally, we replace the downstream QCC objective with BDD size while retaining the remaining components of \texttt{QuantumEvo}.
The BDD-size setting shows a less favorable median validation QCC fitness in \Cref{fig:ablation-val-boxplot}.
Its selected heuristic also underperforms the corresponding QCC-based setting on the benchmark, with lower Best and Strictly Best rates and higher mean QCC.

\begin{table}
\centering
\caption{Benchmark evaluation of the best-validation individual from each setting on the full benchmark set with five random seeds. Starting from the full \texttt{QuantumEvo} setting (bold), each block varies one factor: Guidance, Initialization, or Fitness.}
\label{tab:ablation-main}
\small
\setlength{\tabcolsep}{4pt}
\begin{tabular}{lllccrrr}
\toprule
Initialization & Fitness & Guidance & \makecell{Best/Total\\(\%)} & \makecell{Strictly Best/Total\\(\%)} & \makecell{Mean\\Best QCC} & \makecell{Mean\\Avg QCC} & \makecell{Mean\\Worst QCC} \\
\midrule
\multirow{2}{*}{Sifting, GA, SA} & \multirow{2}{*}{QCC} & LLM & \textbf{70.9} & \textbf{13.5} & \textbf{1923.83} & \textbf{1979.76} & \textbf{2033.46} \\
 & & Non-LLM & 66.9 & 5.4 & 2075.15 & 2220.27 & 2432.07 \\
\midrule
GA & QCC & LLM & 63.5 & 3.4 & 2061.84 & 2172.50 & 2264.26 \\
SA & QCC & LLM & 66.2 & 0.7 & 1949.11 & 2004.30 & 2084.68 \\
Sifting & QCC & LLM & 55.4 & 4.1 & 2203.34 & 2435.04 & 2670.19 \\
\midrule
Sifting, GA, SA & BDD size & LLM & 61.5 & 5.4 & 1944.65 & 2003.03 & 2099.68 \\
\bottomrule
\end{tabular}
\end{table}

\section{Discussion}
\label{sec:discussion}
While the results demonstrate that LLM-driven evolutionary search can discover competitive BDD variable ordering heuristics, several limitations remain.
First, the discovered heuristic is still influenced by the initial heuristic families used to seed the search.
Although the best lineage transitions from SA-family to GA-family individuals, the best heuristic combines and refines ideas related to sifting, SA, and GA rather than introducing an entirely new class of ordering strategy.
Whether alternative framework designs can discover substantially different and more effective ordering strategies remains an open question.
Second, \texttt{QuantumEvo} optimizes an important component of BDD-based reversible synthesis, namely the variable ordering.
While this choice affects the BDD structure and downstream QCC, other resource measures, such as ancilla count, also depend on the synthesis procedure itself.
Future work could combine \texttt{QuantumEvo} with post-synthesis ancilla reduction techniques or alternative synthesis methods that produce lower-cost circuits.
Third, the non-LLM search remains competitive while avoiding the token and inference costs of LLM guidance. This suggests that hybrid strategies could reduce offline discovery cost by relying primarily on predefined evolutionary operators and invoking the LLM selectively when additional algorithmic variation is needed.

\section{Conclusion}
We presented \texttt{QuantumEvo}, an LLM-driven evolutionary framework that discovers interpretable and reusable BDD variable ordering heuristics for reversible circuit synthesis.
The best discovered heuristic, HGA-QE, achieves competitive QCC across diverse benchmark circuits and improves over classical CUDD ordering methods and learning-based baselines on a substantial subset of functions. 
Its implementation as a general CUDD variable-ordering heuristic also enables its reuse in other BDD-based workflows.
More broadly, \texttt{QuantumEvo} illustrates how LLM-based heuristic design can be adapted to domain-specific settings by evaluating candidates directly on the relevant downstream objective. 
This approach may extend to other optimization decisions embedded within specialized computational workflows.

\subsubsection*{Acknowledgments}
This work was supported by the National Research Foundation of Korea (NRF) grant funded by the Korean government (MSIT) (RS-2023-00259550).

\bibliographystyle{tmlr}
\bibliography{reference}

\appendix
\section{Example of Reversible Circuit Synthesis}

\subsection{NCT library and Circuit}
\label{app:circuit-example}
\begin{figure}[h]
\captionsetup[subfigure]{justification=centering}
\centering
\begin{subfigure}[b]{0.48\textwidth}
    \centering
    \small
    \begin{tabular}{ccc||ccc}
        \hline
        $x_1$ & $x_2$ & $x_3$ & $y_1$ & $y_2$ & $y_3$ \\
        \hline
        0 & 0 & 0 & 0 & 0 & 1 \\
        0 & 0 & 1 & 0 & 0 & 0 \\
        0 & 1 & 0 & 0 & 1 & 1 \\
        0 & 1 & 1 & 0 & 1 & 0 \\
        1 & 0 & 0 & 1 & 0 & 1 \\
        1 & 0 & 1 & 1 & 1 & 1 \\
        1 & 1 & 0 & 1 & 0 & 0 \\
        1 & 1 & 1 & 1 & 1 & 0 \\
        \hline
    \end{tabular}
    \caption{Truth table of a 3-bit reversible function}
\end{subfigure}
\hfill
\begin{subfigure}[b]{0.48\textwidth}
    \centering
    \includegraphics[width=\textwidth]{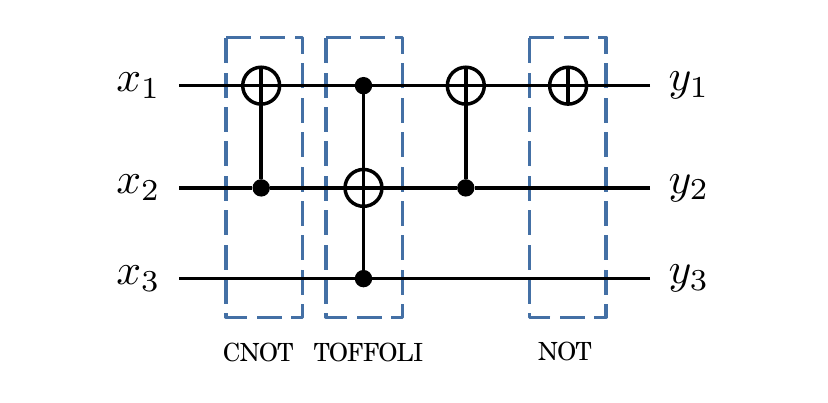}
    \caption{Synthesized reversible circuit (QCC = 8)}
\end{subfigure}
\caption{A 3-bit reversible function and its synthesized NCT circuit.}
\label{fig:example}
\end{figure}

The NCT library consists of three gate types: NOT, CNOT, and Toffoli.
The NOT gate flips a single bit ($x \mapsto \bar{x}$).
The CNOT gate flips the target bit when the control bit is 1, $(x, t) \mapsto (x, t \oplus x)$, and the Toffoli gate flips the target bit when both control bits are 1, $(x_1, x_2, t) \mapsto (x_1, x_2, t \oplus (x_1 \land x_2))$.
\Cref{fig:example} illustrates this with a 3-bit reversible function.
The truth table lists all $2^3 = 8$ input combinations $(x_1, x_2, x_3)$ and their outputs $(y_1, y_2, y_3)$; each output row appears exactly once, confirming bijectivity.
The right panel shows the NCT circuit synthesized from this function, with a total QCC of 8 under the cost model of \citet{szyprowski2013low}.

\subsection{BDD Variable Ordering and Circuit}
\begin{figure}[t]
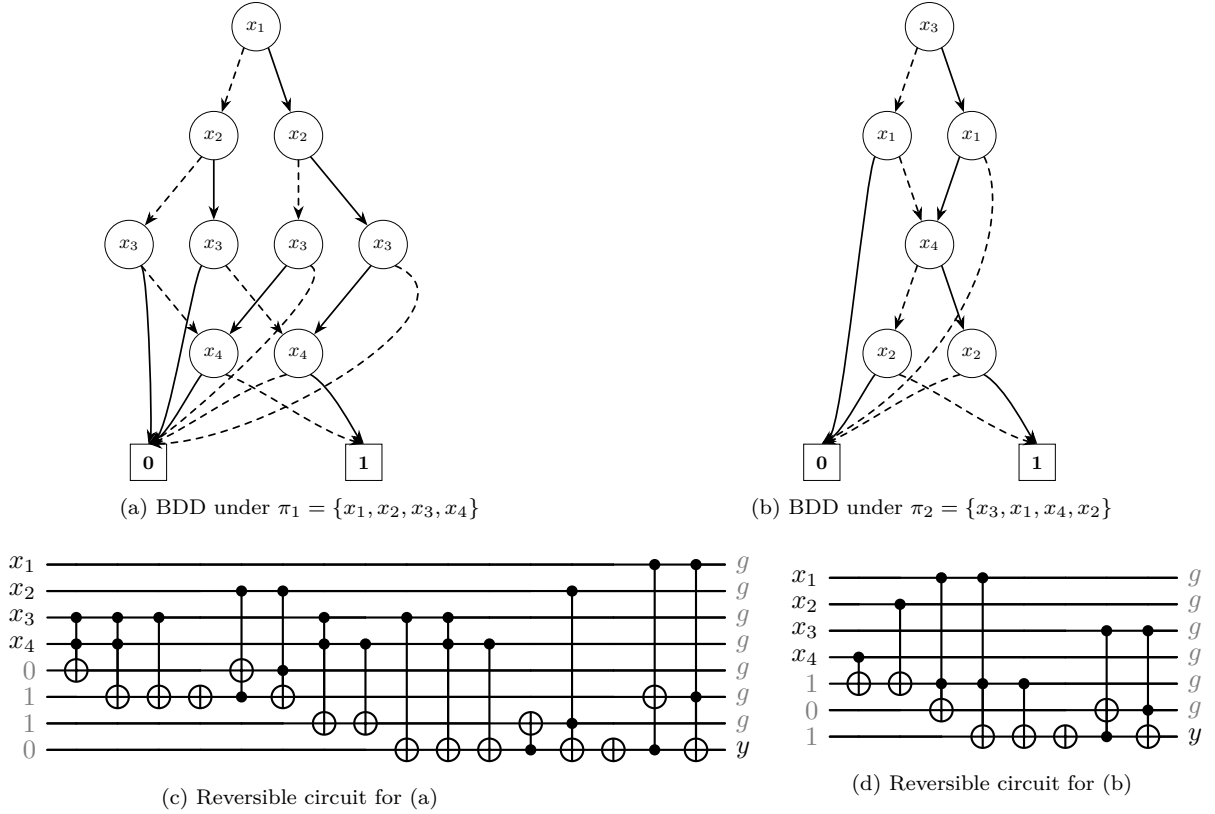

\captionsetup[subfigure]{justification=centering}
    \centering
    \begin{subfigure}[t]{0.5\textwidth}
        \centering
        \scalebox{0.8}{
            \begin{tikzpicture}[
    bdd node/.style={circle, draw, minimum size=8mm, inner sep=0pt, font=\small\bfseries},
    terminal/.style={rectangle, draw, minimum size=6mm, font=\small\bfseries},
    one edge/.style={-{Stealth}, thick, line cap=round, line join=round},
    zero edge/.style={-{Stealth}, dashed, thick, line cap=round, line join=round}
]

    \node[bdd node] (n8) at (0.00,0.00) {$x_1$};
    \node[bdd node] (n4) at (-0.70,-1.80) {$x_2$};
    \node[bdd node] (n7) at (0.70,-1.80) {$x_2$};
    \node[bdd node] (n1) at (-2.10,-3.60) {$x_3$};
    \node[bdd node] (n3) at (-0.70,-3.60) {$x_3$};
    \node[bdd node] (n5) at (0.70,-3.60) {$x_3$};
    \node[bdd node] (n6) at (2.10,-3.60) {$x_3$};
    \node[bdd node] (n0) at (-0.70,-5.40) {$x_4$};
    \node[bdd node] (n2) at (0.70,-5.40) {$x_4$};

    \node[terminal] (const0) at (-1.78,-7.20) {0};
    \node[terminal] (const1) at (1.78,-7.20) {1};

    \draw[zero edge] (n0.300) .. controls (0.07,-5.95) and (1.03,-6.70) .. (const1.north);
    \draw[one edge] (n0.240) .. controls (-1.06,-5.95) and (-1.43,-6.70) .. (const0.north);
    \draw[zero edge] (n1.300) -- (n0);
    \draw[one edge] (n1.300) .. controls (-1.77,-4.19) and (-1.72,-6.66) .. (const0.north);
    \draw[zero edge] (n2.240) .. controls (-0.25,-5.95) and (-1.21,-6.70) .. (const0.north);
    \draw[one edge] (n2.300) .. controls (1.25,-5.95) and (1.62,-6.70) .. (const1.north);
    \draw[zero edge] (n3.300) -- (n2);
    \draw[one edge] (n3.240) .. controls (-1.06,-4.19) and (-1.43,-6.66) .. (const0.north);
    \draw[zero edge] (n4.240) -- (n1);
    \draw[one edge] (n4.270) -- (n3);
    \draw[zero edge] (n5.300) .. controls (1.42,-4.26) and (-1.32,-6.66) .. (const0.north);
    \draw[one edge] (n5.240) -- (n0);
    \draw[zero edge] (n6.300) .. controls (3.86,-4.62) and (0.12,-6.88) .. (const0.north);
    \draw[one edge] (n6.240) -- (n2);
    \draw[zero edge] (n7.270) -- (n5);
    \draw[one edge] (n7.300) -- (n6);
    \draw[zero edge] (n8.240) -- (n4);
    \draw[one edge] (n8.300) -- (n7);

\end{tikzpicture}
        }
        \caption{BDD under $\pi_1 = \{x_1, x_2, x_3, x_4\}$}
    \end{subfigure}
    \hfill
    \begin{subfigure}[t]{0.45\textwidth}
        \centering
        \scalebox{0.8}{
            \begin{tikzpicture}[
    bdd node/.style={circle, draw, minimum size=8mm, inner sep=0pt, font=\small\bfseries},
    terminal/.style={rectangle, draw, minimum size=6mm, font=\small\bfseries},
    one edge/.style={-{Stealth}, thick, line cap=round, line join=round},
    zero edge/.style={-{Stealth}, dashed, thick, line cap=round, line join=round}
]

    \node[bdd node] (n5) at (0.00,0.00) {$x_3$};
    \node[bdd node] (n3) at (-0.70,-1.80) {$x_1$};
    \node[bdd node] (n4) at (0.70,-1.80) {$x_1$};
    \node[bdd node] (n2) at (0.00,-3.60) {$x_4$};
    \node[bdd node] (n0) at (-0.70,-5.40) {$x_2$};
    \node[bdd node] (n1) at (0.70,-5.40) {$x_2$};

    \node[terminal] (const0) at (-1.78,-7.20) {0};
    \node[terminal] (const1) at (1.78,-7.20) {1};

    \draw[zero edge] (n0.300) .. controls (0.07,-5.95) and (1.03,-6.70) .. (const1.north);
    \draw[one edge] (n0.240) .. controls (-1.06,-5.95) and (-1.43,-6.70) .. (const0.north);
    \draw[zero edge] (n1.240) .. controls (-0.25,-5.95) and (-1.21,-6.70) .. (const0.north);
    \draw[one edge] (n1.300) .. controls (1.25,-5.95) and (1.62,-6.70) .. (const1.north);
    \draw[zero edge] (n2.240) -- (n0);
    \draw[one edge] (n2.300) -- (n1);
    \draw[zero edge] (n3.300) -- (n2);
    \draw[one edge] (n3.240) .. controls (-1.06,-2.43) and (-1.43,-6.62) .. (const0.north);
    \draw[zero edge] (n4.300) .. controls (1.34,-3.02) and (0.82,-5.42) .. (const0.north);
    \draw[one edge] (n4.240) -- (n2);
    \draw[zero edge] (n5.240) -- (n3);
    \draw[one edge] (n5.300) -- (n4);

\end{tikzpicture}
        }
        \caption{BDD under $\pi_2 = \{x_3, x_1, x_4, x_2\}$}
    \end{subfigure}

    \vspace{0.3cm}
    \begin{subfigure}[t]{0.5\textwidth}
        \input{fig/circuit_none_2}
        \caption{Reversible circuit for (a)}
    \end{subfigure}
    \hfill
    \begin{subfigure}[t]{0.35\textwidth}
        \input{fig/circuit_sift_2}
        \caption{Reversible circuit for (b)}
    \end{subfigure}
\caption{Impact of variable ordering on QCC for the function \texttt{4mod5\_8}. Solid/dashed edges denote 1/0-branches. The BDD sizes are 9 under $\pi_1$ and 6 under $\pi_2$, whereas the corresponding QCC values are 52 and 24. Here, $y = \overline{(x_1 \oplus x_3)} \cdot \overline{(x_2 \oplus x_4)}$}.
\label{fig:bdd_ordering2}
\end{figure}

\Cref{fig:bdd_ordering2} provides another example illustrating the impact of variable ordering on BDD size and QCC. 
The same function under two orderings yields BDD sizes of 9 and 6, and QCC of 52 and 24, respectively.
This shows that variable ordering can affect both BDD size and QCC.

\section{Detailed Experimental Setup}
\label{app:setup}
This section provides supplementary details needed to reproduce and interpret the experiments.

\subsection{Baseline Ordering Methods in CUDD}
\label{app:cudd}
CUDD~\citep{somenzi2001cudd} is a \texttt{C} library for constructing and manipulating BDDs.
The classical baselines implemented in CUDD are:
\begin{itemize}[nosep]
    \item \textbf{Sifting}~\citep{rudell1993dynamic}: Moves each variable through all possible levels and retains the position minimizing node count. %
    \item \textbf{Symmetric Sifting}~\citep{panda1994symmetry}: Extends sifting by detecting symmetric variable pairs during traversal and grouping them for joint sifting.
    \item \textbf{Group Sifting}~\citep{panda1995variables}: Generalizes symmetric sifting to arbitrary dependent variable groups, sifting each group as a whole.
    \item \textbf{Genetic Algorithm}~\citep{drechsler1996genetic}: Maintains a population of variable orderings; generates new individuals via crossover and sifts each before fitness evaluation, replacing the worst individual if the offspring improves.
    \item \textbf{Simulated Annealing}~\citep{bollig1995simulated}: Randomly perturbs the ordering and accepts or rejects moves according to an annealing schedule.
\end{itemize}

\subsection{Experimental Setting}
\label{app:setting}
\paragraph{Hyperparameters}
\Cref{tab:hyperparams} summarizes the hyperparameters used in all runs. 

\begin{table}[t]
\centering
\caption{Hyperparameters used for all runs.}
\label{tab:hyperparams}
\small
\begin{tabular}{@{}lc|lc@{}}
\toprule
Hyperparameter & Value & Hyperparameter & Value \\
\midrule
Max.\ function evaluations & 200 & Per-iteration timeout & 40 s \\
Population size & 10 & Evaluation seeds per instance & 3 \\
Initial population size & 30 & Search instances & 100 \\
Mutation rate & 0.5 & Validation instances & 50 \\ 
\bottomrule
\end{tabular}
\end{table}

\paragraph{Individual Representation}
Each individual is a program linked directly to CUDD.
Rather than implementing individuals at the Python level, we use functions linked into the BDD process, which avoids per-step interface overhead and provides more direct access to CUDD ordering routines and relevant internal data structures.

\paragraph{Initial Heuristics}
The initial heuristics are generated from three functions corresponding to sifting, GA, and SA.
The sifting and simulated annealing seeds follow the corresponding CUDD implementations, while the GA seed is a lightweight variant of GA in CUDD that only reduces the population size to keep evaluation cost manageable.

\subsection{Analysis of Dataset and Benchmarks}
\label{subsec:dataset-analysis}
\begin{figure}[h]
\captionsetup[subfigure]{justification=centering}
\centering
\begin{subfigure}{0.32\textwidth}
    \centering
    \includegraphics[width=\linewidth]{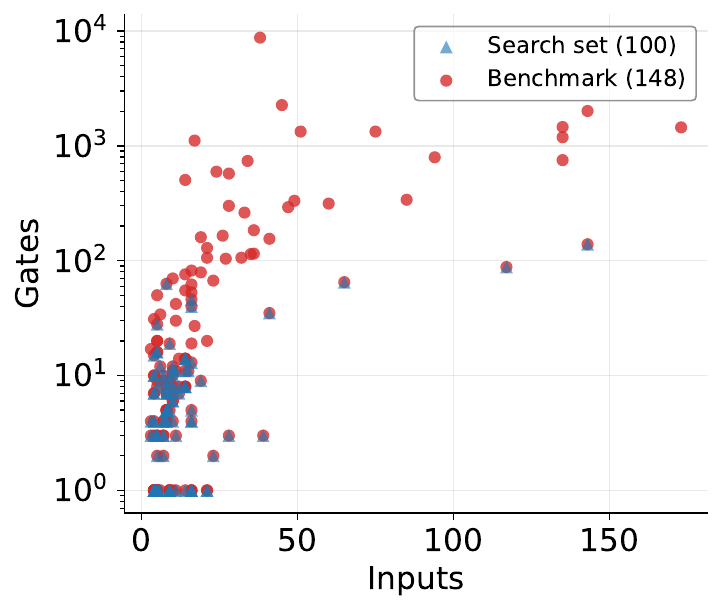}
    \caption{Input-gate coverage}
\end{subfigure}
\hfill
\begin{subfigure}{0.32\textwidth}
    \centering
    \includegraphics[width=\linewidth]{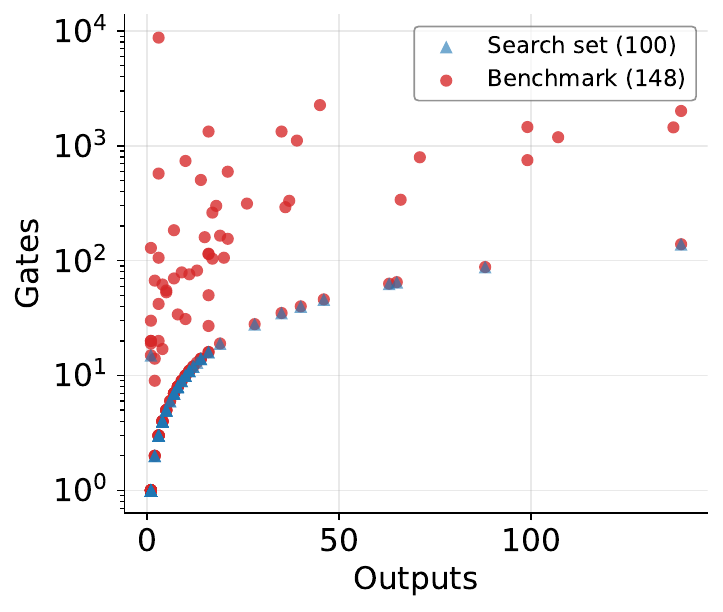}
    \caption{Output-gate coverage}
\end{subfigure}
\hfill
\begin{subfigure}{0.32\textwidth}
    \centering
    \includegraphics[width=\linewidth]{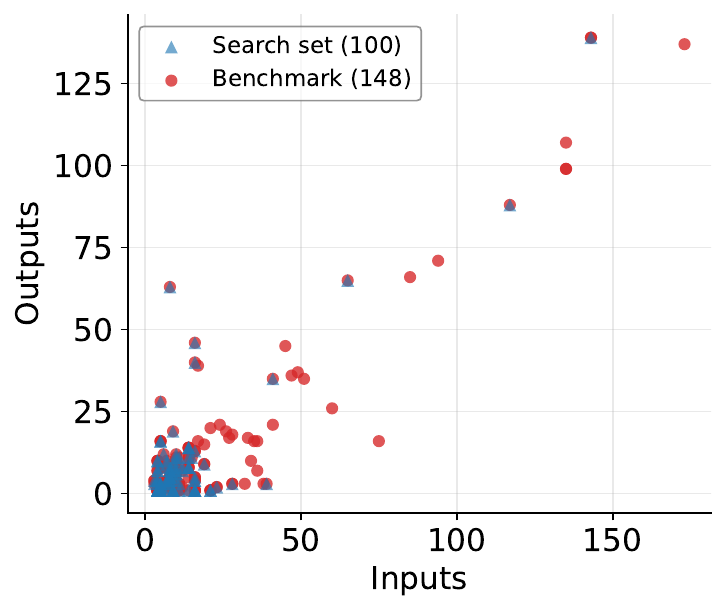}
    \caption{Input-output coverage}
\end{subfigure}
\caption{Structural comparison of the search set (100 functions) and the benchmark set (148 functions)}
\label{fig:dataset-benchmark-analysis}
\end{figure}

RevLib is aligned with reversible circuit synthesis and provides functions that can be repeatedly evaluated within the limited search budget.
To increase the diversity of the search pool, we augment RevLib.
\Cref{fig:dataset-benchmark-analysis} compares the 100 search functions with the 148 benchmark functions by input count, output count, and gate count.
The search set is drawn from augmented RevLib functions, whereas the benchmark set consists of RevLib, LGSynth91, and ISCAS85/89 functions with broader structural variety.
The benchmark points that coincide with the search-set points correspond to RevLib benchmark functions.
This overlap is only at the metadata level, and the search and benchmark sets share no Boolean functions.
By contrast, the benchmark points that do not coincide with any search-set point correspond to the non-RevLib suites, LGSynth91 and ISCAS85/89.
Thus, the non-RevLib suites introduce benchmark functions with input, output, and gate-count configurations not observed in the RevLib-based search set.

\section{Additional Analyses of \texttt{QuantumEvo}}
\subsection{Details of the Non-LLM Evolutionary Search}
\label{app:nonllm-baseline}
We implement the non-LLM control as an evolutionary search that constructs candidates from predefined algorithmic components, following the generation hyper-heuristic paradigm~\citep{burke2009exploring,burke2013hyper,lin2020genetic}.
It uses the same evolutionary procedure, fitness function, initialization scheme, and search and validation sets as \texttt{QuantumEvo}, differing only in heuristic generation.
The candidates are assembled by selecting and combining predefined algorithmic components from the search space summarized in \Cref{tab:nonllm-gene-axes}, rather than being generated and iteratively revised by an LLM.
Each individual encodes all dimensions, but only those associated with the selected family affect the resulting implementation.
The search space consists of three heuristic families, sifting, GA, and SA. 
These correspond to the three seed heuristics used to initialize \texttt{QuantumEvo}. 
The initial population contains ten individuals from each family, matching the round-robin initialization used by \texttt{QuantumEvo}. 
The non-LLM search can modify and combine a broad range of predefined components, including the sifting procedure embedded within the GA.
However, it cannot introduce a new algorithmic idea outside this predefined space, such as \textsc{MiniSift}'s change-aware refinement of only the levels affected by the preceding variation operator.

\begin{table}[h]
\centering
\small
\caption{
Search dimensions available to each heuristic family in the non-LLM evolutionary search.}
\label{tab:nonllm-gene-axes}
\begin{tabular}{llp{7.2cm}}
\toprule
Family & Search dimension & Available choices or range \\
\midrule
Top level
& Heuristic family
& Sifting / SA / GA. The initial population is divided evenly across the three families. \\
\midrule
\multirow[t]{2}{*}{Sifting}
& Search direction
& Bidirectional / top-down only / bottom-up only. \\

& Search range
& Full variable range / sliding window with width $w \in \{2,\ldots,10\}$ /
random subset containing $p \in [0.3,0.9]$ of the variables in each pass. \\
\midrule
\multirow[t]{4}{*}{Simulated annealing}
& Neighborhood operators
& Seed exchange--jump mixture / exchange only / jump only /
adjacent swap only / all four move types. \\

& Move probabilities
& Exchange probability $p_{\mathrm{exc}} \in [0.2,0.6]$;
upward-jump share $p_{\mathrm{up}} \in [0.2,0.8]$. \\

& Cooling schedule
& Multiplicative cooling factor $\alpha \in [0.80,0.95]$;
secondary temperature-scaling parameter $\beta \in [0.3,0.8]$. \\

& Restarts
& Number of additional re-annealing runs $r \in \{1,2,3\}$,
each initialized from the best solution found so far. \\
\midrule
\multirow[t]{6}{*}{Genetic algorithm}
& Population initialization
& Sifting-based / simulated-annealing-based / random-GA /
uniformly random / multi-family initialization. \\

& Population parameters
& Population size $n_{\mathrm{pop}} \in \{8,\ldots,40\}$;
crossover offspring ratio $\rho_{\mathrm{cross}} \in [0.5,3.0]$;
mutation rate $p_{\mathrm{mut}} \in [0,1]$;
elitism ratio $\rho_{\mathrm{elite}} \in [0,0.5]$. \\

& Parent selection
& Roulette-wheel selection / tournament selection with
$k \in \{2,\ldots,6\}$. \\

& Crossover operator
& Partially matched crossover / order crossover. \\

& Perturbation operator
& Adjacent swap / random-pair swap / insertion / block reversal. \\

& Offspring improvement
& Sifting-based refinement using the dimensions above /
adjacent-swap hill climbing. \\
\bottomrule
\end{tabular}
\end{table}

\subsection{Study on the Choice of LLM}
\label{app:ablation}

We compare two LLM choices under the same pipeline, prompt structure, and seed initialization: \texttt{gpt-oss-120b} (120B-parameter open-source MoE, Apache~2.0) and \texttt{gpt-5.4-mini} (OpenAI API).
For each model, we run ten independent searches, select the best individual on the validation set, and evaluate it on the 148 benchmark functions.

\begin{figure}[h]
\captionsetup[subfigure]{justification=centering}
\centering
\begin{subfigure}[t]{0.3\textwidth}
    \centering
    \includegraphics[width=\linewidth]{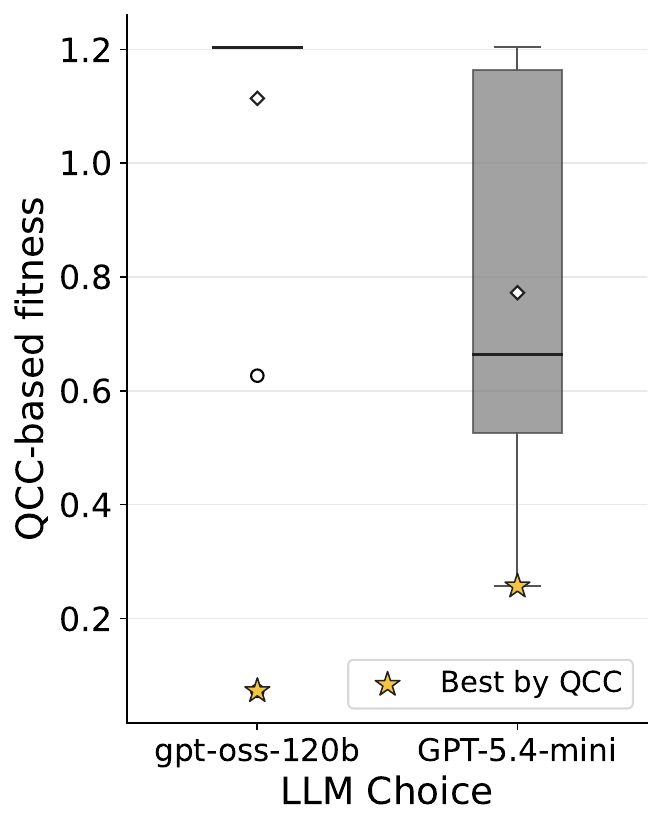}
    \caption{QCC-based fitness}
\end{subfigure}
\hspace{0.05\textwidth}
\begin{subfigure}[t]{0.3\textwidth}
    \centering
    \includegraphics[width=\linewidth]{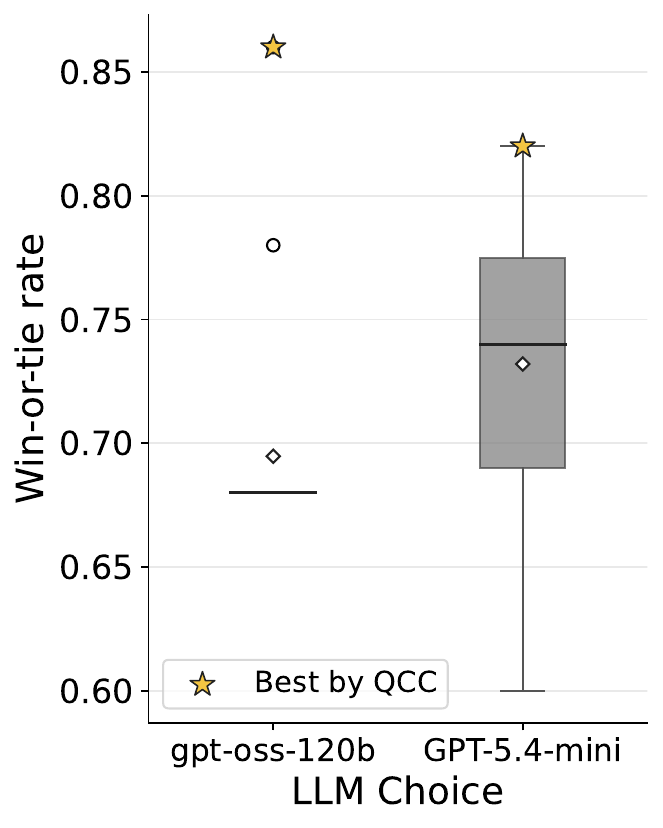}
    \caption{Win-or-tie rate}
\end{subfigure}
\caption{Validation comparison between the two LLM choices over ten independent runs. The star marks the run selected by the lowest validation QCC fitness.}
\label{fig:llm-val-boxplot}
\end{figure}

\begin{figure}[h]
\centering
\includegraphics[width=0.9\textwidth]{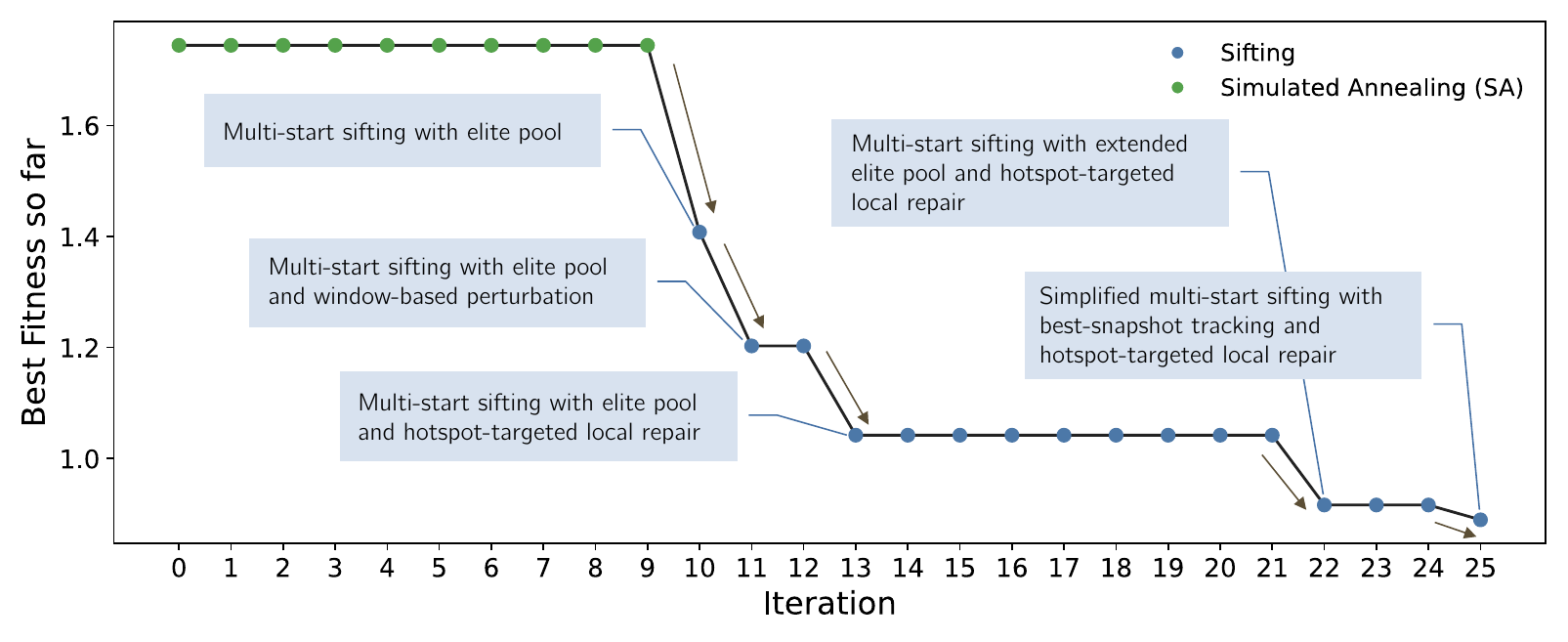}
\caption{Evolution of the Best Discovered Ordering Heuristic for the choice of \texttt{gpt-5.4-mini}}
\label{fig:gpt54mini-best-lineage}
\end{figure}

\Cref{fig:llm-val-boxplot} shows the distribution of validation results across ten independent runs.
Runs using \texttt{gpt-5.4-mini} show wider variance and a lower median validation score, and even the best run does not reach the best result obtained with \texttt{gpt-oss-120b}. 
\Cref{fig:gpt54mini-best-lineage} traces the search trajectory of the best \texttt{gpt-5.4-mini} run, showing how its best heuristic evolved during the search.

The search runs using \texttt{gpt-oss-120b} produce stronger validation results than those using \texttt{gpt-5.4-mini}, with a higher win-or-tie rate and lower mean QCC. 
The two settings also tend to converge to different heuristic families. 
Runs with \texttt{gpt-oss-120b} select SA-family individuals in 9 of 10 runs, whereas runs with \texttt{gpt-5.4-mini} more often converge to sifting-family individuals. 
Based on these validation results, we use the best individual generated with \texttt{gpt-oss-120b} for the main evaluation.

\begin{table}[h]
\centering
\caption{Benchmark comparison of LLM choices. Family counts summarize the best individual selected in each of ten independent runs per model.}
\label{tab:llm-ablation}
\small
\setlength{\tabcolsep}{3.5pt}
\begin{tabular}{lrrrrrrrr}
\toprule
 & \multicolumn{5}{c}{Benchmark} & \multicolumn{3}{c}{Family Type} \\
\cmidrule(lr){2-6}\cmidrule(lr){7-9}
LLM choice & \makecell{Best/Total\\(\%)} & \makecell{Strictly Best/Total\\(\%)} & \makecell{Mean\\Best QCC} & \makecell{Mean\\Avg QCC} & \makecell{Mean\\Worst QCC} & GA & SA & Sifting \\
\midrule
\texttt{gpt-oss-120b} & \textbf{70.9} & \textbf{13.5} & \textbf{1923.83} & \textbf{1979.76} & \textbf{2033.46} & 1 & 9 & 0 \\
\texttt{gpt-5.4-mini} & 66.9 & 6.1 & 2080.71 & 2147.33 & 2211.11 & 2 & 3 & 5 \\
\bottomrule
\end{tabular}
\end{table}

\begin{figure}[h]
\captionsetup[subfigure]{justification=centering}
\centering
\begin{subfigure}{0.9\textwidth}
    \centering
    \includegraphics[width=\linewidth]{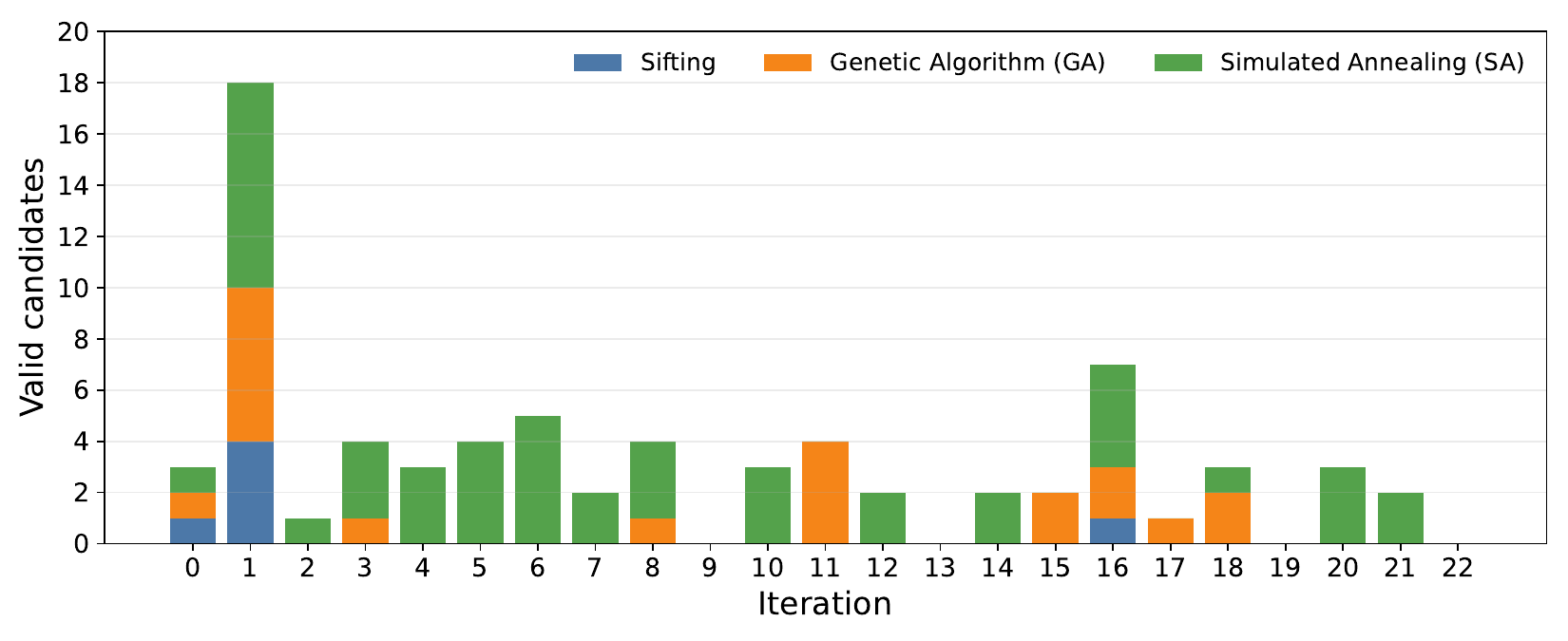}
    \caption{Best run using \texttt{gpt-oss-120b}}
    \label{fig:seed-family-counts-oss}
\end{subfigure}

\vspace{0.3cm}
\begin{subfigure}{0.9\textwidth}
    \centering
    \includegraphics[width=\linewidth]{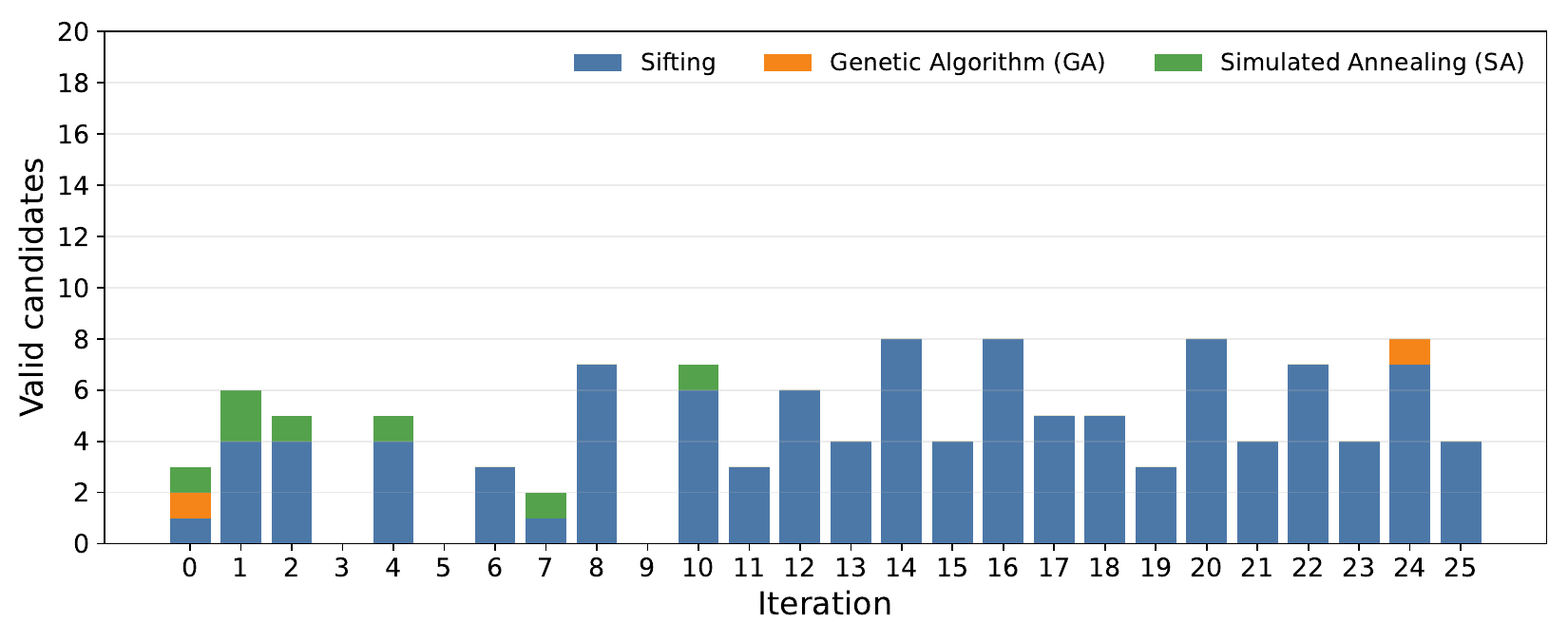}
    \caption{Best run using \texttt{gpt-5.4-mini}}
    \label{fig:seed-family-counts-gpt54mini}
\end{subfigure}
\caption{Code-level family composition of valid individuals across iterations}
\label{fig:seed-family-counts}
\end{figure}

\paragraph{Family Composition across Iterations}
We classify each valid individual by its code-level family.
\Cref{fig:seed-family-counts} traces the family composition over iterations for the best run from each LLM.
In the \texttt{gpt-oss-120b} run, the population shifts from diverse seed families toward SA-family and later GA-family individuals.
In contrast, \texttt{gpt-5.4-mini} tends to converge toward simpler sifting-style variants.

\subsection{Elite-Candidate Comparison across All Settings}

To provide a broader comparison across all seven settings considered in
\Cref{subsec:ablation} and this section, we additionally examine their elite
candidate populations rather than only the finalist selected from each run.
For each setting, we pool all candidates evaluated across its ten runs, rank
them by their logged search-set performance, and evaluate the top 5 and top 10 candidates on the validation set. 
The full \texttt{QuantumEvo} setting, which uses LLM guidance, multi-family initialization, and QCC-based fitness, achieves the best median validation fitness for both candidate pools.
For the top 10 pool, however, its distribution widens substantially, whereas the non-LLM control remains tightly concentrated.
Given the substantial overlap and the trade-off between median performance and variability, these results do not establish a clear overall advantage for either setting.

\begin{figure}[h]
\centering
\includegraphics[width=0.85\linewidth]{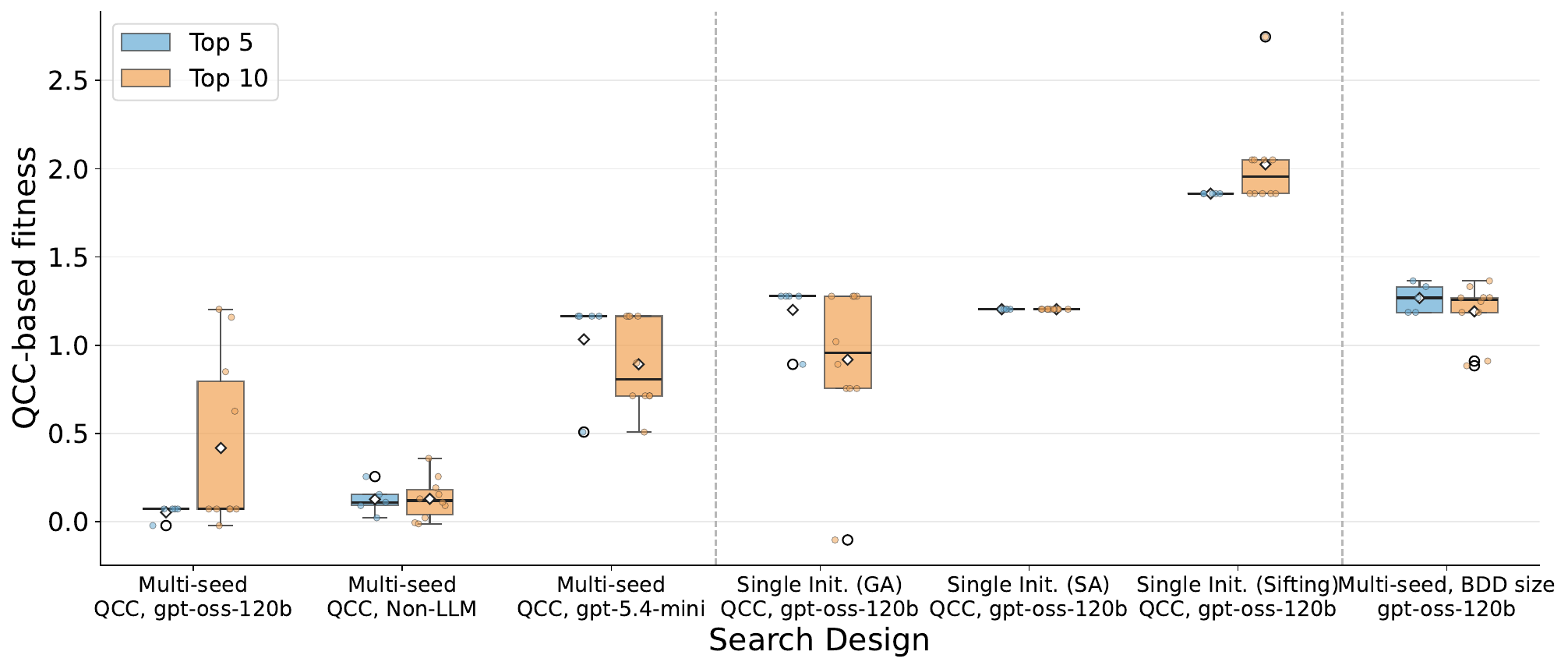}
\caption{Validation-set fitness of the top 5 and top 10 candidates ranked by search-set performance after pooling candidates across the 10 runs for each setting. Dashed lines separate the guidance, initialization, and fitness comparisons.}
\label{fig:elite-candidate-boxplot}
\end{figure}

\subsection{Sensitivity Analysis}

We provide a post-hoc analysis of the search set, validation set, and evaluation timeout. 
The analysis reuses candidates and runtime logs from the completed searches and characterizes sensitivity within the observed search outcomes.

\paragraph{Search set.}
We reevaluate the 30 leading candidates from the search run that produced the HGA-QE algorithm, across the original and nine alternate 100-function search sets, as reported in \Cref{tab:validation-sensitivity}.
HGA-QE ranks first in 5 of these 10 search sets, although the top-ranked candidate varies across sets.
These results characterize sensitivity within the candidate pool observed in the completed run.

\paragraph{Validation set.}
For each setting, the original validation set selects one finalist from 10 independent runs.
We repeat this selection using nine alternate 50-function validation sets and report the same-run rate in \Cref{tab:validation-sensitivity}.
The full \texttt{QuantumEvo} setting and the GA-only initialization are comparatively stable, whereas the other settings are more sensitive to the validation-set choice.
Overall, the HGA-QE finalist selected under the full \texttt{QuantumEvo} setting is more robust to validation-set variation than those selected under the alternative settings.

\begin{table}[h]
\centering
\small
\caption{Sensitivity over the original and nine alternate sets. Rates denote top ranking of HGA-QE among 30 candidates or reselection of the original validation finalist.}
\label{tab:validation-sensitivity}
\begin{tabular}{llllc}
\toprule
Analysis & Initialization & Fitness & Guidance & Rate \\
\midrule
Search set & Multi-seed & QCC & \texttt{gpt-oss-120b} & 5/10 \\
\midrule
\multirow{7}{*}{Validation set}
& \multirow{3}{*}{Multi-seed} & \multirow{3}{*}{QCC} & \textbf{\texttt{gpt-oss-120b}} & 9/10 \\
& & & \texttt{gpt-5.4-mini} & 2/10 \\
& & & Non-LLM & 1/10 \\
\cmidrule{2-5}
& GA only & QCC & \texttt{gpt-oss-120b} & 8/10 \\
& SA only & QCC & \texttt{gpt-oss-120b} & Identical heuristic \\
& Sifting only & QCC & \texttt{gpt-oss-120b} & 4/10 \\
\cmidrule{2-5}
& Multi-seed & BDD size & \texttt{gpt-oss-120b} & 5/10 \\
\bottomrule
\end{tabular}
\end{table}

\paragraph{Timeout.}
We replay the recorded wall-clock times of candidate evaluations from the 10 production runs under shorter hypothetical timeout limits.
An evaluation is counted as a failure when its recorded time exceeds the limit. 
The evaluations that already timed out at 40 seconds remain failures.
The failure rate increases from 8.3\% at 40 seconds to 9.8\% at 32 seconds, then more rapidly to 17.7\% at 20 seconds in \Cref{tab:timeout-sensitivity}.

\begin{table}[h]
\centering
\small
\caption{Candidate-evaluation failure rates under shorter timeout limits, based on the recorded wall times of evaluations from the 10 production runs.}
\label{tab:timeout-sensitivity}
\begin{tabular}{lrr}
\toprule
Budget fraction & Budget (s) & Failure rate \\
\midrule
1.0$\times$ & 40.0 & 8.3\% \\
0.9$\times$ & 36.0 & 9.0\% \\
0.8$\times$ & 32.0 & 9.8\% \\
0.7$\times$ & 28.0 & 11.6\% \\
0.6$\times$ & 24.0 & 14.1\% \\
0.5$\times$ & 20.0 & 17.7\% \\
\bottomrule
\end{tabular}
\end{table}

\section{Additional Analysis of HGA-QE}
\subsection{Mapping to CUDD Implementations}
\label{app:impl-notes}

\Cref{alg:partialsift} and~\Cref{alg:minisift} define the subroutines called by \Cref{alg:cuddFunc}.
\Cref{tab:cudd-correspondence} maps each pseudocode function to its corresponding CUDD function and source file.

\begin{algorithm}[h]
\caption{\textsc{PartialSift}: Individual-order evaluator with \textsc{MiniSift} (subroutine of \Cref{alg:cuddFunc})}
\label{alg:partialsift}
\begin{algorithmic}[1]
\Require BDD manager $T$, individual permutation $\sigma$, range $[\ell, u]$
\Ensure $\sigma$ is replaced by the post-sift ordering; return the resulting BDD size

\State $n \gets u-\ell+1$
\State $\pi[i] \gets$ variable currently at level $\ell+i$, for $i=0,\ldots,n-1$
\Comment{snapshot before any move}
\For{$j = 0$ \textbf{to} $n-1$}
    \State Move variable $\sigma[j]$ to target level $\ell+j$ via adjacent swaps
\EndFor
\State $\Delta \gets \{i \in [0,n) : \sigma[i] \neq \pi[i]\}$
\Comment{levels changed by imposing the target order, before \textsc{MiniSift}}
\If{$\Delta \neq \emptyset$}
    \State $\textsc{MiniSift}(T,\, \ell,\, u,\, \Delta)$
\EndIf
\State $\sigma[i] \gets$ variable currently at level $\ell+i$, for $i=0,\ldots,n-1$
\State \textbf{return} $|T.\mathrm{nodes}|$
\end{algorithmic}
\end{algorithm}

\begin{algorithm}[h]
\caption{\textsc{MiniSift}: targeted local sifting (subroutine of \Cref{alg:cuddFunc} and~\Cref{alg:partialsift})}
\label{alg:minisift}
\begin{algorithmic}[1]
\Require BDD manager $T$, range $[\ell, u]$, changed-level set $\Delta \subseteq [0, n)$
\Ensure Marked levels are used as starting levels for local CUDD sifting within $[\ell, u]$

\Statex \textbf{Upward pass: low levels first}
\For{$i = \ell$ \textbf{to} $u$}
    \If{$i - \ell \notin \Delta$} \textbf{skip} \EndIf
    \State Sift variable at level $i$ upward within $[\ell, u]$; restore the best position found
\EndFor
\Statex \textbf{Downward pass: high levels first}
\For{$i = u$ \textbf{downto} $\ell$}
    \If{$i - \ell \notin \Delta$} \textbf{skip} \EndIf
    \State Sift variable at level $i$ downward within $[\ell, u]$; restore the best position found
\EndFor
\end{algorithmic}
\end{algorithm}

\begin{table}[h]
\captionsetup[subfigure]{justification=centering}
\centering
\caption{Correspondence between pseudocode functions and CUDD
  source.  \emph{W}\,=\,wrapped (unmodified API call),
  \emph{A}\,=\,adapted, \emph{N}\,=\,new.}
\label{tab:cudd-correspondence}
\small
\begin{tabular}{llllp{4.2cm}}
\toprule
Pseudocode & CUDD function & Source file & Type & Notes \\
\midrule
\textsc{Sift}
  & \texttt{cuddSifting}
  & \texttt{cuddReorder.c}
  & W & --- \\
\textsc{SwapAdjacent}
  & \texttt{cuddSwapInPlace}
  & \texttt{cuddReorder.c}
  & W & --- \\
\textsc{PMX}
  & \texttt{PMX}
  & \texttt{cuddGenetic.c}
  & W & --- \\
\textsc{RouletteSelect}
  & \texttt{roulette}
  & \texttt{cuddGenetic.c}
  & A & Spin range unified to \texttt{popsize} for both parent selections \\
\textsc{PartialSift}
  & \texttt{build\_dd}
  & \texttt{cuddGenetic.c}
  & A & Replaces full \texttt{cuddSifting}
        with \textsc{MiniSift} over marked
        starting levels \\
\textsc{RestoreOrder}
  & \texttt{restoreOrder}
  & \texttt{cuddAnneal.c}
  & A & Reimplemented locally; based on
        annealing restore logic
        (cf.\ \texttt{ddShuffle} in
        \texttt{cuddReorder.c}) \\
\textsc{MiniSift}
  & ---
  & ---
  & N & New function; internally calls
        \texttt{ddSiftingUp},
        \texttt{ddSiftingDown},
        \texttt{ddSiftingBackward} \\
\bottomrule
\end{tabular}
\end{table}

\subsection{Diagnostic Analysis of \textsc{MINISIFT} Algorithm}
\label{app:sift_minisift_compare}

This section provides additional diagnostics for the comparison between the standard sifting and \textsc{MiniSift}. 
The goal is to understand why \textsc{MiniSift} can improve downstream QCC.

\paragraph{Metrics}
The diversity statistics are computed over all 100 functions in the search set and three independent seeds.
The unique fraction is the fraction of distinct candidate orderings among the final GA population. 
The Hamming fraction measures the average normalized Hamming distance between pairs of candidate orderings, where a larger value indicates that the candidate orderings differ at more variable positions.
The remaining metrics are computed on the 31 circuits where GA with the standard sifting and GA with \textsc{MiniSift} produce the same BDD size, but \textsc{MiniSift} achieves lower QCC. 
Child-node line preservation cases occur when a child value cannot be overwritten at a combining step because it remains required by another reference later in the synthesis. 
Fewer such cases indicate fewer restrictions on the use of circuit lines and are associated with fewer Toffoli gates and total controls. 
The Toffoli gates, total controls, and QCC are computed from the synthesized reversible circuits.

\paragraph{Statistical tests}
All statistical tests are paired tests between GA with the standard sifting and GA with \textsc{MiniSift} on the same set of circuits or circuit-seed pairs. 
The Wilcoxon signed-rank test additionally uses the ranks of the absolute paired differences, and therefore accounts for both the direction and relative magnitude of the changes. 
Smaller $p$-values indicate stronger evidence that the observed changes are systematic rather than random paired fluctuations.

\begin{table}[h]
    \centering
    \caption{Diagnostic comparison between the standard sifting and \textsc{MiniSift}. Diversity rows use all search-set functions and three seeds; the remaining rows use cases in which both methods produce the same BDD size but \textsc{MiniSift} lowers QCC. The change is relative to the standard sifting.}
    \begin{tabular}{lrrrr}
        \toprule
        Statistic & Sifting & \textsc{MiniSift} & Change (\%) & Wilcoxon $p$ \\
        \midrule
        Unique fraction & 0.553 & 0.615 & $+11.2\%$ &  $4.35{\times}10^{-11}$ \\
        Hamming fraction & 0.507 & 0.537 & $+5.9\%$ & $1.04{\times}10^{-4}$ \\
        \midrule
        Child-node line preservation cases & 999 & 983 & $-1.6\%$ & 0.0100 \\
        Toffoli gates & 3847 & 3797 & $-1.3\%$ & $4.70{\times}10^{-5}$ \\
        Total controls & 10591 & 10479 &$-1.1\%$ & $4.98{\times}10^{-6}$ \\
        Total QCC & 22550 & 22274 & $-1.2\%$  & $1.12{\times}10^{-6}$ \\
        \bottomrule
        \end{tabular}
    \label{tab:statistics}
\end{table}

\Cref{tab:statistics} shows that \textsc{MiniSift} produces a more diverse set of candidate orderings than the standard sifting. 
On the same BDD size subset where \textsc{MiniSift} achieves lower QCC, it also reduces the child-node line preservation cases, Toffoli gates, total controls, and QCC. 
These results suggest that \textsc{MiniSift} can expose ordering candidates that have comparable BDD size but are more favorable for synthesis.

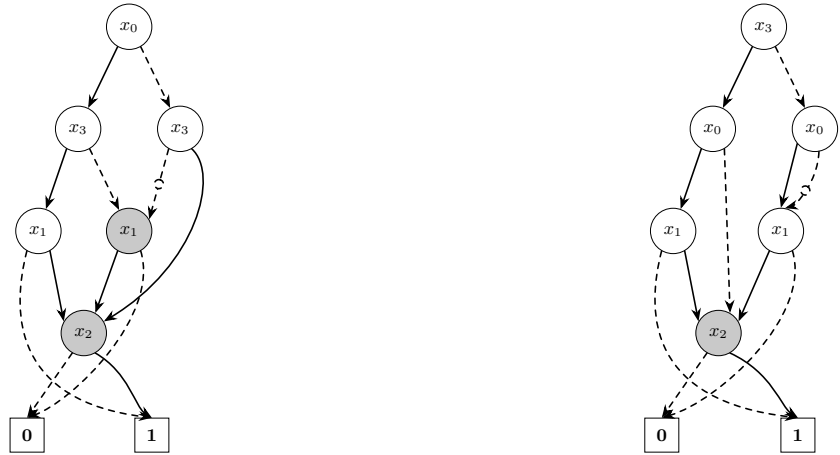
\begin{figure}[h]
\centering
\captionsetup[subfigure]{justification=centering}
\begin{subfigure}[t]{0.48\textwidth}
    \centering
    \scalebox{0.75}{\definecolor{sharedgray}{HTML}{C9C9C9}
\begin{tikzpicture}[
    bdd node/.style={circle, draw, minimum size=8mm, inner sep=0pt, font=\small\bfseries},
    terminal/.style={rectangle, draw, minimum size=6mm, font=\small\bfseries},
    one edge/.style={-{Stealth}, thick, line cap=round, line join=round},
    zero edge/.style={-{Stealth}, dashed, thick, line cap=round, line join=round}
]

    \node[bdd node] (nF) at (0.00,0.00)   {$x_0$};
    \node[bdd node] (nD) at (-0.90,-1.80) {$x_3$};
    \node[bdd node] (nE) at (0.90,-1.80)  {$x_3$};
    \node[bdd node] (nB) at (-1.60,-3.60) {$x_1$};
    \node[bdd node, fill=sharedgray] (nC) at (0.00,-3.60)  {$x_1$};
    \node[bdd node, fill=sharedgray] (nA) at (-0.80,-5.40) {$x_2$};

    \node[terminal] (const0) at (-1.80,-7.20) {0};
    \node[terminal] (const1) at (0.40,-7.20)  {1};

    \draw[one edge]  (nF.240) -- (nD);
    \draw[zero edge] (nF.300) -- (nE);

    \draw[one edge]  (nD.240) -- (nB);
    \draw[zero edge] (nD.300) -- (nC);

    \draw[one edge]  (nE.300) .. controls (1.60,-2.60) and (1.20,-4.20) .. (nA.30);
    \draw[zero edge] (nE.240) -- (nC.30)
        node[midway, sloped, draw, circle, minimum size=1.6mm, inner sep=0pt, fill=white] {};

    \draw[one edge]  (nB.300) -- (nA.150);
    \draw[zero edge] (nB.240) .. controls (-2.10,-5.10) and (-2.00,-6.60) .. (const1.north);

    \draw[one edge]  (nC.240) -- (nA.60);
    \draw[zero edge] (nC.300) .. controls (0.55,-5.10) and (-1.00,-6.60) .. (const0.north);

    \draw[one edge]  (nA.300) .. controls (0.00,-6.10) and (0.10,-6.70) .. (const1.north);
    \draw[zero edge] (nA.240) -- (const0.north);

\end{tikzpicture}}
    \caption{A BDD produced by sifting: Size = 7, QCC = 40}
\end{subfigure}
\hfill
\begin{subfigure}[t]{0.48\textwidth}
    \centering
    \scalebox{0.75}{\definecolor{sharedgray}{HTML}{C9C9C9}
\begin{tikzpicture}[
    bdd node/.style={circle, draw, minimum size=8mm, inner sep=0pt, font=\small\bfseries},
    terminal/.style={rectangle, draw, minimum size=6mm, font=\small\bfseries},
    one edge/.style={-{Stealth}, thick, line cap=round, line join=round},
    zero edge/.style={-{Stealth}, dashed, thick, line cap=round, line join=round}
]

    \node[bdd node] (nF) at (0.00,0.00)   {$x_3$};
    \node[bdd node] (nD) at (-0.90,-1.80) {$x_0$};
    \node[bdd node] (nE) at (0.90,-1.80)  {$x_0$};
    \node[bdd node] (nB) at (-1.60,-3.60) {$x_1$};
    \node[bdd node] (nC) at (0.30,-3.60)  {$x_1$};
    \node[bdd node, fill=sharedgray] (nA) at (-0.80,-5.40) {$x_2$};

    \node[terminal] (const0) at (-1.80,-7.20) {0};
    \node[terminal] (const1) at (0.60,-7.20)  {1};

    \draw[one edge]  (nF.240) -- (nD);
    \draw[zero edge] (nF.300) -- (nE);

    \draw[one edge]  (nD.240) -- (nB);
    \draw[zero edge] (nD.300) -- (nA.60);

    \draw[one edge]  (nE.220) -- (nC.90);
    \draw[zero edge] (nE.280) .. controls (0.95,-2.75) and (0.55,-3.15) .. (nC.80)
        node[midway, sloped, draw, circle, minimum size=1.6mm, inner sep=0pt, fill=white] {};

    \draw[one edge]  (nB.300) -- (nA.150);
    \draw[zero edge] (nB.240) .. controls (-2.10,-5.10) and (-2.00,-6.60) .. (const1.north);

    \draw[one edge]  (nC.240) -- (nA.30);
    \draw[zero edge] (nC.300) .. controls (0.85,-5.10) and (-0.80,-6.60) .. (const0.north);

    \draw[one edge]  (nA.300) .. controls (0.15,-6.10) and (0.30,-6.70) .. (const1.north);
    \draw[zero edge] (nA.240) -- (const0.north);

\end{tikzpicture}}
    \caption{A BDD produced by \textsc{MiniSift}: Size = 7, QCC = 30}
\end{subfigure}
\caption{A representative same-BDD-size example under sifting and \textsc{MiniSift}. Gray nodes are referenced by at least two parents and indicate candidate sites for child-node line preservation. \textsc{MiniSift} preserves the BDD size while eliminating one preservation case in the synthesis trace, reducing QCC.}
\label{fig:minisift-bdd-example}
\end{figure}

\paragraph{Example.}
\Cref{fig:minisift-bdd-example} shows a representative four-input, single-output augmented RevLib function from the search set, which is one of the 31 functions analyzed in \Cref{tab:statistics}. 
Standard sifting makes the
order $(x_0,x_3,x_1,x_2)$, whereas \textsc{MiniSift} makes
$(x_3,x_0,x_1,x_2)$. 
Swapping only the top two variables preserves the BDD
size while reducing QCC from 40 to 30. 
Solid and dashed edges denote then and else branches, respectively.
An open circle marks a complemented edge, indicating that the complemented value of the referenced child function is used.
Gray nodes are referenced by at least two parents.
In the BDD produced by standard sifting, the upper-right $x_3$ node has two children, the gray $x_1$ and gray $x_2$ nodes. 
Because the gray $x_1$ node is also referenced elsewhere, its synthesized value must be preserved at this step, producing a child-node line preservation case. 
In the BDD produced by \textsc{MiniSift}, the corresponding upper-right $x_0$ node instead points to the same $x_1$ child on both branches, with one branch complemented. 
This eliminates that preservation case.
Accordingly, the synthesis trace contains one fewer child-node line preservation case, together with two fewer Toffoli gates and five fewer controls.
Across all 31 same-BDD-size functions, the results in \Cref{tab:statistics} are consistent with this observation, showing fewer
preservation cases together with statistically significant reductions in
Toffoli gates, total controls, and QCC.

\subsection{Per-instance performance comparison}
The following presents statistical tests comparing HGA-QE against the baselines, followed by an analysis of the loss cases.
\subsubsection{Statistical test}
\Cref{tab:stat-tests-full} evaluates both the magnitude and the consistency of HGA-QE’s per-circuit QCC differences against all six baselines. 
Mean $\Delta$QCC measures the average size and direction of the difference, where a negative value favors HGA-QE. 
For each baseline, we apply a paired Wilcoxon signed-rank test to the per-circuit QCC differences.
The raw $p$-value reports the result of that comparison alone, and the Holm-adjusted $p$-value accounts for the six baseline comparisons conducted within the same suite. 
Statistical significance is determined using the adjusted values at $\alpha=0.05$.
Across all 148 functions, HGA-QE achieves significantly lower QCC than every baseline after Holm correction. 
The suite-level evidence is less uniform. 
On RevLib, HGA-QE significantly outperforms the three sifting-based methods and BDD2Seq, while no significant difference is detected relative to GA or SA. 
On LGSynth91, HGA-QE significantly outperforms every baseline except GA. 
No comparison is significant on ISCAS85/89. 
This suite contains only four circuits and several tied results, so the tests have insufficient power and should not be interpreted as evidence that the methods perform equally.

\begin{table}[h]
\centering
\small
\caption{Paired Wilcoxon signed-rank tests comparing the per-circuit best QCC of HGA-QE with each baseline. The six comparisons within each suite are corrected using the Holm--Bonferroni procedure. Mean $\Delta$QCC is computed as HGA-QE minus the baseline, so negative values favor HGA-QE. Bold Holm-adjusted $p$-values indicate significance at $\alpha=0.05$.}
\label{tab:stat-tests-full}
{
\begin{tabular}{llrrr}
\toprule
Suite & Baseline & Mean $\Delta$QCC & $p$ (raw) & $p$ (Holm) \\
\midrule
RevLib (96) 
 & SA & -0.21 & 0.571 & 0.571 \\
 & BDD2Seq & -39.99 & 0.008  & \textbf{0.025} \\
 & GA & -29.24 & 0.069 & 0.139 \\
 & SIFT & -351.20 & $8.1\times10^{-8}$ & $\bm{3.2\times10^{-7}}$ \\
 & SYMM\_SIFT & -341.85 & $5.0\times10^{-8}$ & $\bm{2.5\times10^{-7}}$ \\
 & GROUP\_SIFT & -272.52 & $6.9\times10^{-9}$ & $\bm{4.1\times10^{-8}}$ \\
\midrule
LGSynth91 (48) 
 & SA & -65.04 & 0.003 & \textbf{0.006} \\
 & BDD2Seq & -377.46 & $7.0\times10^{-4}$ & \textbf{0.002} \\
 & GA & -123651.40 & 0.055 & 0.055 \\
 & SIFT & -2599.35 & $4.5\times10^{-6}$ & $\bm{2.2\times10^{-5}}$ \\
 & SYMM\_SIFT & -2594.17 & $4.3\times10^{-6}$ & $\bm{2.2\times10^{-5}}$ \\
 & GROUP\_SIFT & -3674.27 & $2.6\times10^{-6}$ & $\bm{1.5\times10^{-5}}$ \\
\midrule
ISCAS85/89 (4) 
 & SA & -151.75 & 0.317 & 0.653 \\
 & BDD2Seq & -1415.25 & 0.109 & 0.653 \\
 & GA & -161.75 & 0.317 & 0.653 \\
 & SIFT & -7095.75 & 0.125 & 0.653 \\
 & SYMM\_SIFT & -7095.75 & 0.125 & 0.653 \\
 & GROUP\_SIFT & -11351.75 & 0.109 & 0.653 \\
\midrule
All (148) 
 & SA & -25.33 & 0.004 & \textbf{0.008} \\
 & BDD2Seq & -186.61 & $4.0\times10^{-6}$ & $\bm{1.2\times10^{-5}}$ \\
 & GA & -40126.49 & 0.006 & \textbf{0.008} \\
 & SIFT & -1262.61 & $3.2\times10^{-13}$ & $\bm{1.3\times10^{-12}}$ \\
 & SYMM\_SIFT & -1254.87 & $1.3\times10^{-13}$ & $\bm{6.6\times10^{-13}}$ \\
 & GROUP\_SIFT & -1675.23 & $1.9\times10^{-14}$ & $\bm{1.1\times10^{-13}}$ \\

\bottomrule
\end{tabular}
}

\end{table}

\subsubsection{Loss cases}
\Cref{tab:failure-breakdown} provides two complementary views of the loss cases. 
\Cref{tab:failure-breakdown}\subref{tab:failure-loss-rates}
applies the same best-or-not criterion, where ties count as best, to HGA-QE and the three strongest baselines.
HGA-QE has the lowest loss rate overall and on every benchmark suite.
\Cref{tab:failure-breakdown}\subref{tab:failure-loss-breakdown}
further shows that, among HGA-QE's 43 losses, BDD2Seq and SA account for 21 and 11 cases, respectively, confirming that they remain its strongest competitors.

\begin{table}[t]
\centering
\small
\setlength{\tabcolsep}{4pt}
\captionsetup[subtable]{justification=centering,singlelinecheck=false}

\caption{Comparison and breakdown of non-best cases.}
\label{tab:failure-breakdown}

\begin{subtable}{\textwidth}
\centering
\caption{Loss rates by method.}
\label{tab:failure-loss-rates}
\begin{tabular}{lrrrr}
\toprule
Method & RevLib & LGSynth91 & ISCAS85/89 & Overall (148) \\
\midrule
HGA-QE & 26/96 & 17/48 & 0/4 & \textbf{43/148 (29.1\%)} \\
SA & 28/96 & 26/48 & 1/4 & 55/148 (37.2\%) \\
BDD2Seq & 31/96 & 26/48 & 3/4 & 60/148 (40.5\%) \\
GA & 38/96 & 25/48 & 1/4 & 64/148 (43.2\%) \\
\bottomrule
\end{tabular}
\end{subtable}

\vspace{7pt}

\begin{subtable}{\textwidth}
\centering
\caption{Breakdown of HGA-QE losses.}
\label{tab:failure-loss-breakdown}
\begin{tabular}{lrrrrr}
\toprule
Suite & HGA-QE losses & BDD2Seq & SA & GA & Others \\
\midrule
RevLib & 26/96 & 15 & 7 & 1 & 3 \\
LGSynth91 & 17/48 & 6 & 4 & 6 & 1 \\
ISCAS85/89 & 0/4 & 0 & 0 & 0 & 0 \\
\midrule
Overall & 43/148 & 21 & 11 & 7 & 4 \\
\bottomrule
\end{tabular}
\end{subtable}
\end{table}

\subsection{Application of HGA-QE to a Logic-Synthesis Flow}
We further evaluate HGA-QE in a Field-Programmable Gate Array (FPGA) logic-synthesis flow.
Prior work has examined how BDD variable ordering affects the quality
of circuits synthesized from the resulting BDDs
~\citep{popel2002synthesislowpowerdigitalcircuits, drechsler2004synthesis}.
We extend this evaluation to FPGA technology mapping, which transforms a Boolean network into a network of fixed-input lookup tables (LUTs) supported by the target FPGA architecture~\citep{cong1994flowmap}.
For this purpose, each ordered BDD is exported as a Boolean network
in which each BDD node is represented as a multiplexer. 
The structural size of the resulting multiplexer network closely follows the BDD size.
The network is then synthesized and mapped to four-input LUTs using Yosys~\citep{wolf2013yosys}.

We conduct the experiment on 52 functions from LGSynth91 and
ISCAS85/89, as these suites contain conventional Boolean logic
benchmarks that are more directly aligned with FPGA synthesis and
technology mapping than the reversible circuits in RevLib.
The result is summarized in \cref{tab:yosys-combined}.
Sifting is deterministic and is run once, whereas GA, SA, and HGA-QE are each run with five seeds.
For each benchmark and method, we report the best, average, and worst multiplexer-network sizes and LUT counts across the runs.
Aggregate statistics include only benchmarks for which all required seeds completed within the 300s budget for each stage. 
HGA-QE completed only partially on \texttt{c432} and \texttt{c880}, while no seed completed on \texttt{rot\_orig}.
For GA, downstream mapping completed only partially on \texttt{pair\_orig} and did not complete on \texttt{rot\_orig}.
These benchmarks were therefore excluded from the aggregate statistics.
HGA-QE achieves the highest Best/Total count for the exported mux networks. 
But it does not outperform the strongest baselines, SA and GA, under the objective related to FPGA mapping.

\begin{table}[t]
\centering
\caption{Mux-network size and LUT-mapping results under a uniform 300s-per-seed budget for each stage. The Net. Size and \# LUT rows use 49 and 48 complete-case circuits, respectively, for which all methods completed their required runs. Success reports each method's full completion count over all 52 circuits. Best includes ties, whereas Strictly Best denotes a unique best result. Mean quality and geometric mean runtime values are computed over the same complete-case set within each stage.}
\label{tab:yosys-combined}
\resizebox{\linewidth}{!}{
\begin{tabular}{ll|rrr|rrrr}
\toprule
Method & Stage & Success & Best & Strict & Mean Best & Mean Avg & Mean Worst & Geo-mean Runtime (s) \\
\midrule
\multirow{2}{*}{HGA-QE}
 & Net. Size & 49/52 & \textbf{47/49} & 5/49 & 217.39 & 222.12 & \textbf{227.86} & 0.066 \\
 & \# LUT     & 49/52 & 30/48          & 3/48 & 83.46  & 87.45  & 91.73           & \textbf{3.036} \\
\midrule
\multirow{2}{*}{SA}
 & Net. Size & \textbf{52/52} & 36/49          & 1/49 & \textbf{215.47} & \textbf{222.06} & 235.41 & 0.092 \\
 & \# LUT     & \textbf{52/52} & \textbf{35/48} & 6/48 & 85.04           & 90.52           & 98.50  & 3.041 \\
\midrule
\multirow{2}{*}{GA}
 & Net. Size & \textbf{52/52} & 42/49 & 1/49 & 274.24         & 1575.83        & 2564.90        & 0.080 \\
 & \# LUT     & 50/52          & 30/48 & 4/48 & \textbf{83.38} & \textbf{87.24} & \textbf{90.35} & 3.060 \\
\midrule
\multirow{2}{*}{SIFT}
 & Net. Size & \textbf{52/52} & 18/49 & 0/49 & 369.10 & 369.10 & 369.10 & \textbf{0.023} \\
 & \# LUT     & \textbf{52/52} & 30/48 & 4/48 & 134.52 & 134.52 & 134.52 & 3.229 \\
\bottomrule
\end{tabular}

}
\end{table}

\section{Offline and Online Cost}
\label{app:offline-online-cost}

We distinguish between the one-time offline cost of discovering HGA-QE and the recurring online cost of applying the discovered heuristic to a new Boolean function. 
The offline phase includes LLM inference, candidate generation, compilation, evaluation on the search set, reversible synthesis, and QCC evaluation. 
The online phase begins after HGA-QE has been selected and compiled. 
It requires no LLM calls and consists only of applying the fixed \texttt{C} language heuristic, followed by synthesis and QCC evaluation.

\paragraph{Offline Search Cost}
The \texttt{QuantumEvo} setting consists of 10 independent searches using \texttt{gpt-oss-120b}. Across these runs, each search used an average of 291.6 LLM calls, 3.11M input tokens, and 1.04M output tokens. 
The mean wall-clock time was 108.5 minutes. %
The model was self-hosted on a single NVIDIA RTX PRO 6000 GPU. 
We report only the inference resources used during the search and treat the cost of pretraining the underlying LLM as external.
Classical CUDD heuristics require no analogous discovery phase.
BDD2Seq reports its training hardware and hyperparameters, but not the wall-clock time or GPU-hours required for model training. 
Its original offline cost therefore cannot be directly compared with that of \texttt{QuantumEvo}.

\begin{table}[h]
\centering
\caption{Average offline search requirements of one \texttt{QuantumEvo} production run.}
\label{tab:offline-cost}
\small
\begin{tabular}{rrrrl}
\toprule
LLM Calls & Input Tokens & Output Tokens & Wall-clock & Hardware \\
\midrule
291.6 & 3.11M & 1.04M & 108.5 min & 1$\times$ NVIDIA RTX PRO 6000 \\
\bottomrule
\end{tabular}
\end{table}

\paragraph{Offline Compilation and Evaluation Cost}

Each candidate reaching the evaluation stage is compiled and evaluated on the 100-function search set. 
For every function, the evaluation invokes the reversible-synthesis pipeline twice before computing QCC. 
With a maximum of 200 candidate evaluations per run, this corresponds to up to approximately 40,000 synthesis calls per run.
In the \texttt{QuantumEvo} setting, 8.3\% of candidate evaluations reached the 40\,s timeout. 
The completed evaluations required a mean of 9.3\,s, a median of 7.0\,s, and a 95th percentile of 26.0\,s.

\begin{figure}[h]
\centering
\includegraphics[width=0.62\linewidth]{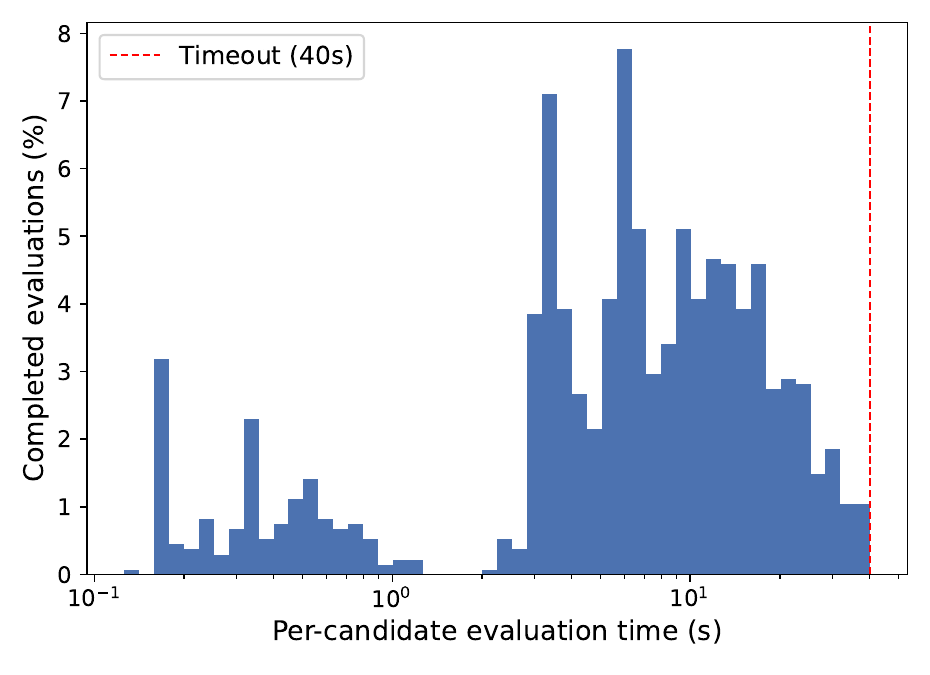}
\caption{Distribution of per-candidate evaluation times for completed evaluations in the production setting. The dashed line indicates the 40\,s timeout threshold.}
\label{fig:runtime-distribution}
\end{figure}

\paragraph{Online Cost}
Applying HGA-QE requires no LLM calls. 
After discovery, HGA-QE is deployed as a fixed compiled heuristic. 
Its online cost therefore consists only of BDD reordering, reversible synthesis, and QCC evaluation, all of which are included in the benchmark runtime measurements.
Unlike HGA-QE, BDD2Seq requires per-instance neural inference.

\section{Prompts}
We show the problem-specific prompts used in \texttt{QuantumEvo}, which comprise a function signature, function description, and external knowledge.
They are tailored to BDD variable ordering heuristic design and to the representation of individuals in \texttt{C}.
The remaining operator prompts, including mutation, crossover, and reflection, are available in the accompanying code.
\begin{promptbox}
int cuddFunc(DdManager *table, int lower, int upper)
\end{promptbox}
\vspace{1mm}{\centering\small Prompt 1: Function Signature\par}

\vspace{2mm}

\begin{promptbox}
Implement `cuddFunc()` for CUDD BDD reordering. \\
Primary goal: reduce native BDD size and Quantum Cost. \\
If output hooks are present, you may also consult `CURRENT\_SIZE(table)`, but native BDD heap size is a valid primary objective. \\

Requirements: \\
- Return `1` on success and `0` on failure. \\
- On OOM, set `table->errorCode = CUDD\_MEMORY\_OUT` and return `0`. \\
- On return, the live order must equal the best order found. \\
- Respect the active level range `[lower, upper]`. \\
- Free owned arrays with `FREE` and owned move lists with `cuddDeallocMove`. \\

Key CUDD contracts: \\
- `cuddSwapInPlace(table, x, x + 1)` takes LEVELS, not variable indices. Pass the lower adjacent level first; `cuddSwapInPlace(table, curLevel, curLevel - 1)` is wrong. \\
- Keep `Cudd\_SetFuncOutputs(...)` and `Cudd\_ClearFuncOutputs(...)` exactly named. \\
- `ddUndoMoves(moves)` returns an inverse history; free the original move list after calling it, and free the full list, not just the head node. \\
- For plain sifting, compose the rollback list as `down + inv` before `ddSiftingBackward(...)`. \\
- `ddSiftingBackward(...)` does not free the move list for you. \\
- For annealing-style code, keep an explicit best-order snapshot and restore the best order before returning.
\end{promptbox}
\vspace{1mm}{\centering\small Prompt 2: Function Description\par}

\vspace{2mm}

\begin{promptbox}
`cuddFunc()` should behave like a native CUDD reorderer. \\

Common mistakes: \\
- Passing a variable index where a level is required, especially in `cuddSwapInPlace()`. \\
- Passing adjacent levels in reverse order. Use `cuddSwapInPlace(table, curLevel - 1, curLevel)`, not `cuddSwapInPlace(table, curLevel, curLevel - 1)`. \\
- Calling `Cudd\_ReduceHeap()` or another nested top-level reorderer inside `cuddFunc()`. \\
- Returning without restoring the best order found.\\
- Mishandling move ownership: freeing `(Move *)CUDD\_OUT\_OF\_MEM`, freeing only the head after `ddUndoMoves(...)`, or double-freeing move lists.\\
- Building the wrong rollback order for plain sifting; use `down + inv`, not `inv + down`.\\

Do not: \\
- Swap outside `[lower, upper]`.\\
- Manually increment `table->ddTotalNumberSwapping`.\\
- Redefine `Move`, `IndexKey`, `ddMin`, `ddMax`, `cuddDynamicAllocNode`, or `cuddDeallocMove`.\\
- Use headers other than `util.h`, `cuddInt.h`, `math.h`, and `string.h`.
\end{promptbox}
\vspace{1mm}{\centering\small Prompt 3: External Knowledge\par}

\section{Full Per-Instance Results}
\label{app:full-results-per-instance}
\Cref{tab:full-stoch-revlib-a}--\Cref{tab:full-base-iscas-lgs} report the full per-function results for all 148 benchmark functions, grouped by the suite.
RevLib instances are listed in \Cref{tab:full-stoch-revlib-a}--\Cref{tab:full-base-revlib-b}.
ISCAS85/89 and LGSynth91 instances are listed in \Cref{tab:full-stoch-iscas-lgs} and~\Cref{tab:full-base-iscas-lgs} and denoted by superscripts \textsuperscript{I} and \textsuperscript{L}, respectively.
For ordering methods with stochastic components, we report best QCC, average QCC with its standard deviation, and average runtime; for BDD2Seq and deterministic methods, we report best QCC and runtime.

\paragraph{BDD2Seq runtime normalization.}
The runtimes reported for BDD2Seq include both variable ordering and reversible synthesis.
BDD2Seq(E*) synthesizes only its final ordering, whereas BDD2Seq(B*) uses beam search with a beam width of 20 and synthesizes all 20 candidate orderings.
We leave the reported BDD2Seq(E*) runtime unchanged because its single synthesis call accounts for only a small fraction of its total runtime.
For BDD2Seq(B*), we normalize only the estimated cost of the 19 additional CPU-based synthesis calls from the original dual Xeon~8375C platform to our single-thread Ryzen~9~5900X platform.
Let $T_{B^*}$ and $T_{E^*}$ denote the reported per-circuit runtimes on the original dual Xeon~8375C platform.
Assuming that $T_{B^*}-T_{E^*}$ corresponds to the 19 additional CPU-based synthesis calls, we rescale this component to our single-thread Ryzen~9~5900X platform using the PassMark single-thread ratio
$\gamma=2474/3465\approx0.714$:
\[
T_{B^*}^{\mathrm{adj}}
\approx
T_{E^*}
+
\frac{20\gamma-1}{19}
\left(T_{B^*}-T_{E^*}\right)
=
T_{E^*}
+
0.699\left(T_{B^*}-T_{E^*}\right).
\]
\par

\begin{table}[p]
\centering
\caption{Full benchmark results, RevLib functions, stochastic methods, part~1 of~2. Best and average QCC compared among QuantumEvo, GA, and SA; lowest values bold. Small values under Avg are $\pm$ one standard deviation of QCC across that function's 5 seeds.}
\label{tab:full-stoch-revlib-a}
\scriptsize
\setlength{\tabcolsep}{2.4pt}
\setlength{\extrarowheight}{0.35pt}
\renewcommand{\arraystretch}{0.86}
\resizebox{\linewidth}{!}{%
\begin{tabular}{lrr|rrr|rrr|rrr}
\toprule
Function & I & O & \multicolumn{3}{c|}{QuantumEvo} & \multicolumn{3}{c|}{GA} & \multicolumn{3}{c}{SA} \\
\cmidrule(lr){4-6}\cmidrule(lr){7-9}\cmidrule(lr){10-12}
 & & & Best & Avg $\pm$Std & Time & Best & Avg $\pm$Std & Time & Best & Avg $\pm$Std & Time \\
\midrule
\texttt{4gt10\_22} & 4 & 1 & \textbf{17} & \textbf{17.00}\,{\tiny$\pm$0} & 0.04 & \textbf{17} & \textbf{17.00}\,{\tiny$\pm$0} & 0.06 & \textbf{17} & \textbf{17.00}\,{\tiny$\pm$0} & 0.05 \\
\texttt{4gt11\_23} & 4 & 1 & \textbf{5} & \textbf{5.00}\,{\tiny$\pm$0} & 0.06 & \textbf{5} & \textbf{5.00}\,{\tiny$\pm$0} & 0.06 & \textbf{5} & \textbf{5.00}\,{\tiny$\pm$0} & 0.05 \\
\texttt{4gt12\_24} & 4 & 1 & \textbf{17} & \textbf{17.00}\,{\tiny$\pm$0} & 0.06 & \textbf{17} & \textbf{17.00}\,{\tiny$\pm$0} & 0.06 & \textbf{17} & \textbf{17.00}\,{\tiny$\pm$0} & 0.06 \\
\texttt{4gt13\_25} & 4 & 1 & \textbf{10} & \textbf{10.00}\,{\tiny$\pm$0} & 0.06 & \textbf{10} & \textbf{10.00}\,{\tiny$\pm$0} & 0.06 & \textbf{10} & \textbf{10.00}\,{\tiny$\pm$0} & 0.05 \\
\texttt{4gt4\_20} & 4 & 1 & \textbf{19} & \textbf{19.00}\,{\tiny$\pm$0} & 0.06 & \textbf{19} & \textbf{19.00}\,{\tiny$\pm$0} & 0.06 & \textbf{19} & \textbf{19.00}\,{\tiny$\pm$0} & 0.06 \\
\texttt{4gt5\_21} & 4 & 1 & \textbf{12} & \textbf{12.00}\,{\tiny$\pm$0} & 0.05 & \textbf{12} & \textbf{12.00}\,{\tiny$\pm$0} & 0.06 & \textbf{12} & \textbf{12.00}\,{\tiny$\pm$0} & 0.06 \\
\texttt{4mod5\_8} & 4 & 1 & \textbf{24} & \textbf{24.00}\,{\tiny$\pm$0} & 0.06 & \textbf{24} & \textbf{24.00}\,{\tiny$\pm$0} & 0.06 & \textbf{24} & \textbf{24.00}\,{\tiny$\pm$0} & 0.05 \\
\texttt{4mod7\_26} & 4 & 3 & \textbf{86} & \textbf{86.00}\,{\tiny$\pm$0} & 0.06 & \textbf{86} & \textbf{86.00}\,{\tiny$\pm$0} & 0.05 & \textbf{86} & \textbf{86.00}\,{\tiny$\pm$0} & 0.06 \\
\texttt{5xp1\_90} & 7 & 10 & \textbf{254} & \textbf{254.00}\,{\tiny$\pm$0} & 0.06 & 280 & 280.00\,{\tiny$\pm$0} & 0.06 & \textbf{254} & \textbf{254.00}\,{\tiny$\pm$0} & 0.07 \\
\texttt{9symml\_91} & 9 & 1 & \textbf{206} & \textbf{206.00}\,{\tiny$\pm$0} & 0.06 & \textbf{206} & \textbf{206.00}\,{\tiny$\pm$0} & 0.06 & \textbf{206} & \textbf{206.00}\,{\tiny$\pm$0} & 0.07 \\
\texttt{add6\_92} & 12 & 7 & \textbf{118} & \textbf{118.00}\,{\tiny$\pm$0} & 0.07 & \textbf{118} & \textbf{118.00}\,{\tiny$\pm$0} & 0.08 & \textbf{118} & 122.60\,{\tiny$\pm$9.20} & 0.09 \\
\texttt{adr4\_93} & 8 & 5 & \textbf{74} & \textbf{74.00}\,{\tiny$\pm$0} & 0.05 & \textbf{74} & \textbf{74.00}\,{\tiny$\pm$0} & 0.06 & \textbf{74} & \textbf{74.00}\,{\tiny$\pm$0} & 0.07 \\
\texttt{alu1\_94} & 12 & 8 & \textbf{139} & \textbf{139.00}\,{\tiny$\pm$0} & 0.06 & \textbf{139} & \textbf{139.00}\,{\tiny$\pm$0} & 0.07 & \textbf{139} & \textbf{139.00}\,{\tiny$\pm$0} & 0.07 \\
\texttt{alu2\_96} & 10 & 6 & \textbf{1266} & \textbf{1276.80}\,{\tiny$\pm$8.82} & 0.08 & 1269 & 1287.00\,{\tiny$\pm$11.22} & 0.09 & \textbf{1266} & 1278.80\,{\tiny$\pm$9.62} & 0.11 \\
\texttt{alu3\_97} & 10 & 8 & \textbf{430} & 434.80\,{\tiny$\pm$3.92} & 0.06 & \textbf{430} & \textbf{430.00}\,{\tiny$\pm$0} & 0.07 & \textbf{430} & 431.60\,{\tiny$\pm$3.20} & 0.10 \\
\texttt{alu4\_98} & 14 & 8 & 4385 & 4422.00\,{\tiny$\pm$30.35} & 0.19 & \textbf{4255} & \textbf{4341.20}\,{\tiny$\pm$47.48} & 0.20 & \textbf{4255} & 4361.40\,{\tiny$\pm$68.79} & 0.33 \\
\texttt{alu\_9} & 5 & 1 & \textbf{29} & \textbf{29.00}\,{\tiny$\pm$0} & 0.05 & \textbf{29} & \textbf{29.00}\,{\tiny$\pm$0} & 0.06 & \textbf{29} & \textbf{29.00}\,{\tiny$\pm$0} & 0.06 \\
\texttt{apex2\_101} & 39 & 3 & 2663 & \textbf{2697.20}\,{\tiny$\pm$25.46} & 0.62 & 2726 & 2929.00\,{\tiny$\pm$157.30} & 1.82 & \textbf{2660} & 2776.00\,{\tiny$\pm$101.95} & 0.89 \\
\texttt{apex4\_103} & 9 & 19 & 8236 & 8243.60\,{\tiny$\pm$3.83} & 0.13 & \textbf{8231} & \textbf{8235.80}\,{\tiny$\pm$4.96} & 0.12 & \textbf{8231} & 8240.40\,{\tiny$\pm$5.85} & 0.26 \\
\texttt{apex5\_104} & 117 & 88 & \textbf{9184} & 9812.40\,{\tiny$\pm$489.75} & 0.97 & 9551 & 10041.40\,{\tiny$\pm$392.32} & 1.24 & 9397 & \textbf{9809.00}\,{\tiny$\pm$448.99} & 1.97 \\
\texttt{apla\_107} & 10 & 12 & \textbf{731} & 734.60\,{\tiny$\pm$2.94} & 0.07 & \textbf{731} & \textbf{732.20}\,{\tiny$\pm$2.40} & 0.08 & \textbf{731} & \textbf{732.20}\,{\tiny$\pm$2.40} & 0.10 \\
\texttt{bw\_116} & 5 & 28 & \textbf{926} & \textbf{930.40}\,{\tiny$\pm$5.39} & 0.06 & \textbf{926} & 934.80\,{\tiny$\pm$4.40} & 0.06 & \textbf{926} & \textbf{930.40}\,{\tiny$\pm$5.39} & 0.08 \\
\texttt{C17\_117} & 5 & 2 & \textbf{37} & \textbf{37.00}\,{\tiny$\pm$0} & 0.06 & \textbf{37} & \textbf{37.00}\,{\tiny$\pm$0} & 0.06 & \textbf{37} & \textbf{37.00}\,{\tiny$\pm$0} & 0.06 \\
\texttt{C7552\_119} & 5 & 16 & \textbf{202} & \textbf{202.00}\,{\tiny$\pm$0} & 0.06 & \textbf{202} & \textbf{202.00}\,{\tiny$\pm$0} & 0.05 & \textbf{202} & \textbf{202.00}\,{\tiny$\pm$0} & 0.06 \\
\texttt{clip\_124} & 9 & 5 & 515 & 521.40\,{\tiny$\pm$4.59} & 0.06 & \textbf{512} & \textbf{517.80}\,{\tiny$\pm$6.27} & 0.07 & 519 & 522.20\,{\tiny$\pm$3.66} & 0.09 \\
\texttt{cm150a\_128} & 21 & 1 & \textbf{186} & \textbf{186.00}\,{\tiny$\pm$0} & 0.52 & \textbf{186} & \textbf{186.00}\,{\tiny$\pm$0} & 0.33 & \textbf{186} & \textbf{186.00}\,{\tiny$\pm$0} & 0.30 \\
\texttt{cm151a\_129} & 19 & 9 & \textbf{298} & \textbf{298.00}\,{\tiny$\pm$0} & 0.10 & \textbf{298} & \textbf{298.00}\,{\tiny$\pm$0} & 0.12 & \textbf{298} & \textbf{298.00}\,{\tiny$\pm$0} & 0.09 \\
\texttt{cm152a\_130} & 11 & 1 & \textbf{62} & \textbf{62.00}\,{\tiny$\pm$0} & 0.07 & \textbf{62} & \textbf{62.00}\,{\tiny$\pm$0} & 0.07 & \textbf{62} & \textbf{62.00}\,{\tiny$\pm$0} & 0.07 \\
\texttt{cm163a\_133} & 16 & 13 & \textbf{244} & 254.40\,{\tiny$\pm$10.52} & 0.08 & \textbf{244} & \textbf{253.20}\,{\tiny$\pm$11.27} & 0.09 & \textbf{244} & 261.20\,{\tiny$\pm$14.05} & 0.07 \\
\texttt{cm42a\_125} & 4 & 10 & \textbf{117} & \textbf{117.00}\,{\tiny$\pm$0} & 0.06 & \textbf{117} & \textbf{117.00}\,{\tiny$\pm$0} & 0.05 & \textbf{117} & \textbf{117.00}\,{\tiny$\pm$0} & 0.06 \\
\texttt{cm82a\_126} & 5 & 3 & \textbf{44} & \textbf{44.00}\,{\tiny$\pm$0} & 0.05 & \textbf{44} & \textbf{44.00}\,{\tiny$\pm$0} & 0.06 & \textbf{44} & \textbf{44.00}\,{\tiny$\pm$0} & 0.06 \\
\texttt{cm85a\_127} & 11 & 3 & \textbf{203} & \textbf{205.40}\,{\tiny$\pm$1.96} & 0.06 & 207 & 210.20\,{\tiny$\pm$1.60} & 0.07 & \textbf{203} & 208.20\,{\tiny$\pm$3.49} & 0.08 \\
\texttt{cmb\_134} & 16 & 4 & \textbf{153} & \textbf{153.00}\,{\tiny$\pm$0} & 0.08 & \textbf{153} & \textbf{153.00}\,{\tiny$\pm$0} & 0.11 & \textbf{153} & \textbf{153.00}\,{\tiny$\pm$0} & 0.09 \\
\texttt{co14\_135} & 14 & 1 & \textbf{159} & \textbf{159.00}\,{\tiny$\pm$0} & 0.08 & \textbf{159} & \textbf{159.00}\,{\tiny$\pm$0} & 0.10 & \textbf{159} & \textbf{159.00}\,{\tiny$\pm$0} & 0.08 \\
\texttt{con1\_136} & 7 & 2 & \textbf{89} & \textbf{90.20}\,{\tiny$\pm$2.40} & 0.06 & \textbf{89} & \textbf{90.20}\,{\tiny$\pm$2.40} & 0.06 & \textbf{89} & 93.80\,{\tiny$\pm$2.40} & 0.06 \\
\texttt{cordic\_138} & 23 & 2 & \textbf{304} & \textbf{308.00}\,{\tiny$\pm$3.74} & 0.16 & 306 & 312.00\,{\tiny$\pm$3.74} & 0.23 & 316 & 323.20\,{\tiny$\pm$3.60} & 0.17 \\
\texttt{cu\_141} & 14 & 11 & \textbf{219} & \textbf{219.80}\,{\tiny$\pm$0.40} & 0.08 & 224 & 224.00\,{\tiny$\pm$0} & 0.09 & 220 & 220.00\,{\tiny$\pm$0} & 0.09 \\
\texttt{dc1\_142} & 4 & 7 & \textbf{186} & \textbf{186.00}\,{\tiny$\pm$0} & 0.06 & \textbf{186} & \textbf{186.00}\,{\tiny$\pm$0} & 0.06 & \textbf{186} & \textbf{186.00}\,{\tiny$\pm$0} & 0.06 \\
\texttt{dc2\_143} & 8 & 7 & \textbf{431} & \textbf{431.00}\,{\tiny$\pm$0} & 0.06 & \textbf{431} & \textbf{431.00}\,{\tiny$\pm$0} & 0.06 & \textbf{431} & \textbf{431.00}\,{\tiny$\pm$0} & 0.07 \\
\texttt{decod24-enable\_32} & 3 & 4 & \textbf{38} & \textbf{38.00}\,{\tiny$\pm$0} & 0.06 & \textbf{38} & \textbf{38.00}\,{\tiny$\pm$0} & 0.05 & \textbf{38} & \textbf{38.00}\,{\tiny$\pm$0} & 0.06 \\
\texttt{decod\_137} & 5 & 16 & \textbf{202} & \textbf{202.00}\,{\tiny$\pm$0} & 0.05 & \textbf{202} & \textbf{202.00}\,{\tiny$\pm$0} & 0.05 & \textbf{202} & \textbf{202.00}\,{\tiny$\pm$0} & 0.06 \\
\texttt{dist\_144} & 8 & 5 & \textbf{975} & 975.80\,{\tiny$\pm$1.60} & 0.06 & 979 & 979.00\,{\tiny$\pm$0} & 0.07 & \textbf{975} & \textbf{975.00}\,{\tiny$\pm$0} & 0.09 \\
\texttt{dk17\_145} & 10 & 11 & \textbf{426} & \textbf{426.00}\,{\tiny$\pm$0} & 0.06 & \textbf{426} & \textbf{426.00}\,{\tiny$\pm$0} & 0.07 & \textbf{426} & \textbf{426.00}\,{\tiny$\pm$0} & 0.09 \\
\texttt{dk27\_146} & 9 & 9 & \textbf{140} & 140.80\,{\tiny$\pm$1.60} & 0.06 & 144 & 144.00\,{\tiny$\pm$0} & 0.06 & \textbf{140} & \textbf{140.00}\,{\tiny$\pm$0} & 0.06 \\
\texttt{e64\_149} & 65 & 65 & \textbf{886} & \textbf{886.00}\,{\tiny$\pm$0} & 0.39 & \textbf{886} & \textbf{886.00}\,{\tiny$\pm$0} & 1.14 & \textbf{886} & 890.20\,{\tiny$\pm$8.40} & 1.07 \\
\texttt{ex1010\_155} & 10 & 10 & 9681 & 9724.60\,{\tiny$\pm$33.76} & 0.23 & \textbf{9643} & \textbf{9703.20}\,{\tiny$\pm$54.16} & 0.22 & 9717 & 9803.60\,{\tiny$\pm$50.53} & 0.36 \\
\texttt{ex1\_150} & 5 & 1 & \textbf{8} & \textbf{8.00}\,{\tiny$\pm$0} & 0.06 & \textbf{8} & \textbf{8.00}\,{\tiny$\pm$0} & 0.05 & \textbf{8} & \textbf{8.00}\,{\tiny$\pm$0} & 0.06 \\
\texttt{ex2\_151} & 5 & 1 & \textbf{60} & \textbf{60.00}\,{\tiny$\pm$0} & 0.06 & \textbf{60} & \textbf{60.00}\,{\tiny$\pm$0} & 0.06 & \textbf{60} & 64.00\,{\tiny$\pm$4.90} & 0.06 \\
\bottomrule
\end{tabular}
}
\end{table}

\clearpage

\begin{table}[p]
\centering
\caption{Full benchmark results, RevLib functions, stochastic methods, part~2 of~2.}
\label{tab:full-stoch-revlib-b}
\scriptsize
\setlength{\tabcolsep}{2.4pt}
\setlength{\extrarowheight}{0.35pt}
\renewcommand{\arraystretch}{0.86}
\resizebox{\linewidth}{!}{%
\begin{tabular}{lrr|rrr|rrr|rrr}
\toprule
Function & I & O & \multicolumn{3}{c|}{QuantumEvo} & \multicolumn{3}{c|}{GA} & \multicolumn{3}{c}{SA} \\
\cmidrule(lr){4-6}\cmidrule(lr){7-9}\cmidrule(lr){10-12}
 & & & Best & Avg $\pm$Std & Time & Best & Avg $\pm$Std & Time & Best & Avg $\pm$Std & Time \\
\midrule
\texttt{ex3\_152} & 5 & 1 & \textbf{38} & \textbf{40.00}\,{\tiny$\pm$4} & 0.06 & \textbf{38} & 44.00\,{\tiny$\pm$4.90} & 0.06 & \textbf{38} & 45.00\,{\tiny$\pm$4} & 0.06 \\
\texttt{ex5p\_154} & 8 & 63 & \textbf{1843} & \textbf{1843.00}\,{\tiny$\pm$0} & 0.11 & \textbf{1843} & \textbf{1843.00}\,{\tiny$\pm$0} & 0.11 & \textbf{1843} & \textbf{1843.00}\,{\tiny$\pm$0} & 0.14 \\
\texttt{example2\_156} & 10 & 6 & \textbf{1266} & \textbf{1276.80}\,{\tiny$\pm$8.82} & 0.08 & 1269 & 1287.00\,{\tiny$\pm$11.22} & 0.09 & \textbf{1266} & 1278.80\,{\tiny$\pm$9.62} & 0.11 \\
\texttt{f2\_158} & 4 & 4 & \textbf{108} & 111.00\,{\tiny$\pm$2.45} & 0.05 & \textbf{108} & \textbf{108.00}\,{\tiny$\pm$0} & 0.05 & 113 & 113.00\,{\tiny$\pm$0} & 0.05 \\
\texttt{f51m\_159} & 14 & 8 & \textbf{2422} & \textbf{2429.20}\,{\tiny$\pm$4.21} & 0.14 & 2427 & 2429.40\,{\tiny$\pm$2.94} & 0.20 & \textbf{2422} & 2430.60\,{\tiny$\pm$4.32} & 0.25 \\
\texttt{frg1\_160} & 28 & 3 & \textbf{559} & \textbf{597.40}\,{\tiny$\pm$38.66} & 0.17 & 560 & 610.20\,{\tiny$\pm$41.06} & 0.24 & 567 & 637.40\,{\tiny$\pm$67.22} & 0.17 \\
\texttt{frg2\_161} & 143 & 139 & 7017 & \textbf{7601.60}\,{\tiny$\pm$506.67} & 1.47 & 9429 & 9429.00\,{\tiny$\pm$0} & 1.63 & \textbf{6972} & 7689.80\,{\tiny$\pm$593.33} & 3.22 \\
\texttt{in0\_162} & 15 & 11 & 2299 & 2313.40\,{\tiny$\pm$14.44} & 0.11 & 2299 & 2317.40\,{\tiny$\pm$13.53} & 0.16 & \textbf{2271} & \textbf{2295.80}\,{\tiny$\pm$12.50} & 0.19 \\
\texttt{inc\_170} & 7 & 9 & \textbf{592} & \textbf{592.00}\,{\tiny$\pm$0} & 0.06 & \textbf{592} & \textbf{592.00}\,{\tiny$\pm$0} & 0.06 & \textbf{592} & \textbf{592.00}\,{\tiny$\pm$0} & 0.08 \\
\texttt{life\_175} & 9 & 1 & \textbf{204} & \textbf{204.00}\,{\tiny$\pm$0} & 0.06 & 210 & 210.00\,{\tiny$\pm$0} & 0.07 & \textbf{204} & \textbf{204.00}\,{\tiny$\pm$0} & 0.07 \\
\texttt{majority\_176} & 5 & 1 & \textbf{41} & \textbf{41.00}\,{\tiny$\pm$0} & 0.06 & \textbf{41} & \textbf{41.00}\,{\tiny$\pm$0} & 0.05 & \textbf{41} & \textbf{41.00}\,{\tiny$\pm$0} & 0.06 \\
\texttt{max46\_177} & 9 & 1 & 556 & 562.80\,{\tiny$\pm$8.89} & 0.07 & 546 & 558.00\,{\tiny$\pm$6.78} & 0.07 & \textbf{545} & \textbf{552.60}\,{\tiny$\pm$4.22} & 0.09 \\
\texttt{misex1\_178} & 8 & 7 & \textbf{279} & 290.40\,{\tiny$\pm$9.31} & 0.06 & 287 & \textbf{287.00}\,{\tiny$\pm$0} & 0.06 & 288 & 288.00\,{\tiny$\pm$0} & 0.07 \\
\texttt{misex3\_180} & 14 & 14 & \textbf{3789} & \textbf{3791.40}\,{\tiny$\pm$2.94} & 0.16 & \textbf{3789} & 3792.60\,{\tiny$\pm$2.94} & 0.22 & \textbf{3789} & 3833.40\,{\tiny$\pm$52.29} & 0.35 \\
\texttt{misex3c\_181} & 14 & 14 & \textbf{3936} & \textbf{3950.40}\,{\tiny$\pm$10.80} & 0.16 & \textbf{3936} & \textbf{3950.40}\,{\tiny$\pm$10.80} & 0.24 & \textbf{3936} & 3979.60\,{\tiny$\pm$73.78} & 0.32 \\
\texttt{mlp4\_184} & 8 & 8 & \textbf{1158} & 1158.40\,{\tiny$\pm$0.49} & 0.07 & 1159 & 1159.00\,{\tiny$\pm$0} & 0.07 & \textbf{1158} & \textbf{1158.00}\,{\tiny$\pm$0} & 0.10 \\
\texttt{mux\_185} & 21 & 1 & \textbf{170} & \textbf{170.00}\,{\tiny$\pm$0} & 0.55 & \textbf{170} & \textbf{170.00}\,{\tiny$\pm$0} & 0.41 & \textbf{170} & \textbf{170.00}\,{\tiny$\pm$0} & 0.41 \\
\texttt{one-two-three\_27} & 3 & 3 & \textbf{44} & \textbf{44.00}\,{\tiny$\pm$0} & 0.06 & \textbf{44} & \textbf{44.00}\,{\tiny$\pm$0} & 0.06 & \textbf{44} & \textbf{44.00}\,{\tiny$\pm$0} & 0.05 \\
\texttt{parity\_188} & 16 & 1 & \textbf{31} & \textbf{31.00}\,{\tiny$\pm$0} & 0.50 & \textbf{31} & \textbf{31.00}\,{\tiny$\pm$0} & 0.49 & \textbf{31} & \textbf{31.00}\,{\tiny$\pm$0} & 0.49 \\
\texttt{pcler8\_190} & 16 & 5 & \textbf{124} & \textbf{124.00}\,{\tiny$\pm$0} & 0.08 & \textbf{124} & \textbf{124.00}\,{\tiny$\pm$0} & 0.08 & \textbf{124} & \textbf{124.00}\,{\tiny$\pm$0} & 0.08 \\
\texttt{pdc\_191} & 16 & 40 & \textbf{6599} & \textbf{6599.00}\,{\tiny$\pm$0} & 0.41 & \textbf{6599} & \textbf{6599.00}\,{\tiny$\pm$0} & 0.60 & \textbf{6599} & \textbf{6599.00}\,{\tiny$\pm$0} & 0.84 \\
\texttt{pm1\_192} & 4 & 10 & \textbf{117} & \textbf{117.00}\,{\tiny$\pm$0} & 0.06 & \textbf{117} & \textbf{117.00}\,{\tiny$\pm$0} & 0.06 & \textbf{117} & \textbf{117.00}\,{\tiny$\pm$0} & 0.06 \\
\texttt{radd\_193} & 8 & 5 & \textbf{74} & \textbf{74.00}\,{\tiny$\pm$0} & 0.06 & \textbf{74} & \textbf{74.00}\,{\tiny$\pm$0} & 0.06 & \textbf{74} & \textbf{74.00}\,{\tiny$\pm$0} & 0.06 \\
\texttt{rd53\_68} & 5 & 3 & \textbf{98} & \textbf{98.00}\,{\tiny$\pm$0} & 0.05 & \textbf{98} & \textbf{98.00}\,{\tiny$\pm$0} & 0.06 & \textbf{98} & \textbf{98.00}\,{\tiny$\pm$0} & 0.05 \\
\texttt{rd73\_69} & 7 & 3 & \textbf{217} & \textbf{217.00}\,{\tiny$\pm$0} & 0.06 & \textbf{217} & \textbf{217.00}\,{\tiny$\pm$0} & 0.06 & \textbf{217} & \textbf{217.00}\,{\tiny$\pm$0} & 0.06 \\
\texttt{rd84\_70} & 8 & 4 & \textbf{304} & \textbf{304.00}\,{\tiny$\pm$0} & 0.06 & \textbf{304} & \textbf{304.00}\,{\tiny$\pm$0} & 0.06 & \textbf{304} & \textbf{304.00}\,{\tiny$\pm$0} & 0.07 \\
\texttt{root\_197} & 8 & 5 & \textbf{444} & \textbf{444.00}\,{\tiny$\pm$0} & 0.06 & 446 & 446.00\,{\tiny$\pm$0} & 0.06 & \textbf{444} & \textbf{444.00}\,{\tiny$\pm$0} & 0.08 \\
\texttt{ryy6\_198} & 16 & 1 & \textbf{107} & \textbf{109.40}\,{\tiny$\pm$4.80} & 0.09 & \textbf{107} & 112.60\,{\tiny$\pm$4.80} & 0.11 & \textbf{107} & 117.20\,{\tiny$\pm$5.23} & 0.09 \\
\texttt{sao2\_199} & 10 & 4 & \textbf{653} & \textbf{653.00}\,{\tiny$\pm$0} & 0.07 & \textbf{653} & 654.80\,{\tiny$\pm$2.23} & 0.07 & \textbf{653} & 655.80\,{\tiny$\pm$2.32} & 0.11 \\
\texttt{seq\_201} & 41 & 35 & \textbf{9080} & 9151.00\,{\tiny$\pm$44.62} & 1.17 & 9113 & \textbf{9142.00}\,{\tiny$\pm$26.48} & 3.03 & 9165 & 9237.40\,{\tiny$\pm$96.16} & 1.99 \\
\texttt{sf\_232} & 4 & 1 & \textbf{21} & \textbf{26.80}\,{\tiny$\pm$3.25} & 0.06 & 43 & 43.00\,{\tiny$\pm$0} & 0.06 & 36 & 36.00\,{\tiny$\pm$0} & 0.06 \\
\texttt{spla\_202} & 16 & 46 & \textbf{5858} & \textbf{5860.00}\,{\tiny$\pm$2.45} & 0.33 & \textbf{5858} & \textbf{5860.00}\,{\tiny$\pm$2.45} & 0.35 & \textbf{5858} & 5861.00\,{\tiny$\pm$2.45} & 0.56 \\
\texttt{sqn\_203} & 7 & 3 & \textbf{356} & 358.00\,{\tiny$\pm$2.45} & 0.06 & \textbf{356} & 358.00\,{\tiny$\pm$2.45} & 0.06 & \textbf{356} & \textbf{357.00}\,{\tiny$\pm$2} & 0.07 \\
\texttt{sqr6\_204} & 6 & 12 & 486 & 486.00\,{\tiny$\pm$0} & 0.06 & \textbf{482} & \textbf{482.00}\,{\tiny$\pm$0} & 0.06 & 486 & 486.00\,{\tiny$\pm$0} & 0.07 \\
\texttt{sqrt8\_205} & 8 & 4 & \textbf{221} & 233.80\,{\tiny$\pm$6.40} & 0.06 & \textbf{221} & \textbf{230.60}\,{\tiny$\pm$7.84} & 0.06 & \textbf{221} & \textbf{230.60}\,{\tiny$\pm$7.84} & 0.07 \\
\texttt{squar5\_206} & 5 & 8 & \textbf{232} & \textbf{232.00}\,{\tiny$\pm$0} & 0.06 & \textbf{232} & \textbf{232.00}\,{\tiny$\pm$0} & 0.05 & \textbf{232} & \textbf{232.00}\,{\tiny$\pm$0} & 0.06 \\
\texttt{sym10\_207} & 10 & 1 & \textbf{253} & \textbf{253.00}\,{\tiny$\pm$0} & 0.07 & \textbf{253} & \textbf{253.00}\,{\tiny$\pm$0} & 0.08 & \textbf{253} & \textbf{253.00}\,{\tiny$\pm$0} & 0.07 \\
\texttt{sym6\_63} & 6 & 1 & \textbf{93} & \textbf{93.00}\,{\tiny$\pm$0} & 0.06 & \textbf{93} & \textbf{93.00}\,{\tiny$\pm$0} & 0.06 & \textbf{93} & \textbf{93.00}\,{\tiny$\pm$0} & 0.06 \\
\texttt{sym9\_71} & 9 & 1 & \textbf{206} & \textbf{206.00}\,{\tiny$\pm$0} & 0.06 & \textbf{206} & \textbf{206.00}\,{\tiny$\pm$0} & 0.07 & \textbf{206} & \textbf{206.00}\,{\tiny$\pm$0} & 0.06 \\
\texttt{t481\_208} & 16 & 1 & \textbf{147} & \textbf{147.00}\,{\tiny$\pm$0} & 0.10 & \textbf{147} & \textbf{147.00}\,{\tiny$\pm$0} & 0.12 & \textbf{147} & \textbf{147.00}\,{\tiny$\pm$0} & 0.09 \\
\texttt{table3\_209} & 14 & 14 & \textbf{5898} & \textbf{5899.40}\,{\tiny$\pm$2.80} & 0.18 & \textbf{5898} & 5902.20\,{\tiny$\pm$3.43} & 0.22 & \textbf{5898} & 5908.80\,{\tiny$\pm$21.60} & 0.43 \\
\texttt{tial\_214} & 14 & 8 & 4460 & 4487.60\,{\tiny$\pm$20.76} & 0.13 & 4354 & 4401.00\,{\tiny$\pm$30.13} & 0.21 & \textbf{4231} & \textbf{4299.20}\,{\tiny$\pm$47.13} & 0.38 \\
\texttt{urf4\_89} & 11 & 11 & \textbf{28406} & \textbf{28537.40}\,{\tiny$\pm$101.35} & 0.41 & 28538 & 28591.80\,{\tiny$\pm$36.85} & 0.54 & 28489 & 28590.80\,{\tiny$\pm$80.86} & 0.97 \\
\texttt{wim\_220} & 4 & 7 & \textbf{107} & \textbf{107.00}\,{\tiny$\pm$0} & 0.06 & \textbf{107} & \textbf{107.00}\,{\tiny$\pm$0} & 0.06 & \textbf{107} & \textbf{107.00}\,{\tiny$\pm$0} & 0.05 \\
\texttt{x2\_223} & 10 & 7 & \textbf{191} & \textbf{191.00}\,{\tiny$\pm$0} & 0.06 & \textbf{191} & \textbf{191.00}\,{\tiny$\pm$0} & 0.07 & \textbf{191} & \textbf{191.00}\,{\tiny$\pm$0} & 0.08 \\
\texttt{xor5\_195} & 5 & 1 & \textbf{8} & \textbf{8.00}\,{\tiny$\pm$0} & 0.05 & \textbf{8} & \textbf{8.00}\,{\tiny$\pm$0} & 0.06 & \textbf{8} & \textbf{8.00}\,{\tiny$\pm$0} & 0.06 \\
\texttt{z4\_224} & 7 & 4 & \textbf{66} & \textbf{66.00}\,{\tiny$\pm$0} & 0.06 & \textbf{66} & \textbf{66.00}\,{\tiny$\pm$0} & 0.05 & \textbf{66} & \textbf{66.00}\,{\tiny$\pm$0} & 0.06 \\
\texttt{z4ml\_225} & 7 & 4 & \textbf{66} & \textbf{66.00}\,{\tiny$\pm$0} & 0.05 & \textbf{66} & \textbf{66.00}\,{\tiny$\pm$0} & 0.06 & \textbf{66} & \textbf{66.00}\,{\tiny$\pm$0} & 0.06 \\
\bottomrule
\end{tabular}
}
\end{table}

\clearpage

\begin{table}[p]
\centering
\caption{Full benchmark results, ISCAS85/89 and LGSynth91 functions, stochastic methods.}
\label{tab:full-stoch-iscas-lgs}
\scriptsize
\setlength{\tabcolsep}{2.4pt}
\setlength{\extrarowheight}{0.35pt}
\renewcommand{\arraystretch}{0.86}
\resizebox{\linewidth}{!}{%
\begin{tabular}{lrr|rrr|rrr|rrr}
\toprule
Function & I & O & \multicolumn{3}{c|}{QuantumEvo} & \multicolumn{3}{c|}{GA} & \multicolumn{3}{c}{SA} \\
\cmidrule(lr){4-6}\cmidrule(lr){7-9}\cmidrule(lr){10-12}
 & & & Best & Avg $\pm$Std & Time & Best & Avg $\pm$Std & Time & Best & Avg $\pm$Std & Time \\
\midrule
\texttt{apex6\_orig}\textsuperscript{L} & 135 & 99 & \textbf{3736} & \textbf{3787.20}\,{\tiny$\pm$36.31} & 0.65 & 4159 & 4368.40\,{\tiny$\pm$169.77} & 0.64 & 4392 & 4551.40\,{\tiny$\pm$148.45} & 1.08 \\
\texttt{apex7\_orig}\textsuperscript{L} & 49 & 37 & \textbf{1431} & \textbf{1440.60}\,{\tiny$\pm$7.45} & 0.28 & 1441 & 1443.40\,{\tiny$\pm$2.94} & 0.58 & 1703 & 1759.20\,{\tiny$\pm$46.63} & 0.44 \\
\texttt{b1\_orig}\textsuperscript{L} & 3 & 4 & \textbf{22} & 23.00\,{\tiny$\pm$2} & 0.06 & \textbf{22} & \textbf{22.00}\,{\tiny$\pm$0} & 0.05 & \textbf{22} & \textbf{22.00}\,{\tiny$\pm$0} & 0.06 \\
\texttt{b9\_orig}\textsuperscript{L} & 41 & 21 & \textbf{713} & 719.40\,{\tiny$\pm$6.59} & 0.17 & 715 & \textbf{719.00}\,{\tiny$\pm$3.58} & 0.27 & 804 & 804.00\,{\tiny$\pm$0} & 0.19 \\
\texttt{c17}\textsuperscript{I} & 5 & 2 & \textbf{37} & \textbf{37.00}\,{\tiny$\pm$0} & 0.06 & \textbf{37} & \textbf{37.00}\,{\tiny$\pm$0} & 0.06 & \textbf{37} & \textbf{37.00}\,{\tiny$\pm$0} & 0.06 \\
\texttt{c432}\textsuperscript{I} & 36 & 7 & \textbf{10494} & \textbf{11053.20}\,{\tiny$\pm$305.36} & 245.19 & 11141 & 11170.60\,{\tiny$\pm$40.35} & 6.20 & 11101 & 11148.60\,{\tiny$\pm$95.20} & 1.46 \\
\texttt{c880}\textsuperscript{I} & 60 & 26 & \textbf{34241} & 34761.40\,{\tiny$\pm$367.34} & 329.32 & \textbf{34241} & 35095.20\,{\tiny$\pm$528.82} & 44.13 & \textbf{34241} & \textbf{34610.60}\,{\tiny$\pm$321.65} & 12.39 \\
\texttt{c8\_orig}\textsuperscript{L} & 28 & 18 & \textbf{436} & \textbf{436.00}\,{\tiny$\pm$0} & 0.14 & \textbf{436} & 446.40\,{\tiny$\pm$6.97} & 0.18 & \textbf{436} & 450.80\,{\tiny$\pm$11.65} & 0.15 \\
\texttt{cc\_orig}\textsuperscript{L} & 21 & 20 & 283 & 289.00\,{\tiny$\pm$5.83} & 0.08 & \textbf{267} & \textbf{280.00}\,{\tiny$\pm$8.12} & 0.11 & 278 & 287.20\,{\tiny$\pm$5.15} & 0.08 \\
\texttt{cht\_orig}\textsuperscript{L} & 47 & 36 & \textbf{714} & \textbf{714.00}\,{\tiny$\pm$0} & 0.12 & \textbf{714} & \textbf{714.00}\,{\tiny$\pm$0} & 0.21 & \textbf{714} & \textbf{714.00}\,{\tiny$\pm$0} & 0.14 \\
\texttt{cm138a\_orig}\textsuperscript{L} & 6 & 8 & \textbf{104} & \textbf{104.00}\,{\tiny$\pm$0} & 0.06 & \textbf{104} & \textbf{104.00}\,{\tiny$\pm$0} & 0.06 & \textbf{104} & \textbf{104.00}\,{\tiny$\pm$0} & 0.06 \\
\texttt{cm150a\_orig}\textsuperscript{L} & 21 & 1 & \textbf{186} & \textbf{186.00}\,{\tiny$\pm$0} & 0.52 & \textbf{186} & \textbf{186.00}\,{\tiny$\pm$0} & 0.45 & \textbf{186} & \textbf{186.00}\,{\tiny$\pm$0} & 0.36 \\
\texttt{cm151a\_orig}\textsuperscript{L} & 12 & 2 & \textbf{92} & \textbf{92.00}\,{\tiny$\pm$0} & 0.07 & \textbf{92} & \textbf{92.00}\,{\tiny$\pm$0} & 0.08 & \textbf{92} & \textbf{92.00}\,{\tiny$\pm$0} & 0.07 \\
\texttt{cm152a\_orig}\textsuperscript{L} & 11 & 1 & \textbf{62} & \textbf{62.00}\,{\tiny$\pm$0} & 0.07 & \textbf{62} & \textbf{62.00}\,{\tiny$\pm$0} & 0.07 & \textbf{62} & \textbf{62.00}\,{\tiny$\pm$0} & 0.07 \\
\texttt{cm162a\_orig}\textsuperscript{L} & 14 & 5 & 222 & 222.00\,{\tiny$\pm$0} & 0.08 & \textbf{219} & \textbf{220.20}\,{\tiny$\pm$1.47} & 0.09 & \textbf{219} & 220.80\,{\tiny$\pm$1.47} & 0.09 \\
\texttt{cm163a\_orig}\textsuperscript{L} & 16 & 5 & \textbf{124} & \textbf{124.00}\,{\tiny$\pm$0} & 0.08 & \textbf{124} & \textbf{124.00}\,{\tiny$\pm$0} & 0.08 & \textbf{124} & \textbf{124.00}\,{\tiny$\pm$0} & 0.07 \\
\texttt{cm42a\_orig}\textsuperscript{L} & 4 & 10 & \textbf{117} & \textbf{117.00}\,{\tiny$\pm$0} & 0.05 & \textbf{117} & \textbf{117.00}\,{\tiny$\pm$0} & 0.05 & \textbf{117} & \textbf{117.00}\,{\tiny$\pm$0} & 0.06 \\
\texttt{cm82a\_orig}\textsuperscript{L} & 5 & 3 & \textbf{44} & \textbf{44.00}\,{\tiny$\pm$0} & 0.06 & \textbf{44} & \textbf{44.00}\,{\tiny$\pm$0} & 0.06 & \textbf{44} & \textbf{44.00}\,{\tiny$\pm$0} & 0.06 \\
\texttt{cm85a\_orig}\textsuperscript{L} & 11 & 3 & \textbf{203} & \textbf{205.40}\,{\tiny$\pm$1.96} & 0.07 & 207 & 210.20\,{\tiny$\pm$1.60} & 0.07 & \textbf{203} & 208.20\,{\tiny$\pm$3.49} & 0.08 \\
\texttt{cmb\_orig}\textsuperscript{L} & 16 & 4 & \textbf{153} & \textbf{153.00}\,{\tiny$\pm$0} & 0.09 & \textbf{153} & \textbf{153.00}\,{\tiny$\pm$0} & 0.11 & \textbf{153} & \textbf{153.00}\,{\tiny$\pm$0} & 0.09 \\
\texttt{comp\_orig}\textsuperscript{L} & 32 & 3 & \textbf{796} & \textbf{796.00}\,{\tiny$\pm$0} & 2.41 & 824 & 824.00\,{\tiny$\pm$0} & 2.28 & 897 & 936.00\,{\tiny$\pm$22.08} & 2.01 \\
\texttt{cordic\_orig}\textsuperscript{L} & 23 & 2 & \textbf{304} & \textbf{308.00}\,{\tiny$\pm$3.74} & 0.12 & 306 & 312.00\,{\tiny$\pm$3.74} & 0.20 & 316 & 323.20\,{\tiny$\pm$3.60} & 0.14 \\
\texttt{count\_orig}\textsuperscript{L} & 35 & 16 & \textbf{441} & \textbf{441.00}\,{\tiny$\pm$0} & 0.17 & \textbf{441} & \textbf{441.00}\,{\tiny$\pm$0} & 0.29 & \textbf{441} & \textbf{441.00}\,{\tiny$\pm$0} & 0.21 \\
\texttt{cu\_orig}\textsuperscript{L} & 14 & 11 & \textbf{219} & \textbf{219.80}\,{\tiny$\pm$0.40} & 0.08 & 224 & 224.00\,{\tiny$\pm$0} & 0.09 & 220 & 220.00\,{\tiny$\pm$0} & 0.09 \\
\texttt{dalu\_orig}\textsuperscript{L} & 75 & 16 & \textbf{5230} & 5301.60\,{\tiny$\pm$66.64} & 71.73 & 5240 & \textbf{5291.20}\,{\tiny$\pm$36.74} & 71.72 & 5316 & 5663.20\,{\tiny$\pm$571.35} & 69.72 \\
\texttt{decod\_orig}\textsuperscript{L} & 5 & 16 & \textbf{202} & \textbf{202.00}\,{\tiny$\pm$0} & 0.06 & \textbf{202} & \textbf{202.00}\,{\tiny$\pm$0} & 0.06 & \textbf{202} & \textbf{202.00}\,{\tiny$\pm$0} & 0.06 \\
\texttt{example2\_orig}\textsuperscript{L} & 85 & 66 & 1790 & 1844.40\,{\tiny$\pm$48.38} & 0.28 & \textbf{1784} & \textbf{1835.80}\,{\tiny$\pm$63.21} & 0.54 & 1894 & 1954.00\,{\tiny$\pm$59.14} & 0.45 \\
\texttt{frg1\_orig}\textsuperscript{L} & 28 & 3 & \textbf{559} & \textbf{597.40}\,{\tiny$\pm$38.66} & 0.15 & 560 & 610.20\,{\tiny$\pm$41.06} & 0.27 & 567 & 637.40\,{\tiny$\pm$67.22} & 0.17 \\
\texttt{frg2\_orig}\textsuperscript{L} & 143 & 139 & 7017 & \textbf{7601.60}\,{\tiny$\pm$506.67} & 1.44 & 9429 & 9429.00\,{\tiny$\pm$0} & 1.69 & \textbf{6972} & 7689.80\,{\tiny$\pm$593.33} & 3.08 \\
\texttt{k2\_orig}\textsuperscript{L} & 45 & 45 & \textbf{10156} & \textbf{10212.60}\,{\tiny$\pm$50.19} & 0.97 & 10200 & 10265.40\,{\tiny$\pm$50.28} & 1.78 & 10276 & 10410.00\,{\tiny$\pm$93.77} & 1.49 \\
\texttt{lal\_orig}\textsuperscript{L} & 26 & 19 & 496 & 496.00\,{\tiny$\pm$0} & 0.10 & \textbf{478} & \textbf{490.80}\,{\tiny$\pm$11.07} & 0.14 & 484 & 507.80\,{\tiny$\pm$16.90} & 0.12 \\
\texttt{majority\_orig}\textsuperscript{L} & 5 & 1 & \textbf{41} & \textbf{41.00}\,{\tiny$\pm$0} & 0.06 & \textbf{41} & \textbf{41.00}\,{\tiny$\pm$0} & 0.06 & \textbf{41} & \textbf{41.00}\,{\tiny$\pm$0} & 0.05 \\
\texttt{mux\_orig}\textsuperscript{L} & 21 & 1 & \textbf{170} & \textbf{170.00}\,{\tiny$\pm$0} & 0.52 & \textbf{170} & \textbf{170.00}\,{\tiny$\pm$0} & 0.39 & \textbf{170} & \textbf{170.00}\,{\tiny$\pm$0} & 0.40 \\
\texttt{my\_adder\_orig}\textsuperscript{L} & 33 & 17 & \textbf{352} & \textbf{352.00}\,{\tiny$\pm$0} & 1.24 & \textbf{352} & \textbf{352.00}\,{\tiny$\pm$0} & 0.66 & \textbf{352} & \textbf{352.00}\,{\tiny$\pm$0} & 0.37 \\
\texttt{pair\_orig}\textsuperscript{L} & 173 & 137 & \textbf{20236} & \textbf{22344.20}\,{\tiny$\pm$1371.13} & 6.24 & 47110 & 602976.40\,{\tiny$\pm$454972.53} & 8.28 & 20729 & 22438.00\,{\tiny$\pm$1928.73} & 20.16 \\
\texttt{parity\_orig}\textsuperscript{L} & 16 & 1 & \textbf{31} & \textbf{31.00}\,{\tiny$\pm$0} & 0.09 & \textbf{31} & \textbf{31.00}\,{\tiny$\pm$0} & 0.11 & \textbf{31} & \textbf{31.00}\,{\tiny$\pm$0} & 0.06 \\
\texttt{pcle\_orig}\textsuperscript{L} & 19 & 9 & \textbf{298} & \textbf{298.00}\,{\tiny$\pm$0} & 0.09 & \textbf{298} & \textbf{298.00}\,{\tiny$\pm$0} & 0.11 & \textbf{298} & \textbf{298.00}\,{\tiny$\pm$0} & 0.10 \\
\texttt{pcler8\_orig}\textsuperscript{L} & 27 & 17 & 639 & 647.00\,{\tiny$\pm$4} & 0.15 & 674 & 674.00\,{\tiny$\pm$0} & 0.23 & \textbf{638} & \textbf{638.00}\,{\tiny$\pm$0} & 0.16 \\
\texttt{pm1\_orig}\textsuperscript{L} & 16 & 13 & \textbf{244} & 254.40\,{\tiny$\pm$10.52} & 0.08 & \textbf{244} & \textbf{253.20}\,{\tiny$\pm$11.27} & 0.09 & \textbf{244} & 261.20\,{\tiny$\pm$14.05} & 0.08 \\
\texttt{rot\_orig}\textsuperscript{L} & 135 & 107 & \textbf{23926} & 26838.00\,{\tiny$\pm$2004.99} & 845.19 & 5928978 & 9069721.00\,{\tiny$\pm$2622975.41} & 62.87 & 24124 & \textbf{25464.60}\,{\tiny$\pm$1500} & 25.99 \\
\texttt{s1196\_orig}\textsuperscript{I} & 14 & 14 & \textbf{3765} & \textbf{3769.40}\,{\tiny$\pm$5.39} & 0.14 & \textbf{3765} & 3771.60\,{\tiny$\pm$5.39} & 0.18 & \textbf{3765} & 3816.80\,{\tiny$\pm$59.21} & 0.29 \\
\texttt{sct\_orig}\textsuperscript{L} & 19 & 15 & 375 & 386.80\,{\tiny$\pm$12.62} & 0.10 & 373 & \textbf{375.80}\,{\tiny$\pm$2.79} & 0.12 & \textbf{369} & 382.40\,{\tiny$\pm$14.07} & 0.11 \\
\texttt{tcon\_orig}\textsuperscript{L} & 17 & 16 & \textbf{88} & \textbf{88.00}\,{\tiny$\pm$0} & 0.06 & \textbf{88} & \textbf{88.00}\,{\tiny$\pm$0} & 0.07 & \textbf{88} & \textbf{88.00}\,{\tiny$\pm$0} & 0.06 \\
\texttt{term1\_orig}\textsuperscript{L} & 34 & 10 & 497 & 500.80\,{\tiny$\pm$3.12} & 0.18 & \textbf{473} & \textbf{489.60}\,{\tiny$\pm$9.33} & 0.33 & 563 & 639.80\,{\tiny$\pm$114.23} & 0.25 \\
\texttt{too\_large\_orig}\textsuperscript{L} & 38 & 3 & 2676 & \textbf{2736.40}\,{\tiny$\pm$41.96} & 0.71 & \textbf{2675} & 2830.80\,{\tiny$\pm$122.28} & 1.83 & 2694 & 2816.60\,{\tiny$\pm$154.97} & 0.90 \\
\texttt{ttt2\_orig}\textsuperscript{L} & 24 & 21 & 736 & 738.80\,{\tiny$\pm$3.43} & 0.13 & \textbf{711} & \textbf{727.20}\,{\tiny$\pm$9.70} & 0.16 & 727 & 730.20\,{\tiny$\pm$6.40} & 0.15 \\
\texttt{unreg\_orig}\textsuperscript{L} & 36 & 16 & \textbf{494} & 494.40\,{\tiny$\pm$0.49} & 0.12 & 495 & 495.00\,{\tiny$\pm$0} & 0.19 & \textbf{494} & \textbf{494.00}\,{\tiny$\pm$0} & 0.11 \\
\texttt{vda\_orig}\textsuperscript{L} & 17 & 39 & 4169 & 4181.00\,{\tiny$\pm$9.80} & 0.16 & 4169 & 4177.00\,{\tiny$\pm$9.80} & 0.26 & \textbf{4107} & \textbf{4176.20}\,{\tiny$\pm$50.98} & 0.28 \\
\texttt{x1\_orig}\textsuperscript{L} & 51 & 35 & 3062 & 3133.60\,{\tiny$\pm$60.90} & 0.50 & \textbf{3030} & \textbf{3089.80}\,{\tiny$\pm$60.19} & 1.06 & 3144 & 3308.40\,{\tiny$\pm$192.40} & 0.76 \\
\texttt{x2\_orig}\textsuperscript{L} & 10 & 7 & \textbf{191} & \textbf{191.00}\,{\tiny$\pm$0} & 0.06 & \textbf{191} & \textbf{191.00}\,{\tiny$\pm$0} & 0.07 & \textbf{191} & \textbf{191.00}\,{\tiny$\pm$0} & 0.08 \\
\texttt{x3\_orig}\textsuperscript{L} & 135 & 99 & \textbf{3736} & \textbf{3787.20}\,{\tiny$\pm$36.31} & 0.62 & 4159 & 4368.40\,{\tiny$\pm$169.77} & 0.66 & 4392 & 4551.40\,{\tiny$\pm$148.45} & 1.15 \\
\texttt{x4\_orig}\textsuperscript{L} & 94 & 71 & \textbf{2437} & \textbf{2513.60}\,{\tiny$\pm$87.43} & 0.37 & 2505 & 2809.40\,{\tiny$\pm$231.03} & 0.65 & 2738 & 2863.20\,{\tiny$\pm$109.57} & 0.69 \\
\bottomrule
\end{tabular}
}
\par\smallskip{\footnotesize\raggedright Superscripts denote source dataset: \textsuperscript{I}~ISCAS85/89; \textsuperscript{L}~LGSynth91.\par}
\end{table}

\clearpage

\begin{table}[p]
\centering
\caption{Full benchmark results, RevLib functions, prior baselines, part~1 of~2. Best QCC compared among QuantumEvo, BDD2Seq, and deterministic CUDD baselines; lowest values bold. BDD2Seq(B*) Time is hardware-adjusted to our platform; BDD2Seq(E*) Time is as originally reported.}
\label{tab:full-base-revlib-a}
\scriptsize
\setlength{\tabcolsep}{2.4pt}
\setlength{\extrarowheight}{0.35pt}
\renewcommand{\arraystretch}{0.86}
\resizebox{\linewidth}{!}{%
\begin{tabular}{l|rr|rr|rr|rr|rr|rr}
\toprule
Function & \multicolumn{2}{|c|}{QuantumEvo} & \multicolumn{2}{c|}{BDD2Seq(B*)} & \multicolumn{2}{c|}{BDD2Seq(E*)} & \multicolumn{2}{c|}{SIFT} & \multicolumn{2}{c|}{SYMM\_SIFT} & \multicolumn{2}{c}{GROUP\_SIFT} \\
\cmidrule(lr){2-3}\cmidrule(lr){4-5}\cmidrule(lr){6-7}\cmidrule(lr){8-9}\cmidrule(lr){10-11}\cmidrule(lr){12-13}
 & Best & Time & Best & Time & Best & Time & Best & Time & Best & Time & Best & Time \\
\midrule
\texttt{4gt10\_22} & \textbf{17} & 0.04 & \textbf{17} & 1.49 & \textbf{17} & 2.13 & \textbf{17} & 0.02 & 23 & 0.01 & 23 & 0.02 \\
\texttt{4gt11\_23} & \textbf{5} & 0.06 & \textbf{5} & 0.05 & \textbf{5} & 0.01 & \textbf{5} & 0.01 & \textbf{5} & 0.01 & \textbf{5} & 0.03 \\
\texttt{4gt12\_24} & 17 & 0.06 & 17 & 0.04 & 17 & 0.01 & 17 & 0.01 & \textbf{16} & 0.01 & \textbf{16} & 0.03 \\
\texttt{4gt13\_25} & \textbf{10} & 0.06 & \textbf{10} & 0.05 & \textbf{10} & 0.01 & \textbf{10} & 0.01 & \textbf{10} & 0.01 & \textbf{10} & 0.03 \\
\texttt{4gt4\_20} & \textbf{19} & 0.06 & \textbf{19} & 0.04 & 25 & 0.01 & \textbf{19} & 0.03 & 23 & 0.01 & 23 & 0.03 \\
\texttt{4gt5\_21} & \textbf{12} & 0.05 & \textbf{12} & 0.05 & \textbf{12} & 0.01 & \textbf{12} & 0.03 & 18 & 0.01 & 18 & 0.03 \\
\texttt{4mod5\_8} & \textbf{24} & 0.06 & \textbf{24} & 0.05 & \textbf{24} & 0.02 & \textbf{24} & 0.03 & \textbf{24} & 0.01 & \textbf{24} & 0.02 \\
\texttt{4mod7\_26} & 86 & 0.06 & 86 & 0.04 & 86 & 0.01 & 86 & 0.03 & 86 & 0.01 & \textbf{80} & 0.03 \\
\texttt{5xp1\_90} & \textbf{254} & 0.06 & \textbf{254} & 0.09 & 280 & 0.01 & \textbf{254} & 0.03 & \textbf{254} & 0.01 & \textbf{254} & 0.03 \\
\texttt{9symml\_91} & \textbf{206} & 0.06 & \textbf{206} & 0.15 & \textbf{206} & 0.03 & \textbf{206} & 0.03 & \textbf{206} & 0.01 & \textbf{206} & 0.03 \\
\texttt{add6\_92} & \textbf{118} & 0.07 & \textbf{118} & 0.21 & 566 & 0.01 & 499 & 0.03 & 499 & 0.02 & 474 & 0.03 \\
\texttt{adr4\_93} & \textbf{74} & 0.05 & \textbf{74} & 0.12 & \textbf{74} & 0.02 & \textbf{74} & 0.03 & \textbf{74} & 0.01 & \textbf{74} & 0.03 \\
\texttt{alu1\_94} & \textbf{139} & 0.06 & \textbf{139} & 0.21 & \textbf{139} & 0.02 & \textbf{139} & 0.03 & \textbf{139} & 0.01 & \textbf{139} & 0.03 \\
\texttt{alu2\_96} & 1266 & 0.08 & \textbf{1218} & 0.17 & 1415 & 0.03 & 1436 & 0.03 & 1436 & 0.01 & 1366 & 0.03 \\
\texttt{alu3\_97} & \textbf{430} & 0.06 & \textbf{430} & 0.16 & 464 & 0.01 & 644 & 0.03 & 644 & 0.01 & 497 & 0.03 \\
\texttt{alu4\_98} & \textbf{4385} & 0.19 & 4455 & 0.33 & 4934 & 0.09 & 7222 & 0.04 & 7222 & 0.02 & 7623 & 0.04 \\
\texttt{alu\_9} & \textbf{29} & 0.05 & \textbf{29} & 0.06 & \textbf{29} & 0.02 & \textbf{29} & 0.03 & \textbf{29} & 0.01 & 35 & 0.03 \\
\texttt{apex2\_101} & \textbf{2663} & 0.62 & 3103 & 1.70 & 3178 & 0.16 & 5922 & 0.11 & 5393 & 0.08 & 6439 & 0.16 \\
\texttt{apex4\_103} & \textbf{8236} & 0.13 & 8409 & 0.35 & 8813 & 0.19 & 8343 & 0.02 & 8343 & 0.02 & 8381 & 0.04 \\
\texttt{apex5\_104} & \textbf{9184} & 0.97 & 9852 & 12.83 & 10697 & 0.20 & 10358 & 0.06 & 10228 & 0.03 & 10227 & 0.06 \\
\texttt{apla\_107} & \textbf{731} & 0.07 & 732 & 0.16 & 955 & 0.03 & 1002 & 0.03 & 1002 & 0.03 & 979 & 0.03 \\
\texttt{bw\_116} & 926 & 0.06 & \textbf{924} & 0.06 & 937 & 0.02 & 943 & 0.03 & 943 & 0.03 & 943 & 0.03 \\
\texttt{C17\_117} & \textbf{37} & 0.06 & \textbf{37} & 0.06 & \textbf{37} & 0.01 & 49 & 0.03 & 49 & 0.01 & \textbf{37} & 0.03 \\
\texttt{C7552\_119} & \textbf{202} & 0.06 & \textbf{202} & 0.06 & \textbf{202} & 0.01 & \textbf{202} & 0.03 & \textbf{202} & 0.01 & \textbf{202} & 0.03 \\
\texttt{clip\_124} & \textbf{515} & 0.06 & 695 & 0.13 & 698 & 0.02 & 704 & 0.03 & 704 & 0.02 & 520 & 0.03 \\
\texttt{cm150a\_128} & 186 & 0.52 & \textbf{136} & 0.66 & \textbf{136} & 0.24 & 186 & 0.21 & 186 & 0.16 & 186 & 0.20 \\
\texttt{cm151a\_129} & \textbf{298} & 0.10 & \textbf{298} & 0.43 & \textbf{298} & 0.02 & \textbf{298} & 0.03 & 310 & 0.02 & \textbf{298} & 0.03 \\
\texttt{cm152a\_130} & \textbf{62} & 0.07 & \textbf{62} & 0.18 & \textbf{62} & 0.02 & \textbf{62} & 0.03 & \textbf{62} & 0.02 & \textbf{62} & 0.03 \\
\texttt{cm163a\_133} & \textbf{244} & 0.08 & \textbf{244} & 0.32 & 267 & 0.02 & 273 & 0.03 & \textbf{244} & 0.02 & 267 & 0.03 \\
\texttt{cm42a\_125} & \textbf{117} & 0.06 & \textbf{117} & 0.05 & \textbf{117} & 0.01 & \textbf{117} & 0.03 & \textbf{117} & 0.02 & \textbf{117} & 0.03 \\
\texttt{cm82a\_126} & \textbf{44} & 0.05 & \textbf{44} & 0.05 & \textbf{44} & 0.01 & 82 & 0.03 & \textbf{44} & 0.01 & \textbf{44} & 0.03 \\
\texttt{cm85a\_127} & \textbf{203} & 0.06 & 227 & 0.18 & 283 & 0.03 & 275 & 0.03 & 221 & 0.01 & 221 & 0.03 \\
\texttt{cmb\_134} & \textbf{153} & 0.08 & \textbf{153} & 0.32 & \textbf{153} & 0.02 & 158 & 0.03 & \textbf{153} & 0.01 & \textbf{153} & 0.03 \\
\texttt{co14\_135} & \textbf{159} & 0.08 & \textbf{159} & 0.26 & \textbf{159} & 0.02 & \textbf{159} & 0.02 & \textbf{159} & 0.01 & \textbf{159} & 0.03 \\
\texttt{con1\_136} & 89 & 0.06 & \textbf{88} & 0.10 & 103 & 0.03 & 96 & 0.02 & 96 & 0.02 & 96 & 0.01 \\
\texttt{cordic\_138} & \textbf{304} & 0.16 & 344 & 0.62 & 446 & 0.05 & 325 & 0.02 & 323 & 0.02 & 318 & 0.04 \\
\texttt{cu\_141} & \textbf{219} & 0.08 & \textbf{219} & 0.26 & 230 & 0.02 & 220 & 0.01 & 220 & 0.01 & 224 & 0.02 \\
\texttt{dc1\_142} & 186 & 0.06 & \textbf{160} & 0.05 & 168 & 0.01 & \textbf{160} & 0.03 & \textbf{160} & 0.03 & 186 & 0.03 \\
\texttt{dc2\_143} & \textbf{431} & 0.06 & \textbf{431} & 0.11 & \textbf{431} & 0.03 & \textbf{431} & 0.03 & \textbf{431} & 0.02 & \textbf{431} & 0.03 \\
\texttt{decod24-enable\_32} & \textbf{38} & 0.06 & \textbf{38} & 0.04 & \textbf{38} & 0.01 & \textbf{38} & 0.03 & \textbf{38} & 0.03 & \textbf{38} & 0.03 \\
\texttt{decod\_137} & \textbf{202} & 0.05 & \textbf{202} & 0.06 & \textbf{202} & 0.01 & \textbf{202} & 0.03 & \textbf{202} & 0.03 & \textbf{202} & 0.03 \\
\texttt{dist\_144} & \textbf{975} & 0.06 & \textbf{975} & 0.13 & 979 & 0.04 & \textbf{975} & 0.03 & \textbf{975} & 0.03 & 979 & 0.03 \\
\texttt{dk17\_145} & \textbf{426} & 0.06 & 429 & 0.16 & 429 & 0.03 & \textbf{426} & 0.03 & \textbf{426} & 0.03 & 574 & 0.03 \\
\texttt{dk27\_146} & 140 & 0.06 & \textbf{131} & 0.13 & \textbf{131} & 0.02 & 140 & 0.03 & 140 & 0.03 & 141 & 0.03 \\
\texttt{e64\_149} & \textbf{886} & 0.39 & 1019 & 4.17 & 1208 & 0.17 & 907 & 0.05 & 907 & 0.05 & 1124 & 0.04 \\
\texttt{ex1010\_155} & 9681 & 0.23 & \textbf{9633} & 0.33 & 9810 & 0.14 & 9766 & 0.07 & 9766 & 0.04 & 9696 & 0.05 \\
\texttt{ex1\_150} & \textbf{8} & 0.06 & \textbf{8} & 0.07 & \textbf{8} & 0.01 & \textbf{8} & 0.03 & \textbf{8} & 0.01 & \textbf{8} & 0.03 \\
\texttt{ex2\_151} & \textbf{60} & 0.06 & \textbf{60} & 0.07 & 68 & 0.01 & 73 & 0.03 & 70 & 0.03 & \textbf{60} & 0.03 \\
\bottomrule
\end{tabular}
}
\end{table}

\clearpage

\begin{table}[p]
\centering
\caption{Full benchmark results, RevLib functions, prior baselines, part~2 of~2.}
\label{tab:full-base-revlib-b}
\scriptsize
\setlength{\tabcolsep}{2.4pt}
\setlength{\extrarowheight}{0.35pt}
\renewcommand{\arraystretch}{0.86}
\resizebox{\linewidth}{!}{%
\begin{tabular}{l|rr|rr|rr|rr|rr|rr}
\toprule
Function & \multicolumn{2}{|c|}{QuantumEvo} & \multicolumn{2}{c|}{BDD2Seq(B*)} & \multicolumn{2}{c|}{BDD2Seq(E*)} & \multicolumn{2}{c|}{SIFT} & \multicolumn{2}{c|}{SYMM\_SIFT} & \multicolumn{2}{c}{GROUP\_SIFT} \\
\cmidrule(lr){2-3}\cmidrule(lr){4-5}\cmidrule(lr){6-7}\cmidrule(lr){8-9}\cmidrule(lr){10-11}\cmidrule(lr){12-13}
 & Best & Time & Best & Time & Best & Time & Best & Time & Best & Time & Best & Time \\
\midrule
\texttt{ex3\_152} & \textbf{38} & 0.06 & \textbf{38} & 0.08 & \textbf{38} & 0.02 & 61 & 0.03 & 61 & 0.03 & 43 & 0.03 \\
\texttt{ex5p\_154} & 1843 & 0.11 & \textbf{1837} & 0.17 & 1871 & 0.06 & 1843 & 0.04 & 1843 & 0.05 & 1843 & 0.05 \\
\texttt{example2\_156} & 1266 & 0.08 & \textbf{1218} & 0.18 & 1415 & 0.03 & 1436 & 0.03 & 1436 & 0.03 & 1366 & 0.02 \\
\texttt{f2\_158} & \textbf{108} & 0.05 & \textbf{108} & 0.05 & 113 & 0.01 & 113 & 0.03 & 113 & 0.03 & 113 & 0.02 \\
\texttt{f51m\_159} & \textbf{2422} & 0.14 & 2427 & 0.46 & 2517 & 0.14 & 5392 & 0.05 & 5392 & 0.05 & 4603 & 0.05 \\
\texttt{frg1\_160} & \textbf{559} & 0.17 & 598 & 0.95 & 629 & 0.03 & 747 & 0.03 & 747 & 0.03 & 827 & 0.03 \\
\texttt{frg2\_161} & \textbf{7017} & 1.47 & 8584 & 18.95 & 10768 & 0.37 & 12468 & 0.07 & 12361 & 0.08 & 12224 & 0.09 \\
\texttt{in0\_162} & 2299 & 0.11 & 2296 & 0.31 & 2417 & 0.04 & \textbf{2283} & 0.01 & \textbf{2283} & 0.03 & 2328 & 0.04 \\
\texttt{inc\_170} & 592 & 0.06 & \textbf{579} & 0.09 & \textbf{579} & 0.01 & \textbf{579} & 0.01 & \textbf{579} & 0.03 & 592 & 0.03 \\
\texttt{life\_175} & \textbf{204} & 0.06 & \textbf{204} & 0.14 & 210 & 0.01 & \textbf{204} & 0.01 & 210 & 0.01 & \textbf{204} & 0.03 \\
\texttt{majority\_176} & \textbf{41} & 0.06 & \textbf{41} & 0.06 & \textbf{41} & 0.01 & \textbf{41} & 0.01 & \textbf{41} & 0.02 & \textbf{41} & 0.03 \\
\texttt{max46\_177} & \textbf{556} & 0.07 & 562 & 0.14 & 596 & 0.02 & 598 & 0.01 & 598 & 0.03 & 598 & 0.03 \\
\texttt{misex1\_178} & \textbf{279} & 0.06 & 281 & 0.12 & 289 & 0.02 & 288 & 0.01 & 288 & 0.03 & 289 & 0.03 \\
\texttt{misex3\_180} & \textbf{3789} & 0.16 & 3830 & 0.33 & 4054 & 0.09 & 4661 & 0.03 & 4661 & 0.05 & 4619 & 0.05 \\
\texttt{misex3c\_181} & \textbf{3936} & 0.16 & 3969 & 0.29 & 4457 & 0.05 & 4769 & 0.02 & 4769 & 0.04 & 4823 & 0.04 \\
\texttt{mlp4\_184} & \textbf{1158} & 0.07 & 1191 & 0.13 & 1228 & 0.03 & \textbf{1158} & 0.02 & \textbf{1158} & 0.03 & \textbf{1158} & 0.03 \\
\texttt{mux\_185} & 170 & 0.55 & \textbf{135} & 0.90 & 224 & 0.26 & 170 & 0.23 & 170 & 0.25 & 170 & 0.35 \\
\texttt{one-two-three\_27} & \textbf{44} & 0.06 & \textbf{44} & 0.04 & \textbf{44} & 0.01 & \textbf{44} & 0.01 & \textbf{44} & 0.03 & \textbf{44} & 0.01 \\
\texttt{parity\_188} & \textbf{31} & 0.50 & \textbf{31} & 0.82 & \textbf{31} & 0.44 & \textbf{31} & 0.18 & \textbf{31} & 0.21 & \textbf{31} & 0.17 \\
\texttt{pcler8\_190} & \textbf{124} & 0.08 & \textbf{124} & 0.32 & 137 & 0.01 & \textbf{124} & 0.01 & \textbf{124} & 0.03 & 137 & 0.01 \\
\texttt{pdc\_191} & 6599 & 0.41 & 6742 & 0.57 & 6742 & 0.23 & \textbf{6500} & 0.13 & \textbf{6500} & 0.19 & \textbf{6500} & 0.13 \\
\texttt{pm1\_192} & \textbf{117} & 0.06 & \textbf{117} & 0.05 & \textbf{117} & 0.02 & \textbf{117} & 0.01 & \textbf{117} & 0.01 & \textbf{117} & 0.01 \\
\texttt{radd\_193} & \textbf{74} & 0.06 & \textbf{74} & 0.11 & 140 & 0.01 & 217 & 0.01 & 192 & 0.03 & 212 & 0.01 \\
\texttt{rd53\_68} & \textbf{98} & 0.05 & \textbf{98} & 0.05 & \textbf{98} & 0.01 & \textbf{98} & 0.01 & \textbf{98} & 0.03 & \textbf{98} & 0.01 \\
\texttt{rd73\_69} & \textbf{217} & 0.06 & \textbf{217} & 0.09 & \textbf{217} & 0.02 & \textbf{217} & 0.01 & \textbf{217} & 0.03 & \textbf{217} & 0.01 \\
\texttt{rd84\_70} & \textbf{304} & 0.06 & \textbf{304} & 0.12 & \textbf{304} & 0.02 & \textbf{304} & 0.01 & \textbf{304} & 0.03 & \textbf{304} & 0.01 \\
\texttt{root\_197} & \textbf{444} & 0.06 & \textbf{444} & 0.11 & 446 & 0.01 & \textbf{444} & 0.01 & \textbf{444} & 0.03 & \textbf{444} & 0.01 \\
\texttt{ryy6\_198} & 107 & 0.09 & \textbf{103} & 0.33 & 107 & 0.02 & 133 & 0.02 & 132 & 0.03 & 119 & 0.03 \\
\texttt{sao2\_199} & \textbf{653} & 0.07 & 657 & 0.16 & 698 & 0.01 & 667 & 0.01 & 667 & 0.03 & 684 & 0.03 \\
\texttt{seq\_201} & \textbf{9080} & 1.17 & 9349 & 2.12 & 9908 & 0.33 & 19362 & 0.24 & 19350 & 0.28 & 15309 & 0.27 \\
\texttt{sf\_232} & \textbf{21} & 0.06 & 31 & 0.05 & 36 & 0.02 & 36 & 0.01 & 36 & 0.03 & \textbf{21} & 0.03 \\
\texttt{spla\_202} & \textbf{5858} & 0.33 & 5947 & 0.52 & 6092 & 0.19 & 5925 & 0.08 & 5925 & 0.11 & 5925 & 0.10 \\
\texttt{sqn\_203} & \textbf{356} & 0.06 & 374 & 0.10 & 426 & 0.02 & 426 & 0.01 & 426 & 0.03 & 426 & 0.03 \\
\texttt{sqr6\_204} & 486 & 0.06 & \textbf{470} & 0.08 & 524 & 0.02 & 486 & 0.01 & 486 & 0.03 & 486 & 0.03 \\
\texttt{sqrt8\_205} & \textbf{221} & 0.06 & 237 & 0.11 & 283 & 0.01 & 240 & 0.01 & 240 & 0.03 & 240 & 0.03 \\
\texttt{squar5\_206} & \textbf{232} & 0.06 & \textbf{232} & 0.07 & 264 & 0.02 & 253 & 0.01 & 253 & 0.03 & \textbf{232} & 0.03 \\
\texttt{sym10\_207} & \textbf{253} & 0.07 & \textbf{253} & 0.18 & \textbf{253} & 0.03 & \textbf{253} & 0.02 & \textbf{253} & 0.03 & \textbf{253} & 0.03 \\
\texttt{sym6\_63} & \textbf{93} & 0.06 & \textbf{93} & 0.08 & \textbf{93} & 0.01 & \textbf{93} & 0.01 & \textbf{93} & 0.03 & \textbf{93} & 0.03 \\
\texttt{sym9\_71} & \textbf{206} & 0.06 & \textbf{206} & 0.15 & \textbf{206} & 0.03 & \textbf{206} & 0.01 & \textbf{206} & 0.03 & \textbf{206} & 0.03 \\
\texttt{t481\_208} & 147 & 0.10 & \textbf{139} & 0.34 & 140 & 0.03 & 147 & 0.02 & 152 & 0.03 & 152 & 0.03 \\
\texttt{table3\_209} & \textbf{5898} & 0.18 & 5954 & 0.36 & 7005 & 0.08 & 6276 & 0.02 & 6276 & 0.04 & 5927 & 0.03 \\
\texttt{tial\_214} & \textbf{4460} & 0.13 & 4475 & 0.33 & 5157 & 0.09 & 7609 & 0.02 & 7609 & 0.04 & 4852 & 0.04 \\
\texttt{urf4\_89} & \textbf{28406} & 0.41 & 28488 & 0.68 & 28833 & 0.47 & 28523 & 0.06 & 28523 & 0.10 & 28523 & 0.11 \\
\texttt{wim\_220} & 107 & 0.06 & \textbf{103} & 0.05 & 108 & 0.01 & 107 & 0.01 & 107 & 0.03 & 107 & 0.03 \\
\texttt{x2\_223} & \textbf{191} & 0.06 & \textbf{191} & 0.17 & \textbf{191} & 0.03 & 273 & 0.01 & 273 & 0.03 & 283 & 0.03 \\
\texttt{xor5\_195} & \textbf{8} & 0.05 & \textbf{8} & 0.05 & \textbf{8} & 0.01 & \textbf{8} & 0.01 & \textbf{8} & 0.03 & \textbf{8} & 0.03 \\
\texttt{z4\_224} & \textbf{66} & 0.06 & \textbf{66} & 0.09 & \textbf{66} & 0.01 & \textbf{66} & 0.01 & \textbf{66} & 0.03 & \textbf{66} & 0.03 \\
\texttt{z4ml\_225} & \textbf{66} & 0.05 & \textbf{66} & 0.10 & \textbf{66} & 0.02 & \textbf{66} & 0.01 & \textbf{66} & 0.03 & \textbf{66} & 0.03 \\
\bottomrule
\end{tabular}
}
\end{table}

\clearpage

\begin{table}[p]
\centering
\caption{Full benchmark results, ISCAS85/89 and LGSynth91 functions, prior baselines.}
\label{tab:full-base-iscas-lgs}
\scriptsize
\setlength{\tabcolsep}{2.4pt}
\setlength{\extrarowheight}{0.35pt}
\renewcommand{\arraystretch}{0.86}
\resizebox{\linewidth}{!}{%
\begin{tabular}{l|rr|rr|rr|rr|rr|rr}
\toprule
Function & \multicolumn{2}{|c|}{QuantumEvo} & \multicolumn{2}{c|}{BDD2Seq(B*)} & \multicolumn{2}{c|}{BDD2Seq(E*)} & \multicolumn{2}{c|}{SIFT} & \multicolumn{2}{c|}{SYMM\_SIFT} & \multicolumn{2}{c}{GROUP\_SIFT} \\
\cmidrule(lr){2-3}\cmidrule(lr){4-5}\cmidrule(lr){6-7}\cmidrule(lr){8-9}\cmidrule(lr){10-11}\cmidrule(lr){12-13}
 & Best & Time & Best & Time & Best & Time & Best & Time & Best & Time & Best & Time \\
\midrule
\texttt{apex6\_orig}\textsuperscript{L} & \textbf{3736} & 0.65 & 3895 & 16.72 & 4021 & 0.16 & 4822 & 0.05 & 4810 & 0.03 & 4269 & 0.03 \\
\texttt{apex7\_orig}\textsuperscript{L} & \textbf{1431} & 0.28 & 1455 & 2.40 & 1456 & 0.07 & 2653 & 0.04 & 2653 & 0.04 & 2377 & 0.04 \\
\texttt{b1\_orig}\textsuperscript{L} & \textbf{22} & 0.06 & \textbf{22} & 0.04 & \textbf{22} & 0.01 & \textbf{22} & 0.03 & \textbf{22} & 0.03 & \textbf{22} & 0.03 \\
\texttt{b9\_orig}\textsuperscript{L} & \textbf{713} & 0.17 & 737 & 1.71 & 737 & 0.05 & 804 & 0.03 & 810 & 0.03 & 836 & 0.03 \\
\texttt{c17}\textsuperscript{I} & \textbf{37} & 0.06 & \textbf{37} & 0.06 & 49 & 0.01 & 49 & 0.03 & 49 & 0.03 & \textbf{37} & 0.03 \\
\texttt{c432}\textsuperscript{I} & \textbf{10494} & 245.19 & 12255 & 10.01 & 20391 & 6.79 & 11101 & 0.04 & 11101 & 0.02 & 11141 & 0.04 \\
\texttt{c880}\textsuperscript{I} & \textbf{34241} & 329.32 & 37802 & 23.00 & 39992 & 15.52 & 61079 & 0.53 & 61079 & 0.65 & 78113 & 0.59 \\
\texttt{c8\_orig}\textsuperscript{L} & 436 & 0.14 & \textbf{432} & 0.86 & 436 & 0.04 & 445 & 0.03 & 445 & 0.02 & 486 & 0.03 \\
\texttt{cc\_orig}\textsuperscript{L} & 283 & 0.08 & \textbf{272} & 0.51 & 278 & 0.03 & 385 & 0.03 & 385 & 0.01 & 379 & 0.03 \\
\texttt{cht\_orig}\textsuperscript{L} & \textbf{714} & 0.12 & \textbf{714} & 2.22 & \textbf{714} & 0.06 & \textbf{714} & 0.03 & \textbf{714} & 0.02 & \textbf{714} & 0.03 \\
\texttt{cm138a\_orig}\textsuperscript{L} & \textbf{104} & 0.06 & \textbf{104} & 0.08 & \textbf{104} & 0.01 & \textbf{104} & 0.02 & \textbf{104} & 0.02 & \textbf{104} & 0.03 \\
\texttt{cm150a\_orig}\textsuperscript{L} & 186 & 0.52 & \textbf{136} & 0.75 & \textbf{136} & 0.34 & 186 & 0.38 & 186 & 0.25 & 186 & 0.32 \\
\texttt{cm151a\_orig}\textsuperscript{L} & 92 & 0.07 & \textbf{70} & 0.21 & \textbf{70} & 0.02 & 92 & 0.03 & 92 & 0.02 & 92 & 0.03 \\
\texttt{cm152a\_orig}\textsuperscript{L} & \textbf{62} & 0.07 & \textbf{62} & 0.18 & 81 & 0.02 & \textbf{62} & 0.03 & \textbf{62} & 0.02 & \textbf{62} & 0.03 \\
\texttt{cm162a\_orig}\textsuperscript{L} & 222 & 0.08 & 222 & 0.26 & 222 & 0.03 & \textbf{192} & 0.03 & 258 & 0.02 & 244 & 0.03 \\
\texttt{cm163a\_orig}\textsuperscript{L} & \textbf{124} & 0.08 & 136 & 0.32 & 136 & 0.03 & \textbf{124} & 0.03 & \textbf{124} & 0.02 & 137 & 0.03 \\
\texttt{cm42a\_orig}\textsuperscript{L} & \textbf{117} & 0.05 & \textbf{117} & 0.04 & \textbf{117} & 0.02 & \textbf{117} & 0.03 & \textbf{117} & 0.01 & \textbf{117} & 0.03 \\
\texttt{cm82a\_orig}\textsuperscript{L} & \textbf{44} & 0.06 & \textbf{44} & 0.05 & \textbf{44} & 0.01 & 82 & 0.03 & \textbf{44} & 0.01 & \textbf{44} & 0.03 \\
\texttt{cm85a\_orig}\textsuperscript{L} & \textbf{203} & 0.07 & 225 & 0.18 & 225 & 0.02 & 275 & 0.03 & 221 & 0.01 & 221 & 0.03 \\
\texttt{cmb\_orig}\textsuperscript{L} & \textbf{153} & 0.09 & \textbf{153} & 0.32 & \textbf{153} & 0.03 & 158 & 0.03 & \textbf{153} & 0.01 & \textbf{153} & 0.03 \\
\texttt{comp\_orig}\textsuperscript{L} & \textbf{796} & 2.41 & 824 & 40.21 & 824 & 28.75 & 961 & 1.77 & 1071 & 1.96 & 884 & 1.79 \\
\texttt{cordic\_orig}\textsuperscript{L} & \textbf{304} & 0.12 & 320 & 0.62 & 320 & 0.06 & 325 & 0.01 & 323 & 0.01 & 318 & 0.03 \\
\texttt{count\_orig}\textsuperscript{L} & \textbf{441} & 0.17 & \textbf{441} & 1.27 & \textbf{441} & 0.04 & \textbf{441} & 0.01 & \textbf{441} & 0.01 & \textbf{441} & 0.02 \\
\texttt{cu\_orig}\textsuperscript{L} & \textbf{219} & 0.08 & 224 & 0.26 & 231 & 0.02 & 220 & 0.01 & 220 & 0.01 & 224 & 0.03 \\
\texttt{dalu\_orig}\textsuperscript{L} & \textbf{5230} & 71.73 & 5297 & 55.53 & 5652 & 28.17 & 31145 & 66.13 & 31145 & 71.95 & 47170 & 75.07 \\
\texttt{decod\_orig}\textsuperscript{L} & \textbf{202} & 0.06 & \textbf{202} & 0.05 & \textbf{202} & 0.01 & \textbf{202} & 0.03 & \textbf{202} & 0.03 & \textbf{202} & 0.03 \\
\texttt{example2\_orig}\textsuperscript{L} & \textbf{1790} & 0.28 & 1800 & 6.83 & 1806 & 0.08 & 2066 & 0.03 & 2066 & 0.02 & 2041 & 0.03 \\
\texttt{frg1\_orig}\textsuperscript{L} & \textbf{559} & 0.15 & 757 & 0.95 & 762 & 0.03 & 747 & 0.03 & 747 & 0.02 & 827 & 0.02 \\
\texttt{frg2\_orig}\textsuperscript{L} & \textbf{7017} & 1.44 & 7189 & 19.35 & 7189 & 0.35 & 12468 & 0.03 & 12361 & 0.06 & 12224 & 0.07 \\
\texttt{k2\_orig}\textsuperscript{L} & \textbf{10156} & 0.97 & 10432 & 2.38 & 10686 & 0.19 & 11058 & 0.07 & 11058 & 0.13 & 11275 & 0.09 \\
\texttt{lal\_orig}\textsuperscript{L} & 496 & 0.10 & \textbf{478} & 0.76 & 496 & 0.04 & 527 & 0.01 & 768 & 0.02 & 724 & 0.03 \\
\texttt{majority\_orig}\textsuperscript{L} & \textbf{41} & 0.06 & \textbf{41} & 0.05 & \textbf{41} & 0.01 & \textbf{41} & 0.01 & \textbf{41} & 0.03 & \textbf{41} & 0.03 \\
\texttt{mux\_orig}\textsuperscript{L} & 170 & 0.52 & \textbf{135} & 0.84 & \textbf{135} & 0.40 & 170 & 0.24 & 170 & 0.35 & 170 & 0.26 \\
\texttt{my\_adder\_orig}\textsuperscript{L} & \textbf{352} & 1.24 & \textbf{352} & 59.50 & \textbf{352} & 48.54 & \textbf{352} & 0.15 & \textbf{352} & 0.35 & \textbf{352} & 0.16 \\
\texttt{pair\_orig}\textsuperscript{L} & \textbf{20236} & 6.24 & 20754 & 28.19 & 20799 & 1.06 & 46491 & 0.11 & 46391 & 0.17 & 49917 & 0.15 \\
\texttt{parity\_orig}\textsuperscript{L} & \textbf{31} & 0.09 & \textbf{31} & 0.83 & \textbf{31} & 0.61 & \textbf{31} & 0.01 & \textbf{31} & 0.03 & \textbf{31} & 0.01 \\
\texttt{pcle\_orig}\textsuperscript{L} & \textbf{298} & 0.09 & \textbf{298} & 0.43 & \textbf{298} & 0.02 & \textbf{298} & 0.01 & 310 & 0.03 & \textbf{298} & 0.01 \\
\texttt{pcler8\_orig}\textsuperscript{L} & 639 & 0.15 & 765 & 0.80 & 765 & 0.02 & \textbf{638} & 0.01 & \textbf{638} & 0.04 & 639 & 0.01 \\
\texttt{pm1\_orig}\textsuperscript{L} & 244 & 0.08 & \textbf{234} & 0.33 & 273 & 0.03 & 273 & 0.01 & 244 & 0.01 & 267 & 0.01 \\
\texttt{rot\_orig}\textsuperscript{L} & \textbf{23926} & 845.19 & 38453 & 37.98 & 48344 & 15.69 & 78639 & 0.56 & 78374 & 0.62 & 110936 & 0.74 \\
\texttt{s1196\_orig}\textsuperscript{I} & \textbf{3765} & 0.14 & 4104 & 0.30 & 4110 & 0.05 & 4691 & 0.02 & 4691 & 0.04 & 4653 & 0.04 \\
\texttt{sct\_orig}\textsuperscript{L} & \textbf{375} & 0.10 & 405 & 0.43 & 513 & 0.02 & 545 & 0.01 & 545 & 0.03 & 548 & 0.03 \\
\texttt{tcon\_orig}\textsuperscript{L} & \textbf{88} & 0.06 & \textbf{88} & 0.36 & \textbf{88} & 0.02 & \textbf{88} & 0.01 & \textbf{88} & 0.03 & \textbf{88} & 0.02 \\
\texttt{term1\_orig}\textsuperscript{L} & 497 & 0.18 & \textbf{480} & 1.25 & 483 & 0.06 & 1193 & 0.02 & 1151 & 0.04 & 1070 & 0.04 \\
\texttt{too\_large\_orig}\textsuperscript{L} & \textbf{2676} & 0.71 & 3744 & 2.19 & 5379 & 0.46 & 5922 & 0.07 & 5393 & 0.12 & 6334 & 0.10 \\
\texttt{ttt2\_orig}\textsuperscript{L} & 736 & 0.13 & 749 & 0.66 & 1006 & 0.04 & \textbf{727} & 0.02 & \textbf{727} & 0.04 & 1139 & 0.03 \\
\texttt{unreg\_orig}\textsuperscript{L} & \textbf{494} & 0.12 & \textbf{494} & 1.36 & \textbf{494} & 0.04 & \textbf{494} & 0.01 & \textbf{494} & 0.03 & \textbf{494} & 0.03 \\
\texttt{vda\_orig}\textsuperscript{L} & \textbf{4169} & 0.16 & 4224 & 0.42 & 4224 & 0.08 & 4477 & 0.03 & 4477 & 0.05 & 4460 & 0.05 \\
\texttt{x1\_orig}\textsuperscript{L} & \textbf{3062} & 0.50 & 3567 & 2.68 & 3703 & 0.16 & 3678 & 0.02 & 4189 & 0.04 & 4172 & 0.05 \\
\texttt{x2\_orig}\textsuperscript{L} & \textbf{191} & 0.06 & \textbf{191} & 0.15 & \textbf{191} & 0.02 & 273 & 0.01 & 273 & 0.03 & 283 & 0.03 \\
\texttt{x3\_orig}\textsuperscript{L} & \textbf{3736} & 0.62 & 3941 & 16.78 & 4645 & 0.19 & 4822 & 0.03 & 4810 & 0.05 & 4269 & 0.03 \\
\texttt{x4\_orig}\textsuperscript{L} & \textbf{2437} & 0.37 & 2662 & 8.31 & 2782 & 0.11 & 4470 & 0.02 & 4470 & 0.04 & 4334 & 0.04 \\
\bottomrule
\end{tabular}
}
\par\smallskip{\footnotesize\raggedright Superscripts denote source dataset: \textsuperscript{I}~ISCAS85/89; \textsuperscript{L}~LGSynth91.\par}
\end{table}

\end{document}